\documentclass{article}

\PassOptionsToPackage{numbers, compress}{natbib}

\usepackage[main, final]{neurips_2026}
\usepackage[utf8]{inputenc} 
\ifdefined\DeclareUnicodeCharacter
  \DeclareUnicodeCharacter{03B2}{\ensuremath{\beta}}
\fi
\usepackage[T1]{fontenc}    
\usepackage{hyperref}       
\usepackage{url}            
\usepackage{booktabs}       
\usepackage{amsfonts}       
\usepackage{nicefrac}       
\usepackage{microtype}      
\usepackage{xcolor}         
\usepackage{threeparttable}
\usepackage{multirow}
\usepackage{makecell}
\usepackage{enumitem}
\usepackage{amsmath}
\usepackage{amssymb}
\usepackage{mathtools}
\usepackage{amsthm}
\usepackage{graphicx}
\usepackage{caption}
\usepackage{xcolor}
\usepackage{wrapfig}

\hypersetup{
    colorlinks=true,
    linkcolor=blue,
    citecolor=black,
    urlcolor=magenta
}
\newcommand{\blackdiamond}{\blacklozenge}

\usepackage{pifont}   
\usepackage{xcolor}   

\usepackage[textsize=tiny]{todonotes}

\title{Learning Biological Hierarchies in Single-Cell Foundation Models}

\author{
  Xiangyu Guo \\
  Duke University \\
  \texttt{xiangyu.guo@duke.edu}
  \And
  Ricardo Henao \\
  Duke University \\
  \texttt{ricardo.henao@duke.edu}
}

\begin{document}

\maketitle

\begin{abstract}
Cell types are organized by taxonomic relationships that define lineages in the so called cell ontology. However, most existing foundation models typically ignore cell-type lineages encoded in this cell ontology. We introduce a biologically informed foundation model called \textbf{scOntoFM}\footnote{The source code and pre-trained model are available on
\href{https://github.com/woweizhi/scOntoFM}{GitHub} and
\href{https://huggingface.co/xiangyguo/scOntoFM}{Hugging Face},
respectively.}, which embeds hierarchical cell ontology into representation learning. Using Lowest Common Ancestor (LCA) distances from the ontology graph, our framework pairs efficient offline triplet sampling with hierarchical cell-ontology learning.
This approach jointly preserves local neighborhoods and shapes the global embedding geometry to reflect the ontology structure. Across diverse cell- and gene-level benchmarks, the model consistently improves performance, with especially strong gains in zero-shot settings. Our model also balances batch integration with biological structure preservation, providing complementary perspectives to existing embedding evaluation frameworks and supporting scalable biological discovery.
\end{abstract}

\section{Introduction}
Current single-cell RNA sequencing (scRNA-seq) technology has revolutionized our ability to characterize cellular heterogeneity and uncover the molecular mechanisms underlying healthy and diseased tissues \cite{scRNA-review}. Driven by advances in high-throughput sequencing and large-scale atlas construction, the field has witnessed an explosion of massive single-cell datasets in recent years \cite{HuMap2019, Brain_atlas_ECKER2017542, Lung_atlas_Sikkema2023, CZI_2023}. This data deluge has created an urgent demand for computational frameworks capable of fully utilizing these resources for broad biological discovery. Inspired by the transformative success of Large Language Models (LLMs) in Natural Language Processing (NLP), the community has recently developed Single-Cell Foundation Models (scFMs), such as Geneformer \cite{Geneformer-Theodoris2023}, scGPT \cite{scGPT-Cui2024}, scFoundation \cite{scFoundation-Hao2024}, scCello \cite{scCello-yuan2024cell}, CellWhisperer \cite{cellWhisper-Schaefer2025}, {\em etc}. By pretraining on large corpora of cell-level transcriptomic data, these models aim to learn generalizable cellular representations that can be transferred to diverse downstream tasks, ranging from cell-type annotation to gene-perturbation prediction \cite{scFM_review_Szaata2024}. We provide a detailed description of the Related Work in Appendix~\ref{suppl:related_work}.

Contrastive learning is widely adopted in current scFMs as an auxiliary objective to learn discriminative features between cell types \cite{zhao2023largeScaleRepLearning,SCimilarity-Heimberg2024,scCello-yuan2024cell,Tabulading2026}. Among these approaches, ontology-aware methods such as SCimilarity \cite{SCimilarity-Heimberg2024} and scCello \cite{scCello-yuan2024cell} further incorporate cell ontology as supervised signals during model training, leveraging online triplet sampling and vanilla contrastive objectives to encourage biologically meaningful representation learning. However, scaling such frameworks to atlas-level datasets introduces substantial computational challenges, as standard online triplet sampling is highly inefficient for massive datasets. In random mini-batches, informative positive pairs are rarely encountered (Appendix Figure (AFigure)~\ref{fig:online_triplet_sampling}), leading to weak supervision and suboptimal metric learning \cite{hierarchical-tripletLoss-Ge_2018_ECCV, dynamic-metric-learning}. Moreover, existing methods such as scCello mainly focus on local cell-type relational alignment guided by the ontology structure, while overlooking the global hierarchical structure encoded in the cell ontology \cite{cellOnto-perspective-OsumiSutherland2021, concord-Zhu2026}. Consequently, the learned latent space may not faithfully reflect biological phylogeny and developmental lineages, limiting both biological interpretability and the ability to capture dynamic cellular processes \cite{scGraph-Wang2025, Wu2025-my}.

To address these challenges, we propose scOntoFM, a novel biologically informed foundation model that explicitly integrates hierarchical cell ontology into a representation learning framework. Cell Ontology \cite{cellOnto-perspective-OsumiSutherland2021} provides a curated hierarchy of cell-type identities that complements transcriptomic supervision. Consistent with prior single-cell studies linking ontology relationships to transcriptomic similarity \cite{onClass-Wang2021, UCE-Rosen2026, Wu2025-my}, we used the Lowest Common Ancestor (LCA) distance \cite{LCA-shi2024, LCA-bertinetto2020} from the cell ontology graph as an interpretable structural prior and an approximate proxy for transcriptomic distance. This prior serves as a guide for both data sampling and loss optimization in scOntoFM. Our contributions are threefold.

\begin{itemize}[leftmargin=8pt,itemsep=0mm,topsep=0mm]
    \item \textbf{Efficient Ontology-aware Offline Sampling:} We introduce the \emph{Cell-Dataset Index Map}, a hierarchical indexing structure for efficient offline triplet sampling across cell types and datasets. Guided by the LCA distance matrix, it prioritizes cross-dataset positives and semi-hard and hard negatives drawn from semantically close cell types in the ontology graph, thus overcoming the inefficiency of random online sampling commonly used in current scFMs, {\em e.g.}, scCello \cite{scCello-yuan2024cell}.
    \item \textbf{Hierarchical Structure Learning:} We incorporate the ontology structure into representation learning through a \emph{dynamic margin triplet loss}, where the margin is determined by the LCA distance between the anchor and negative cell types and the ontology depth of the anchor cell type. In addition, a complementary \emph{rank-order alignment loss} enforces consistency between the relative ordering of cell-type centroid distances in the embedding space and their ontology-based LCA distances within each batch. These two synergistic losses enforce both fine-grained local discrimination and global hierarchical consistency, producing an embedding space that better reflects the structure of the cell-type ontology.
    \item \textbf{Comprehensive Evaluation:} We benchmark our model against state-of-the-art (SOTA) methods across diverse cell- and gene-level tasks. Our results demonstrate superior generalizability, particularly in zero-shot settings. Crucially, as highlighted by recent critiques of embedding metrics ({\em e.g.}, the ``drifting islands'' phenomenon) \cite{scGraph-Wang2025}, purely aligning cell embeddings with cell-type annotations will lead to biologically problematic embeddings. 
    Our evaluation confirms that our approach achieves a superior balance between the well established scIB metrics \cite{scIB-Luecken2021}, which measure the proximity of cells that belong to the same cell-type cluster, and how well cells of the same cell type mix-up across batches/datasets in single cell studies,
    and the more rigorous biological structure preservation metric of model embeddings \cite{Wu2025-my}.
\end{itemize}

\section{Methods}
\label{sec:methods}
\begin{figure*}[t]
\centering
\includegraphics[width=1\textwidth]
{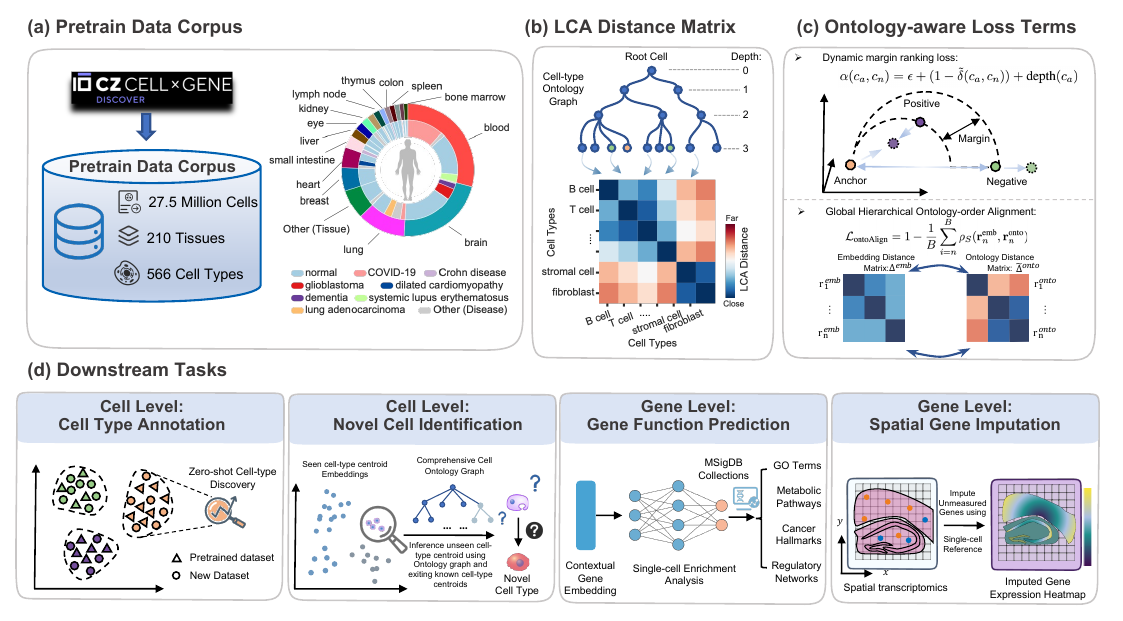}
\vspace{-3mm}
\caption{Proposed model workflow. (a) scOntoFM pre-training data corpus details (left), and the distribution of major tissues and diseases in the corpus (right). (b) An illustration of cell ontology graph and its derived LCA distance matrix. (c) Ontology-aware loss terms in scOntoFM, the dynamic margin $\alpha(c_a,c_n)$ is controlled by the distance between the anchor cell-type $c_a$ and negative cell-type $c_n$, and the depth of the anchor $c_a$ in the ontology graph. Down below is the hierarchical ontology alignment loss, in which $\Delta^{\text{emb}}$ and $ \tilde{\Delta}^{\text{onto}} \in \mathbb{R}^{B \times B}$ represent correspondingly the distance matrix of our model embeddings and its corresponding LCA distance matrix in the batch (of size $B$), and where $r_n^{emb}$ and $r_n^{onto}$ are the soft-ranking rows in their respective distance matrices. $\rho_S(\cdot, \cdot)$ is the Spearman correlation for the soft-ranking vectors. (d) Downstream tasks of scOntoFM, from left to right and grouped into cell- and gene-level tasks as presented in the results.}
\label{fig:main_fig}
\vspace{-3mm}
\end{figure*}

Figure~\ref{fig:main_fig} provides a high-level overview of the proposed scOntoFM workflow. We first describe the pre-training data processing, the construction of the Lowest Common Ancestor (LCA) distance matrix
in Section~\ref{sec:data_process}, which is followed by the offline triplet sampling, model architecture, and model pre-training objectives in Section~\ref{sec:offline_sampling}, Section~\ref{sec:model_architecture} and \ref{sec:objectives}, respectively.

\subsection{Data Processing}
\label{sec:data_process}
{\bf Single-cell Data}
Let $\mathcal{D} = \{x_n, c_n\}_{n=0}^N$ denote a single-cell RNA-seq dataset, where $x_n \in \mathbb{R}^{|\mathcal{M}|}$ is the raw gene expression vector in cell $n$, and $\mathcal{M}=\{m_1, \dots, m_{\mathcal{|M|}}\}$ represents the global gene set (vocabulary) of size $|\mathcal{M}|$ defined by our pre-training corpus. Each cell is associated with a cell type label $c_n \in \mathcal{V}$, where $\mathcal{V}$ represents the set of all cell type nodes in the cell ontology graph, which we will introduce in the next section. An overview of our pre-training data corpus is presented in Figure~\ref{fig:main_fig}(a).

We follow the same data pre-processing steps 
as Geneformer \cite{Geneformer-Theodoris2023}: first we normalize the library size of each cell to 10,000 to account for sequencing depth variation, and re-normalize each gene expression value by its non-zero median over the entire pre-training dataset, {\em i.e.},
\vspace{-5pt}
\[
\resizebox{0.8\hsize}{!}{$ 
\begin{aligned}
\label{eq:data_normalization}
x'_{n,m} = 10^4 \frac{x_{n,m}}{\sum_{m' \in \mathcal{M}} x_{n,m'}}, \hspace{4mm}
x''_{n,m} = \frac{x'_{n,m}}{\operatorname{median}\!\left\{
x'_{n,m} \;\middle|\; x'_{n,m} > 0\right\}} , \qquad n=1,\dots,N,\; m \in \mathcal{M} ,
\end{aligned}
$}
\]
%
%
%
where $x''_{n,m}$ is the normalized gene expression for cell $n$ and gene $m$.
Following normalization, we perform rank-value encoding to generate the inputs of the model. For each cell $n$, we sort the normalized expression values $x''_{n,m}$ in descending order.
The resulting input is a sequence of gene tokens $S_n = (m_{(1)}, m_{(2)}, \dots, m_{(L)})$, where each token $m_{(k)} \in \mathcal{M}$ is the gene corresponding to the $k$-th highest expression value in cell $n$. We truncate this gene sequence to a maximum length of $L=2,000$, and then append the \texttt{[CLS]} and \texttt{[EOS]} tokens at the beginning and the end of the sequence, respectively. By defining the sequence elements directly as members of $\mathcal{M}$, we ensure that the input tokens map one-to-one with the pre-defined model vocabulary. 

{\bf LCA Distance Matrix Construction}
To quantify biological similarity, we leverage the hierarchical structure of the cell ontology graph $\mathcal{G} = (\mathcal{V}, \mathcal{E})$ from the Open Biological and Biomedical Ontology (OBO) Foundry \cite{cellOntology-obo-foundry2007}, where $\mathcal{V}$ denotes cell-type nodes in the graph and every cell label $c_i$ in our dataset is an instance of a node $v \in \mathcal{V}$; $\mathcal{E}$ is the directed edge between two nodes, encoding the hierarchical relationship (``is a subtype of'') between cell types. 
We construct a pairwise distance matrix $\Delta \in \mathbb{R}^{|\mathcal{V}| \times |\mathcal{V}|}$ derived from the Lowest Common Ancestor (LCA) based on $\mathcal{G}$ \cite{LCA-distance-Aho1973}. 
The general concept of this process is shown in Figure~\ref{fig:main_fig}(b). For any two types of cells $c_i, c_j \in \mathcal{V}$, their raw semantic distance $d_{\mathcal{G}}(c_i, c_j)$ is calculated as the sum of the shortest path distances from each node to their common $c_s = \text{LCA}(c_i, c_j)$:
\begin{align}
\Delta_{i,j} = \delta_{\mathcal{G}}(c_i, c_j) = \delta_{path}(c_i, c_s) + \delta_{path}(c_j, c_s) ,
\label{eq:LCA_distance}
\end{align}
where $\delta_{path}(c_i, c_s)$ and $\delta_{path}(c_j, c_s)$ denote the number of edges on the shortest path between nodes $(c_i, c_j)$ to their lowest common ancestor $c_s$ in $\mathcal{G}$, respectively. To facilitate stable training with the dynamic triplet margin loss in Section~\ref{sec:objectives}, we compute the normalized distance $\tilde{\delta}(c_i, c_j)$ element wise and obtain $\tilde{\Delta}_{i,j}$, by scaling $\delta_{\mathcal{G}}(c_i, c_j)$ by the maximum distance on the ontology graph. The matrix $\tilde{\Delta}$, with elements $\tilde{\Delta}_{i,j}$, serves as the ground truth for biological distinctness during the extraction of negative pairs in Section~\ref{sec:offline_sampling} and to determine the dynamic margin for the triplet loss defined in Section~\ref{sec:objectives}. We provide more empirical evidence about the relationship between LCA-based distance and transcriptomic similarity in Section \ref{sec:results} and Appendix \ref{suppl:lca_vs_expr}.

\subsection{Offline Triplet Sampling}
\label{sec:offline_sampling}
Contrastive learning \cite{chen2020simple} is a representation learning framework that aims to learn an embedding space such that it encourages semantically similar samples to be closer while pushing dissimilar samples apart.
In contrastive learning, an {\em anchor sample} serves as a reference that is used to compare, by proxy, positive and negative samples.
The general concept is shown in Figure~\ref{fig:main_fig}(c).
Efficiently mining of informative triplets is critical for contrastive learning.
While standard online triplet sampling, like in SCimilarity and scCello, is highly inefficient at massive scale, since random mini-batches are unlikely to contain sufficiently informative positive pairs. Here, we propose a hierarchical indexing structure, which we term the {\em Cell-Dataset Index Map} ($\Omega$), to enable $O(1)$ retrieval of candidate samples.
Let $\mathcal{V}$ be the set of cell types and $\mathcal{K}$ be the set of all dataset identifiers ({\em e.g.}, studies). We define $\Omega$ as the collection of index sets $\mathcal{I}_{c,k}$, each of which contains the indices of cells belonging to type $c \in \mathcal{V}$ and dataset $k \in \mathcal{K}$: $\Omega = 
\{ \mathcal{I}_{c,k} 
\}_{c\in{\cal V},k\in{\cal K}}$,
%
%
where again $\mathcal{I}_{c,k}$ denotes the set of indices for all cells of type $c$ found in dataset $k$.
Additionally, for any cell type $c$, we denote the set of datasets containing this cell type as $\mathcal{K}_c = \{ k \in \mathcal{K} \mid \mathcal{I}_{c,k} \neq \emptyset \}$. For each training step, we select an anchor cell $x_a$ with cell type $c_a$ from dataset $k_a$.
Specifically, we construct triplets $(x_a, x_p, x_n)$ as follows.

{\bf Positive Sampling}
To promote batch mixing for the same cell type, we prioritize sampling {\em cross-dataset} positive pairs from different datasets.
We define a pool of positive candidates, $\mathcal{P}^+$, designed to prioritize cross-dataset pairs while also handling edge cases where a cell type is unique to a single dataset.
If the cell type $c_a$ exists in multiple datasets ($|\mathcal{K}_{c_a}| > 1$), we construct the pool by taking the union of all instances from the datasets in ${\cal K}_{c_a}$ other than $k_a$: $\mathcal{P}^+ = \bigcup_{k \in \mathcal{K}_{c_a} \setminus {k_a}} \mathcal{I}_{c_a, k}$, and then we uniformly sample $x_p \sim \text{Uniform}(\mathcal{P}^+)$. If the cell type only exists in the anchor's dataset ({\em i.e.}, $\mathcal{P}^+ = \emptyset$), we default to sampling $x_p$ from the anchor's own dataset index $\mathcal{I}_{c_a, k_a}$.

{\bf Negative Sampling} 
Randomly selected negative samples ({\em e.g.}, ``T cell'' {\em vs.} ``Neuron'') are often too easy to distinguish, producing zero loss and vanishing gradients.
To ensure robust learning, we specifically mine {``semi-hard'' to ``hard'' negatives}—cell types that are biologically distinct but semantically close to the anchor.

We utilize the Cell Ontology graph to measure semantic proximity. Recall that $\Delta$ is the pre-computed LCA distance matrix between all cell types. For an anchor with cell-type $c_a$, we first define the set of \textit{valid negative cell types}, denoted as $\mathcal{V}_{\text{n}}$, excluding the anchor type $c_a$ and its lineage (ancestors and descendants) to avoid conflicting training signals (ambiguous pairs) during training: $\mathcal{V}_{\text{n}} = \{ v \in \mathcal{V} \mid v \neq c_a \land v \notin \text{Lineage}(c_a) \}$.
%
%
To identify the most informative negatives, we rank all valid types based on their proximity to the anchor. Let $\text{rank}(c_a, v)$ denote the ordinal rank of the anchor cell type $c_a$ and all other valid negative cell types $v \in \mathcal{V}_{\text{n}}$ when sorted by ascending LCA distance $\Delta(c_a, v)$. We select the set of {\em hard to semi-hard negative candidates} by following:
$\mathcal{C}_{\text{hard}} = \{ v \in \mathcal{V}_{\text{n}} \mid \text{rank}(c_a, v) \le M \}$, and we set $M=50$ by default, which provides a diverse set of closely-related cell types to sample from, stabilizing training while maintaining optimization ``difficulty''.
By discriminating against these biologically similar types, the model learns to capture high-resolution lineage markers rather than just broad lineage differences.
Then the negative cell instance $x_n$ is uniformly sampled from:
$x_n \sim \bigcup_{c \in \mathcal{C_{\text{hard}}}} \mathcal{I}_{c, k}$.

Since the collection of index sets $\Omega$ 
pre-groups indices, we avoid costly filtering operations during training.
The sampling logic described above is implemented as a vectorized look-up operation.
We pre-compute the valid positive dataset indices and negative cell type candidates for every batch, allowing us to generate millions of triplets in parallel using efficient tensor gathering operations on the GPU. To support this empirically, the computational cost of constructing the {\em Cell-Dataset Index Map} is summarized in ATable \ref{tab:cell_dataset_index_map_cost}.
%


\subsection{Model Architecture}
\label{sec:model_architecture}
We adopt a transformer architecture similar to the Geneformer \cite{Geneformer-Theodoris2023}. The model consists of 6 transformer layers with 256 hidden units and 4 attention heads. The input to the model is the rank-ordered sequence of gene tokens $S_n$, which is embedded as follows: $\mathbf{H}_0 = \text{E}(S_n) + \mathbf{E}_{pos}$, 
%
%
where $\text{E}(\cdot)$ is the gene-token embedding layer, $\mathbf{E}_{pos}$ represents the learnable positional embedding, and the embedding $\mathbf{H}_0$ is the embedded input used by the model.
A special \texttt{[CLS]} token is added to the beginning of the sequence to aggregate global cell representations, and \texttt{[EOS]} is added to the end of the input gene sequence $S_n$. We use the output of the final hidden layer of the transformer model associated with the \texttt{[CLS]} token, denoted as $\mathbf{z}_n$, as the low-dimensional cell embedding.

\subsection{Learning Objective}
\label{sec:objectives}
We optimize our model using three objectives, namely $i$) the standard MLM term, $ii$) the dynamic margin triplet loss, and $iii$) the global ontology alignment loss. The latter two loss objectives leverage the LCA distance matrix to guide the model to learn biologically meaningful representations. Compared with existing methods such as scCello \cite{scCello-yuan2024cell}, which employs a fixed scaling parameter to learn ontology-informed pairwise relational alignment in its contrastive loss and focuses on local cell-type relationships, our approach uses a dynamic margin triplet loss to enable finer-grained discrimination of cell states, while the rank-order alignment loss further preserves the global hierarchical structure in the embedding space.

{\bf Masked Language Modeling (MLM)} Like in BERT \cite{BERT-devlin-etal-2019}, we apply the MLM strategy with our input sequence $S_n$,  and the masking ratio is set to $15\%$. For the selected masking position, we apply the same masking rule as used in BERT, ($i$) there is 80\% chance that the token is replaced by the \texttt{[MASK]} token; ($ii$) 10\% probability that the token will be replaced by a random token; and ($iii$) 10\% chance that the token will remain unchanged. Let $\mathcal{J}_{mask} \subset {1, \dots, L}$, denote the set of masked position indices. We write the MLM loss as follows
\begin{equation}
\label{eq:MLM_loss}
\mathcal{L}_{\rm MLM} = -\textstyle\sum_{j \in \mathcal{J}_{mask}} \log P(t_j | S_{n \setminus \mathcal{J}_{mask}}) ,
\end{equation}
where $t_j$ is the original token in the masked position and $S_{n \setminus \mathcal{J}_{mask}}$ represents the masked sequence. This objective enables the model to learn gene–gene co-expression patterns that can be used to reconstruct missing gene expression values as previously described in \cite{Geneformer-Theodoris2023}. 

{\bf Dynamic Ontology-Aware Triplet Loss}
To enforce the biological hierarchy in the embedding space, we employ a triplet loss with a dynamic margin $\alpha(\cdot,\cdot)$ (Figure~\ref{fig:main_fig}(c) top panel). 
Specifically, $\alpha(\cdot,\cdot)$ adapts to the specific semantic relationship between the anchor and the negative sample based on their positions in the cell-type ontology instead of setting a fixed margin as in the original triplet loss \cite{hierarchical-tripletLoss-Ge_2018_ECCV}. Specifically, we define the dynamic margin $\alpha(c_{a}, c_{n})$ as:
\begin{equation}
\alpha(c_{a}, c_{n}) = \epsilon + (1 - \tilde{\delta}(c_{a}, c_{n})) + \text{depth}(c_{a}) ,
\end{equation}
where $\epsilon$ is a small constant offset (set to 0.1 by default), $\tilde{\delta}$ denotes the normalized LCA distance between cell types $c_a$ and $c_n$, and $\text{depth}(c_{a})$ is the normalized depth of the anchor node in the ontology graph. This design encourages a larger separation margin for semantically similar cell types, {\em i.e.}, small $\tilde{\delta}$, leading to larger margin, and for anchors corresponding to fine-grained cell types, {\em i.e.}, large $\text{depth}$, as well. The offset term $\epsilon$ ensures a minimum separation for shallow and  biologically distinct nodes, in which these cases are easily distinguishable by the model.
The resulting triplet loss function is formulated as follows.
\begin{equation}
\label{eq:Triplet_loss}
\mathcal{L}_{\rm Tri} = \max\left(0, \lVert \mathbf{z}_{a} - \mathbf{z}_{p} \rVert_2^2 - \lVert \mathbf{z}_{a} - \mathbf{z}_{n} \rVert_2^2 + \alpha(c_{a}, c_{n}) \right) .
\end{equation}
Consequently, this design forces the model to learn more discriminative representations that capture subtle biological distinctions, {\em i.e.}, CD$4^+$ T cells and CD$8^+$ T cells, relative to naive T cells.

{\bf Global Cell-type Ontology Alignment} 
Although the dynamic margin triplet loss effectively enforces {\em local discrimination}—ensuring that fine-grained, semantically similar subtypes are pushed apart to prevent collapse—, however, it does not inherently preserve the broader cell-type hierarchical lineage in a mini-batch.
Here, we introduce a complementary objective that aligns the geometry of the embedding space with the global hierarchical relationships encoded by the cell. Concretely, we propose a \textit{Rank-Order Alignment Loss} that encourages consistency between ($i$) pairwise distances in the learned embedding space, and ($ii$) reference ontological distances, by matching their relative (rank) order (see Figure~\ref{fig:main_fig}(c) lower panel).

Let $\mathbf{Z} \in \mathbb{R}^{B \times d}$ denote cell embeddings in a mini-batch, where $B$ is the batch size and $d$ is the embedding dimension, which we set to $d=256$ in our case.
We firstly compute the pairwise cosine distance matrix in the embedding space in the current batch:
$
\Delta^{\text{emb}} = 1 - \frac{\mathbf{Z}\mathbf{Z}^\top}{\|\mathbf{Z}\|_2 \|\mathbf{Z}^\top\|_2} .
$
Following this, we get the embedding distance matrix $\Delta^{\text{emb}} \in \mathbb{R}^{B \times B}$ in the batch.
Regarding the reference ontology distances, recall that $\tilde{\Delta}$ is the normalized ontology LCA distance, here we denote $\tilde{\Delta}^{\text{onto}} \in \mathbb{R}^{B \times B}$ as the corresponding LCA ontology distance sub-matrix in the current batch. 
Let $\Delta_n^{\text{emb}}$ and $\tilde{\Delta}_n^{\text{onto}}$ be the vectors obtained from the 
$n$-th row of $\Delta^{\text{emb}}$, and $\tilde{\Delta}^{\text{onto}}$, respectively.
Our goal is to make the ranking of neighbors induced by $\Delta_n^{\text{emb}}$ match the ranking induced by $\tilde{\Delta}_n^{\text{onto}}$, {\em i.e.}, to maximize the row-wise Spearman rank correlation between the two distance vectors. Since the Spearman correlation involves a non-differentiable sorting/ranking operation, we adopt the differentiable soft-ranking operator~\cite{softRanking-blondel2020fast}. Specifically, we compute soft rank vectors: 
\[
\mathbf{r}_n^{\text{emb}} = \mathcal{R}_{\text{soft}}(\Delta_n^{\text{emb}}), \quad \mathbf{r}_n^{\text{onto}} = \mathcal{R}_{\text{soft}}(\tilde{\Delta}_n^{\text{onto}}) ,
\]
where $\mathcal{R}_{\text{soft}}(\cdot)$ denotes the differentiable ranking operator. We then define the global ontology alignment objective as follows.
\begin{equation}
\label{eq:OntologyAlign_loss}
\mathcal{L}_{\text{ontoAlign}} = 1 - \frac{1}{B} \textstyle\sum_{i=n}^{B} \rho_S(\mathbf{r}_n^{\text{emb}}, \mathbf{r}_n^{\text{onto}}) ,
\end{equation}
where $\rho_S(\cdot, \cdot)$ is the Spearman rank correlation. Minimizing $\mathcal{L}_{\text{ontoAlign}}$ encourages the learned embedding space to preserve, for each cell, the relative ordering of ontological distances within each batch, thus aligning the global embedding structure with the hierarchical relationships defined by the ontology.

{\bf Complete Objective}
The complete loss function is a weighted sum of the three terms in (\ref{eq:MLM_loss}), (\ref{eq:Triplet_loss}) and (\ref{eq:OntologyAlign_loss}): $\mathcal{L}_{Total} = \mathcal{L}_{\text{MLM}} + \lambda_1 \mathcal{L}_{\text{Tri}} +  \lambda_2 \mathcal{L}_{\text{ontoAlign}}$, 
%
%
where $\lambda_1$ and $\lambda_2$  control the contribution of local and global ontology structure preserving losses, respectively. For simplicity, we set $\lambda_1=\lambda_2=1$ in all of our experiments. The dynamic triplet margin loss enables us to discriminate subtle cell-state differences, while the ontology alignment loss imposes a global hierarchical structure on the learned representation to preserve biological fidelity. The trade-off and synergy between these complementary objectives are explicitly analyzed in the ablation study in Section ~\ref{sec:results}.
%

\section{Experiments}
\label{sec:experiments}
\subsection{Pre-Training}
{\bf Pretrained Dataset Construction} We downloaded 234 \textit{Homo sapiens} scRNA-seq datasets from the CZI \cite{CZI_2023} data portal (2025-01-30 release), covering 210 tissues, 566 cell types, 58 disease types in the pretraining phase (see Figure \ref{fig:main_fig}(a)). All datasets are from \textit{Homo Sapiens} and we retained only those sequencing using the 10X platform. We filtered out low-quality cells with fewer than 200 detected genes and genes expressed in fewer than 5 cells, and excluded malignant cells, non-primary samples, and cells whose annotated cell types are not present in the Cell Ontology graph. After such filtering, approximately 27.5 million cells remained.

{\bf Pretrained Model Architecture and Configurations} We adopt the same BERT-style architecture as scCello \cite{scCello-yuan2024cell} and Geneformer \cite{Geneformer-Theodoris2023}, with six Transformer layers and four attention heads per layer, a 512-dimensional intermediate layer and a 256-dimensional hidden size. We pretrain the model using AdamW (learning rate 1e-3, weight decay 0.001) with a warm-up of approximately 1\% of the total training steps. Training is performed for three epochs on six NVIDIA H200 GPUs with a batch size of 64 per GPU and a gradient accumulation step of 4 (effective batch size = 64 × number of GPUs × 4); the complete training process takes approximately 30 hours. Additional training details and hyperparameter settings are provided in Appendix~\ref{supp:model_config}.

\subsection{Downstream Task Settings}
In the following, we describe the state-of-the-art (SOTA) baselines and provide details on the datasets and evaluation metrics used for each downstream task. Additional information on the baselines and evaluation metrics is provided in Appendix~\ref{suppl:Baselines} and Appendix~\ref{suppl:evaluation_metrics}.

{\bf Benchmark baselines}
We benchmark our model against five SOTA scFMs, namely, Geneformer \cite{Geneformer-Theodoris2023}, scGPT \cite{scGPT-Cui2024}, scCello \cite{scCello-yuan2024cell}, SCimilarity\cite{SCimilarity-Heimberg2024} and CellPLM \cite{cellPLM-wen2024cellplm}, on four downstream tasks spanning both cell- and gene-level evaluations. 
The illustration of the downstream tasks considered is shown in Figure~\ref{fig:main_fig}(d).

{\bf Cell-type Annotation}
To benchmark zero-shot cell-type clustering, we evaluate on nine datasets excluded from pretraining, spanning blood \cite{HIHA-Gong2025}, heart \cite{Heart-Kanemaru2023}, liver \cite{Liver-Edgar2025}, kidney \cite{kidney-Lake2025}, pancreas \cite{Pancreas-Bandesh2025}, lung \cite{Lung-Kim2020}, breast \cite{Breast-Gray2022}, adipose tissue \cite{Adipose-tissue-Emont2022}, and bone marrow \cite{BoneMarrow-Zeng2025}, with sizes ranging from 30k to 348k cells (Appendix~\ref{suppl:dataset-zero-shot}). Specifically, the HIHA (blood), kidney and Azimuth PBMC \cite{Azimuth-Hao2021} datasets are selected for model robustness analysis in cell-type annotation, as they provide author-annotated labels spanning multiple levels of granularity, from coarse to fine. We further construct an evaluation set of 340 overlapping cell types from 212 unseen datasets (ATable~\ref{tab:unseen_datasetid}).
Performance is assessed using scIB metrics \cite{scIB-Luecken2021}, including ARI, NMI, and ASW$_{label}$, and their arithmetic mean (AvgBio), along with ASW$_{label/batch}$ and GraphConn, and their mean (AvgBatch). The overall scIB score is calculated as a weighted average of these two aggregated metrics (0.6$\times$AvgBio + 0.4$\times$AvgBatch). We also report scGraph-RWR and scGraph-LCA \cite{scGraph-Wang2025, Wu2025-my} to evaluate the preservation of biological structure captured by model embeddings. For cell-type annotation fine-tuning (Appendix \ref{suppl:cell-anno-finetune}), we use the pancreas and HIHA datasets, reporting accuracy and macro-F1 for the predicted labels. Moreover, we compare with hierarchical cross-entropy (HCE) classifiers \cite{HCE-2026}, using the authors' released MLP and TabNet checkpoints.

{\bf Novel Cell Identification}
We use the same 212 unseen datasets as previously mentioned in the cell-type annotation task. However, here we focus on cell types that are absent from pretraining. After filtering rare types (<2,000 cells) and subsampling abundant ones (>10,000 cells), the benchmark contains 99 novel cell types and 774,215 cells (ATable~\ref{tab:novelCL_full_table}, Appendix~\ref{suppl:novel_cell}).
Performance is measured using per-cell-type accuracy and macro-accuracy to account for class imbalance.

{\bf Gene Function Prediction}
We evaluate both static and contextual gene embeddings. For static embeddings, Gene Ontology (GO) gene sets from MSigDB \cite{MSigDB-Liberzon2011} are grouped into Biological Process (BP), Cellular Component (CC), and Molecular Function (MF), and evaluated separately. For contextual embeddings, we conduct a case study on lung adenocarcinoma \cite{Lung-Kim2020} using \textit{decoupleR} \cite{decoupleR-BadiaiMompel2022}. Details are provided in Appendix~\ref{suppl:gene_func_pred}.
We report micro/macro precision and micro/macro F1 scores.

{\bf Cross-modal Spatial Gene Imputation}
We evaluate spatial gene imputation using scRNA-seq references from lung cancer \cite{Lung-Kim2020}, breast cancer \cite{breastCancer-NG-Wu2021}, and brain tissue from SEA-AD \cite{SEAAD-Gabitto2024}, paired with corresponding spatial datasets. Additional details are summarized in Appendix~\ref{suppl:spatial_gene_imputation}.
Performance is quantified using Pearson, Spearman, and Kendall’s $\tau$ correlations between imputed and ground-truth spatial expression \cite{scikit-learn}.
\vspace{-2mm}
\section{Results}
\label{sec:results}
We first examine the relationship between cell ontology distance and transcriptomic distance to assess the biological motivation for ontology-guided learning. We then evaluate scOntoFM on cell-level tasks, including cell-type annotation and novel cell-type identification, followed by gene-level tasks, namely, gene function prediction and spatial gene imputation. Finally, we analyze the contributions of the training objectives through ablation studies. For all tables reported in the following, \textbf{bold} indicates the best performance and \underline{underlining} denotes the second-best results.

\begin{figure*}[t]
  \centering

  \begin{minipage}[t]{0.48\textwidth}
    \centering
    \vspace{0pt}
    \includegraphics[width=5cm]{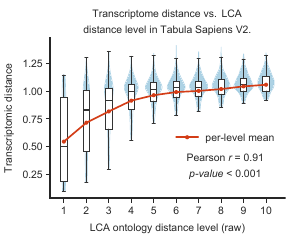}
    \vspace{-2mm}
    \captionsetup{type=figure,font=small}
    \captionof{figure}{Cell-type pairwise transcriptomic distance in PCA space across ten LCA ontology levels in Tabula Sapiens v2. Boxes show the distance distributions, and the \textcolor{red}{red line} denotes the per-level mean.}
    \label{fig:lca_level_expr_boxplot}
  \end{minipage}
  \hfill
  \begin{minipage}[t]{0.48\textwidth}
    \centering
    \vspace{-2mm}
    \captionsetup{type=figure,font=small}
    \captionof{table}{Correlation between LCA distance and cell-type centroids cosine distance in PCA space over the nine zero-shot clustering datasets. ``Overall'' pools pairs across all the datasets.}
    \label{tab:lca_nine}
    \small
    \setlength{\tabcolsep}{5pt}
    \renewcommand{\arraystretch}{1.05}
    \scalebox{0.85}{
    \begin{tabular}{lccc}
      \toprule
      \textbf{Dataset} & \textbf{Pearson $r$} & \textbf{Spearman $\rho$} & \textbf{\# Cell types} \\
      \midrule
      HIHA           & 0.58 & 0.60 & 38 \\
      Heart          & 0.47 & 0.43 & 12 \\
      Lung           & 0.51 & 0.55 & 10 \\
      Liver          & 0.37 & 0.29 & 18 \\
      Kidney         & 0.31 & 0.27 & 33 \\
      Pancreas       & 0.38 & 0.30 & 12 \\
      Breast TNBC    & 0.54 & 0.31 & 6 \\
      Adipose tissue & 0.35 & 0.28 & 16 \\
      Bone marrow    & 0.39 & 0.36 & 40 \\
      \midrule
      Overall        & 0.41 & 0.40 & 130 \\
      \bottomrule
    \end{tabular}
    }
  \end{minipage}
  \vspace{-4mm}
\end{figure*}

{\bf Relationship Between Ontology and Transcriptomic Distances} To examine the relationship between ontology proximity and transcriptional similarity, we compared LCA distances with the pairwise cosine distances between cell-type centroids in PCA space, using Tabula Sapiens v2 and the nine datasets used for zero-shot clustering. As shown in Figure \ref{fig:lca_level_expr_boxplot}, the mean transcriptomic distance between cell types increased monotonically over the first ten LCA distance levels (Pearson $r=0.91$, Spearman $\rho=1$), then plateaued (AFigure \ref{fig:lca_level_expr_boxplot_all_levels}). Consistent with this trend, we observed positive correlations across all nine zero-shot clustering datasets, with Pearson $r$ ranging from 0.31 to 0.58 and Spearman $\rho$ from 0.27 to 0.60 (Table \ref{tab:lca_nine}). Together, these results indicate that ontology-near cell types tend, on average, to be more transcriptionally similar, supporting LCA distance as an informative structural prior, while its limitations and more details are discussed in Appendix \ref{suppl:lca_vs_expr}.

%

\begin{figure}[t]
  \centering
  \begin{minipage}[t]{0.485\columnwidth}
    \includegraphics[width=\linewidth, trim=0 0 5 0, clip]{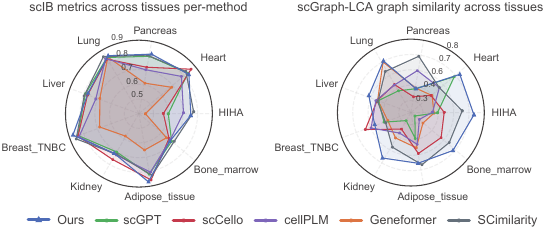}
    \captionsetup{type=figure,font=small}
    \vspace{-2mm}
    \caption{Radar plots of zero-shot clustering benchmark across nine tissues in scIB (left) and scGraph-LCA (Spearman correlation) (right) metrics across six SOTA methods.}
    \label{fig:zero_shot_clustering}
  \end{minipage}
  \hspace{0.01\columnwidth}
  \begin{minipage}[t]{0.485\columnwidth}
    \includegraphics[width=\linewidth, trim=2 1 1 1, clip]{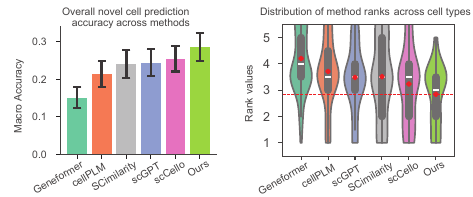}
    \captionsetup{type=figure,font=small}
    \vspace{-2mm}
    \caption{Bar plot of macro-accuracy for novel cell identification (left) and violin plot of per-cell type accuracy rank distributions (right); $\color{red}\blackdiamond$ denotes each method’s average rank.}
    \label{fig:novelCL}
  \end{minipage}
  \vspace{-6mm}
\end{figure}

{\bf Cell Type Annotation (Zero-shot Clustering)} We first evaluate the performance of our model with SOTA methods across nine  datasets of diverse tissues.
As shown in the radar plot of Figure~\ref{fig:zero_shot_clustering} and ATables~\ref{tab:scIB_overall_tissue_performance} and \ref{tab:scIB_full_benchmark_results}, scOntoFM achieves a mean scIB score comparable to SCimilarity (0.7958 {\em vs.} 0.7960) and above the remaining baselines.
We then investigate performance using biologically structure preserving metrics (scGraph-LCA and scGraph-RWR) in Figure \ref{fig:zero_shot_clustering} right panel, ATables~\ref{tab:scGraph-LCA_metrics} and \ref{tab:scGraph-RWR_performance}. Notably, our model achieves substantially higher performance than SOTA methods on these two metrics across most tissues, highlighting the complementary nature of biological conservation {\em versus} scIB metrics. 
To further illustrate this, we examined the cell-type embedding distance heatmap of Bone Marrow tissue in AFigure~\ref{fig:BoneMarrow_emb_heatmap}(c), which aligns well with the LCA-based ontology distance heatmap as well as the RWR-based ontology distance heatmap (AFigure~\ref{fig:scGraph_reference_heatmap}).
Specifically, we investigate the hematopoietic stem cell and mark it as the rooted cell type in the distance umap plot (see AFigure~\ref{fig:BoneMarrow_emb_heatmap}(a,b,d)), which precisely illustrates the developmental lineage of hematopoietic and myeloid-related cell types, as well as aligns well with G2M and S phase cell cycle scores reported in the original study \cite{BoneMarrow-Zeng2025}.

We further evaluate the robustness of our model embeddings (Table \ref{tab:hiha_kidney_robust_labels}) under varying annotation granularity in HIHA and Kidney dataset, which contain author annotations from coarse to fine-grained labels. We calculate the agreement scores (ARI) (ATable \ref{tab:agreement_scores}) and visualize the cell-type annotation flow in sankey plot (AFigure \ref{fig:HIHA_sankey_plot}, \ref{fig:Kidney_sankey_plot})) between the cell ontology reference and different annotation levels. Leveraging its ontology-aware design, scOntoFM achieves competitive clustering across varying resolutions, particularly at finer annotation levels ({\em e.g.}, AIFI\_L2/L3 and Subclass\_L2), indicating its superior ability to capture subtle cellular states. Although performance is comparatively lower on coarse-grained annotations ({\em e.g.}, AIFI\_L1 and Subclass\_L1), this is expected due to their weaker alignment with the fine-grained Cell Ontology labels used during pretraining.

\begin{table*}[t]
\centering
\setlength{\tabcolsep}{3pt}
\renewcommand{\arraystretch}{0.95}
\begin{minipage}[t]{0.5\textwidth}
\centering
\captionsetup{type=table,font=small}
\caption{ Evaluation of zero-shot clustering robustness under varying annotation levels using the scIB score in HIHA and Kidney datasets.
}
\label{tab:hiha_kidney_robust_labels}
\scriptsize
\vspace{-2mm}
\scalebox{0.9}{
\begin{tabular}{lccccc}
\toprule
& \multicolumn{3}{c}{\textbf{HIHA}} & \multicolumn{2}{c}{\textbf{Kidney}} \\
\cmidrule(lr){2-4} \cmidrule(lr){5-6}
\textbf{Model} 
& \textbf{AIFI\_L1} 
& \textbf{AIFI\_L2} 
& \textbf{AIFI\_L3} 
& \textbf{Subclass\_L1} 
& \textbf{Subclass\_L2} \\
\midrule
Ours 
& 0.681 & \textbf{0.782} & \textbf{0.671} & \underline{0.740} & \textbf{0.648} \\

Geneformer 
& 0.510 & 0.473 & 0.398 & 0.585 & 0.520 \\

scGPT 
& 0.670 & 0.622 & 0.563 & 0.713 & 0.631 \\

scCello 
& \underline{0.701} & 0.600 & 0.506 & \textbf{0.770} & 0.639 \\

cellPLM 
& \textbf{0.760} & \underline{0.722} & \underline{0.666} & 0.721 & \underline{0.640} \\
\bottomrule
\end{tabular}
}
\end{minipage}
\hspace{0.01\columnwidth}
\begin{minipage}[t]{0.47\textwidth}
\centering
\captionsetup{type=table,font=small}
\caption{ Cell-type annotation benchmark (fine-tuning) on HIHA and Pancreas datasets across six SOTA methods.
}
\label{tab:cellAnno_finetune}
\vspace{-2mm}
\scalebox{0.65}{
\begin{tabular}{lcccc}
\toprule
& \multicolumn{2}{c}{\textbf{HIHA}} & \multicolumn{2}{c}{\textbf{Pancreas}} \\
\cmidrule(lr){2-3} \cmidrule(lr){4-5}
\textbf{Model} & \textbf{Accuracy} & \textbf{Macro F1} & \textbf{Accuracy} & \textbf{Macro F1} \\
\midrule
Ours
& \textbf{0.915} $\pm$ 0.00
& \underline{0.771} $\pm$ 0.01
& \textbf{0.992} $\pm$ 0.00
& \textbf{0.969} $\pm$ 0.01 \\

Geneformer
& 0.701 $\pm$ 0.01
& 0.395 $\pm$ 0.01
& 0.935 $\pm$ 0.00
& 0.840 $\pm$ 0.01 \\

scGPT
& 0.839 $\pm$ 0.00
& 0.613 $\pm$ 0.02
& 0.973 $\pm$ 0.00
& 0.919 $\pm$ 0.01 \\

scCello
& 0.811 $\pm$ 0.01
& 0.543 $\pm$ 0.01
& 0.977 $\pm$ 0.00
& \underline{0.960} $\pm$ 0.00 \\

cellPLM
& 0.907 $\pm$ 0.00
& \textbf{0.792} $\pm$ 0.01
& 0.968 $\pm$ 0.00
& 0.902 $\pm$ 0.01 \\

HCE-TabNet
& \underline{0.910} $\pm$ 0.00
& 0.710 $\pm$ 0.02
& \underline{0.980} $\pm$ 0.00
& 0.900 $\pm$ 0.01 \\
\bottomrule
\end{tabular}
}
\end{minipage}
\vspace{-3mm}
\end{table*}

To assess the generalization of our model in real world scenarios, we evaluated 340 cell types present in the pre-training cell-type labels but sourced from entirely new, unseen datasets.
Although our method shows comparable performance to cellPLM in standard scIB metrics (ATable \ref{tab:seenCL_new_scib_scgraphlca_multicol}), it outperforms cellPLM by a significant margin in scGraph-LCA.
Collectively, these results demonstrate that our embeddings better preserve cell-type ontology structure and exhibit superior biological conservation, a critical aspect often under-emphasized in prior foundation model evaluation.

{\bf Cell Type Annotation (Fine-tuning)} We conduct cell type classification benchmarks on the HIHA and Pancreas datasets, with fixed held-out donors and identical hyperparameters for all the  and HCE classifiers. Averaged over five random seeds, scOntoFM achieves the best overall accuracy and maintains strong Macro-F1 scores (Table \ref{tab:cellAnno_finetune}), indicating its effectiveness in learning discriminative and generalizable representations for accurate cell type classification beyond zero-shot clustering. To further evaluate our model robustness to label noise, we progressively inject noise into the training annotations of the well-annotated Azimuth PBMC dataset by randomly replacing fine-grained labels with their corresponding coarser categories. scOntoFM demonstrates highly stable and accurate cell annotation even  under increasing noise levels (ATable \ref{tab:label_noise_analysis}). When the noise ratio reaches 50\%, the model reliably falls back to predicting broader parent categories. This behavior is consistent with biological hierarchies, ensuring structurally coherent predictions despite unreliable fine-grained annotations.

{\bf Novel Cell Identification} To evaluate the hierarchical representations of known cell types and their generalizability to unseen cell types, we consider a more challenging setting: \emph{novel cell-type identification}.
We leverage the full cell-ontology LCA distance matrix to infer embeddings for unseen cell-type centroids via a weighted average of seen centroid (see Appendix~\ref{suppl:novel_cell} for details).
First, we examine the biological structure coherence of \emph{seen} cell types.
As visualized in AFigure~\ref{fig:coarse_seenEmb_umap_plot}(a, b), the centroid embeddings exhibit distinct separation among major lineages, including endothelial cells, leukocytes, myeloid cells, and hematopoietic oligopotent progenitors, confirming that the model captures biologically meaningful embeddings.
Quantitatively, as shown in Figure~\ref{fig:novelCL}, our method outperforms all SOTA baselines, achieving an overall macro-accuracy of 28.59\% across 99 novel cell types (ATable \ref{tab:novelCL_full_table}). And the per-cell-type rank distribution reveals that our model attains the lowest average rank (2.85) across the 99 unseen types—an improvement of 13.65\% over the second-best method, scCello (3.24).
This underscores the model's superior capability in recognizing previously unseen biological states.


To thoroughly examine this ability, we selected three unseen cell types from the hematopoietic lineage where scOntoFM achieved high predictive accuracy: megakaryocyte progenitor cells, immature neutrophils, and CD14-positive/CD16-low monocytes (AFigure~\ref{fig:coarse_seenEmb_umap_plot}(c, d)).
We retrieved the top 5 closest \emph{seen} cell types for each of these unseen targets from the pretraining dataset and performed co-clustering with the unseen cell types.
As shown in AFigure~\ref{fig:case_LCA_cluster_plot}(a, b, c), the unseen cell types cluster proximally to their biologically related seen counterparts in the embedding space.
Crucially, this embedding proximity is consistent with the LCA distance heatmap (AFigure~\ref{fig:case_LCA_cluster_plot}(d)), which validates that the model successfully extrapolates the correct ontological neighborhood for novel cells.

{\bf Gene Function Prediction} scFMs inherently learn informative gene embeddings from diverse cellular contexts and disease conditions. In this section, we systematically explore the biological knowledge encoded by these embeddings; specifically, we compare the \textit{static} gene representations learned by scFMs to make functional predictions about GO terms, namely, BP, CC, and MF. As shown in Table~\ref{tab:geneFunc_performance}, our method achieves the highest classification performance for all baselines, achieving top scores in both Micro-F1 (0.917±0.001) and Macro-F1 (0.916±0.001) using 20-fold cross validation.
This indicates that our model offers a superior balance of precision and recall when assigning functional labels. Furthermore, in terms of Average Precision (AP), our method remains highly competitive, performing on par with Geneformer and scGPT ({\em e.g.}, Micro AP of 0.921 {\em vs.} 0.922) and outperforming scCello. These results demonstrate that our biologically informed training strategy enables the model to capture robust, semantically meaningful gene representations that generalize well to standard functional annotation tasks. We also show GO namespace classification results by aggregating the corresponding gene embeddings for each GO term in ATable \ref{tab:aggregated_GO_embedding}.

\begin{table*}[t]
\centering
\setlength{\tabcolsep}{3pt}
\renewcommand{\arraystretch}{0.95}
\begin{minipage}[t]{0.46\textwidth}
\centering
\captionsetup{type=table,font=small}
\caption{Gene function prediction performance across SOTA methods. Classification accuracy (mean $\pm$ std) is calculated over 20-fold cross validation, best values are highlighted in \textbf{bold}.}
\label{tab:geneFunc_performance}
\scriptsize
\vspace{-1mm}
\resizebox{\linewidth}{!}{%
\begin{tabular}{lcccc}
\toprule
Method & F1 Micro & F1 Macro & AP Micro & AP Macro \\
\midrule
Ours & $\mathbf{0.917 \pm 0.001}$ & $\mathbf{0.916 \pm 0.001}$ & $\underline{0.921} \pm 0.002$ & $\underline{0.900} \pm 0.002$ \\
Geneformer & $0.912 \pm 0.001$ & $\underline{0.911} \pm 0.001$ & $\mathbf{0.922 \pm 0.002}$ & $\mathbf{0.902 \pm 0.003}$ \\
scCello & $\underline{0.913} \pm 0.000$ & $\underline{0.911} \pm 0.000$ & $0.916 \pm 0.002$ & $0.891 \pm 0.003$ \\
scGPT & $0.908 \pm 0.001$ & $0.906 \pm 0.001$ & $\mathbf{0.922 \pm 0.002}$ & $\mathbf{0.902 \pm 0.003}$ \\
\bottomrule
\end{tabular}%
}
\end{minipage}
\hspace{0.01\columnwidth}
\begin{minipage}[t]{0.50\textwidth}
\centering
\captionsetup{type=table,font=small}
\caption{Spatial gene imputation performance on Lung and Breast cancer datasets. Pearson, Spearman, and Kendall's $\tau$ correlation are calculated over 10-fold cross validation on held-out gene sets, best values are in \textbf{bold}.}
\label{tab:impute_correlation_metrics}
\scriptsize
\vspace{-1.5mm}
\resizebox{\linewidth}{!}{%
\begin{tabular}{lcccccc}
\toprule
& \multicolumn{3}{c}{\textbf{Lung cancer}} & \multicolumn{3}{c}{\textbf{Breast cancer}} \\
\cmidrule(lr){2-4} \cmidrule(lr){5-7}
Method & Pearson & Spearman & Kendall's $\tau$ & Pearson & Spearman & Kendall's $\tau$ \\
\midrule
Ours & $\mathbf{0.319 \pm 0.204}$ & $\mathbf{0.282 \pm 0.195}$ & $\mathbf{0.223 \pm 0.148}$ & $\mathbf{0.388 \pm 0.198}$ & $\mathbf{0.369 \pm 0.198}$ & $\mathbf{0.279 \pm 0.147}$ \\
Geneformer & $\underline{0.314} \pm 0.204$ & $\underline{0.275} \pm 0.198$ & $\underline{0.218} \pm 0.150$ & $\underline{0.369} \pm 0.184$ & $\underline{0.357} \pm 0.198$ & $\underline{0.270} \pm 0.147$ \\
CellPLM & $0.272 \pm 0.190$ & $0.225 \pm 0.168$ & $0.179 \pm 0.128$ & $0.152 \pm 0.107$ & $0.125 \pm 0.097$ & $0.093 \pm 0.072$ \\
scCello & $0.274 \pm 0.187$ & $0.233 \pm 0.175$ & $0.185 \pm 0.131$ & $0.105 \pm 0.098$ & $0.101 \pm 0.092$ & $0.075 \pm 0.070$ \\
scGPT & $0.308 \pm 0.201$ & $0.255 \pm 0.191$ & $0.202 \pm 0.144$ & $0.290 \pm 0.176$ & $0.271 \pm 0.168$ & $0.201 \pm 0.121$ \\
\bottomrule
\end{tabular}%
}
\end{minipage}
\vspace{-6mm}
\end{table*}





\begin{figure*}[t]
\centering
\begin{minipage}[t]{0.48\textwidth}
\vspace{3mm}
\centering
\includegraphics[width=\textwidth, trim=0 1 0 1, clip]{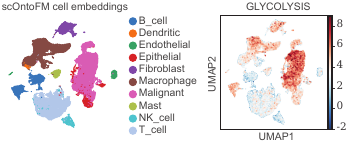}
\vspace{0.1mm}
\captionsetup{type=figure,font=small}
\caption{scOntoFM-derived cell-type clustering (Left) in lung cancer and the enrichment heatmap (Right) for the \textit{GLYCOLYSIS} hallmark gene set.}
\label{fig:geneFunc_caseStudy}
\end{minipage}
\hfill
\begin{minipage}[t]{0.5\textwidth}
\vspace{0pt}
\centering
\includegraphics[width=0.96\columnwidth]{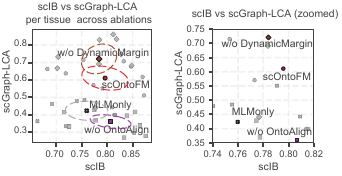}
\vspace{-1mm}
\captionsetup{type=figure,font=small}
\caption{Pareto analysis of scOntoFM loss components in scIB and scGraph-LCA. Highlighted points are the mean across nine tissues.}
\label{fig:Pareto_ablation_plot}

\end{minipage}
\vspace{-6mm}
\end{figure*}

Beyond static representations, we evaluate the model's capacity to generate context-aware gene embeddings by applying our model on a metastatic lung adenocarcinoma dataset \cite{Lung-Kim2020}. We focus on the \textit{GLYCOLYSIS} hallmark gene set to visualize how the model captures pathway activity across different cellular contexts, and we also visualize the enrichment of score of \textit{COAGULATION} and \textit{ALLOGRAFT\_REJECTION} hallmarks in AFigure~\ref{fig:geneFunc_violin_plot}.
As shown in Figure~\ref{fig:geneFunc_caseStudy}, the enrichment scores for the \textit{GLYCOLYSIS} hallmark are notably elevated in malignant cell populations, as well as cancer-associated fibroblasts (CAFs), and tumor-associated macrophages (TAMs).
This pattern reflects the high proliferative and metastatic activity of tumor cells and the immunosuppressive phenotype of associated immune cells within the tumor microenvironment \cite{glycolysis-Qiao2024}.

{\bf Cross-modal Spatial Gene Imputation} High-resolution spatial transcriptomics technologies are often constrained by the limited number of detectable genes. To address this challenge, we extend our framework to a cross-modal setting that leverages single-cell reference data to impute unmeasured genes in spatial datasets, a critical task to enhance spatial resolution \cite{imputation-benchmark-Li2022, guo25a-aistats}. As shown in Table~\ref{tab:impute_correlation_metrics}, our model consistently outperforms all baseline methods across all evaluation metrics. In particular, we achieve the highest Pearson, Spearman, and Kendall’s $\tau$ correlations in both the Lung and Breast cancer dataset, outperforming the nearest rival (Geneformer). 
We further evaluate our method on an independent SEA-AD brain spatial transcriptomics dataset (ATable~\ref{tab:AD_imputation_results}, ~\ref{fig:imputation_violin_plot}).
Despite being surpassed only by CellPLM which was trained on spatial data, our model consistently outperforms other single-modality baselines, demonstrating scOntoFM's strong cross-modal generalization by recovering meaningful spatial patterns.
{\bf Ablation Study} We conduct an ablation study to assess the contributions of the MLM loss, dynamic margin ranking loss, and ontology alignment (Figure \ref{fig:Pareto_ablation_plot}, ATable~\ref{tab:ablation_components}, ATable~\ref{tab:ablation_full_table}) in the zero-shot clustering task across nine tissues. To ensure fair comparison, all ablated variants are pretrained on the same 27.5M-cell corpus as scOntoFM. Removing ontology alignment improves scIB performance but leads to a substantial degradation in scGraph-LCA, indicating weakened ontology consistency.
In contrast, excluding the dynamic margin ranking loss enhances scGraph-LCA but slightly reduces scIB, suggesting complementary roles of these two components. Although individual components favor specific metrics, the full model (scOntoFM) achieves a balanced trade-off, simultaneously maintaining strong scIB performance and substantially improving scGraph-LCA. These results demonstrate that jointly optimizing dynamic margin ranking and ontology alignment is essential to balancing integration quality and ontology-aware structural consistency.




\section{Conclusion}
\label{sec:conclusion}
We presented scOntoFM, a novel foundation model that integrates hierarchical cell ontologies directly into the pretraining framework.
Through the use of efficient offline triplet sampling and a dynamic, hierarchy-aware loss function, our approach addresses key limitations in existing methods regarding biological interpretability.
Our results show that scOntoFM not only achieves SOTA performance across diverse downstream tasks, but also significantly improves the preservation of biological structure, offering a more robust tool for single-cell analysis. 

{\bf Limitations} This work has several limitations. $i)$ scOntoFM depends on the quality of the cell ontology and source annotations: incomplete coverage, uneven granularity, and inconsistent labels can introduce ambiguous supervision and affect the learned representations. $ii)$ triplet sampling relies on ontology relationships without jointly considering transcriptomic similarity. Consequently, it may overlook context-specific variation, as closely related cell types can exhibit distinct expression patterns across tissues, diseases, or developmental stages. $iii)$ pretraining requires substantial computational resources and curated annotations, while the storage requirements of the {\em Cell-Dataset Index Map} increase with corpus size. Future work will focus on developing a unified cell embedding space that supports harmonized, ontology-grounded annotations across cell atlases, together with covariate-conditioned sampling. These extensions aim to preserve a shared global structure while capturing context-specific variation and disease- and development-associated changes in cell state.

\bibliographystyle{plainnat}
\bibliography{scOntoFM}

\newpage
\appendix
\onecolumn

\section{Related Work}
\label{suppl:related_work}
Existing single-cell foundation models (scFMs) can be broadly categorized into three groups based on their tokenization and data processing strategies \cite{scFM_review_Szaata2024}. The first group \cite{Geneformer-Theodoris2023, scCello-yuan2024cell,Cab-Chen2026}, exemplified by Geneformer, employs rank-based tokenization, where genes are ordered by expression level to form sequence-like inputs for Masked Language Modeling (MLM). The second group \cite{scBERT-Yang2022, scGPT-Cui2024}, ({\em e.g.}, scGPT) extends this by discretizing gene expression values into consecutive bins, hence enabling joint learning of gene and cell representations. The third group \cite{cellPLM-wen2024cellplm, scFoundation-Hao2024}, known as value projection methods ({\em e.g.}, CellPLM, scFoundation), bypasses discrete tokenization by projecting raw expression counts directly into a continuous embedding space. Other recent studies \cite{GenePT-Chen2024, cell2sentence-levine2024cellsentence, cellWhisper-Schaefer2025}, have explored multi-modal learning to align single-cell data with natural language, thus leveraging the power of large language models; these methods demonstrate great potential for downstream biological discovery.

Hierarchy-aware representation learning has a long history outside single-cell biology. Hierarchical triplet loss \cite{hierarchical-tripletLoss-Ge_2018_ECCV} and dynamic metric learning \cite{dynamic-metric-learning} adapt margins to a class taxonomy; Bertinetto et al. \cite{LCA-bertinetto2020} show that LCA-height-weighted losses produce "better mistakes" on ImageNet; and Shi et al. \cite{LCA-shi2024} find that LCA-based taxonomic distance predicts out-of-distribution generalization across vision models. Within single-cell biology, OnClass \cite{onClass-Wang2021} propagates labels over the Cell Ontology to classify unseen types, treeArches \cite{treeArches-Michielsen2023} progressively builds and extends cell atlas according to the hierarchies of cell types, UCE \cite{UCE-Rosen2026} evaluates embeddings against ontology-derived distances, and hierarchical cross-entropy \cite{HCE-2026} trains annotation models against the ontology tree. scOntoFM draws on both lines: it uses LCA distance both to mine triplets offline and as the target of a rank-order alignment loss during foundation-model pre-training.

In terms of incorporating cell ontology into model pretraining, \citet{SCimilarity-Heimberg2024} proposed SCimilarity, a deep metric learning–based foundation model that combines a standard triplet loss with a reconstruction objective to embed single-cell RNA-sequencing data into a unified space. Specifically, it leverages cell ontology information to guide the sampling of semi-hard negative triplets within mini-batches. Along a similar line, scCello \cite{scCello-yuan2024cell} employs the Personalized PageRank (PPR) algorithm to quantify structural similarity between nodes in the cell ontology graph, thereby guiding the model to learn pairwise cell-type distance relationships that aligns with PPR structure.

Unlike previous approaches, we proposes a more coherent and unified framework that integrates ontology information not only into the triplet sampling stage but also into biologically constrained representation learning throughout pre-training. Concretely, compared to scCello and SCimilarity, scOntoFM enjoys specific novelty as follows: $i)$ Efficient offline ontology-aware triplet sampling, which mines more informative positive and negative pairs and facilitate effective contrastive learning. While standard online triplet sampling, like in scCello and SCimilarity, is highly inefficient for massive datasets and the probability of encountering informative positive pairs is low in a random mini-batch (see AFigure~\ref{fig:online_triplet_sampling}). $ii)$ The ontology-aware dynamic margin and global ontology alignment loss terms enables the model capture more fine-grained cell type information (see Tables \ref{tab:hiha_kidney_robust_labels}), while also maintains a faithful biological structure versus scCello (see Figure \ref{fig:zero_shot_clustering} right panel, AFigure \ref{fig:BoneMarrow_emb_heatmap}, \ref{fig:scGraph_reference_heatmap}). $iii)$ Comprehensive evaluation on both cell- and gene level task. While scCello mainly concentrates on cell level tasks, scOntoFM provides a more dedicated evaluation for scFMs in gene-level tasks (see Section \ref{sec:results}).

\section{EXPERIMENTS}
\label{suppl:experiments}
\subsection{Online-triplet sampling}
\label{suppl:online_triplet_sampling}
\begin{figure}[h]
\centering 
\includegraphics{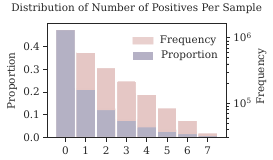} 
\caption{Distribution of positives per sample under online random sampling in 10\% of our pretraining dataset. Pink bars show the absolute frequency of positive samples, and purple bars show the corresponding proportions. 
}
\label{fig:online_triplet_sampling}
\end{figure}

In AFigure \ref{fig:online_triplet_sampling}, we show an experiment of online triplet sampling for 10\% of our pre-training data. Around 46.78\% of them have zero positives, and 21.03\% only have one positives, indicating the highly in-efficiency for online triplet sampling in large data corpus.

\subsection{Relationship between LCA distance and transcriptomic similarity}
\label{suppl:lca_vs_expr}

To test the premise that proximity in the cell ontology reflects transcriptomic proximity, we computed per cell-type transcriptomic centroids in PCA space by averaging the pre-processed gene expression through Scanpy pipeline (library size normalization, log-transform, with highly variable genes set to 2000, and principle components set to 50) over all cells belonging to that type, and measured the cosine distance between every pair of cell-type profiles. We compared these transcriptomic distances to the raw LCA distances $\Delta_{i,j}$ from \ref{eq:LCA_distance} in Tabula Sapiens v2 dataset (173 cell types across 28 organs) and the nine diverse tissue datasets from our zero-shot clustering evaluation.

\paragraph{Correlation Analysis on Tabula Sapiens v2.} As shown in Table~\ref{tab:lca_ts2} and Figure~\ref{fig:lca_level_expr_boxplot}, grouping cell pairs by the first 10 LCA distance reveals a strong positive correlation and shows a strictly monotone increase in mean expression distance (Pearson $r = 0.91$, $p<0.001$). The within-level standard deviation shrinks with LCA distance: close relatives are heterogeneous (some sibling types are transcriptionally far apart), whereas distant types are uniformly dissimilar. This broad, multi-organ analysis supports the biological premise that ontology-near cell types tend, on average, to be more transcriptionally similar, which also aligns with findings in OnClass \cite{onClass-Wang2021} and UCE \cite{UCE-Rosen2026}.

\begin{table}[ht]
\centering
\caption{The relationship between ontology LCA distance and transcriptomic distance in Tabula Sapiens v2.}
\label{tab:lca_ts2}
\scalebox{0.62}{%
\begin{tabular}{lcccccccccc}
\toprule
\textbf{LCA distance}
& \textbf{1} & \textbf{2} & \textbf{3} & \textbf{4} & \textbf{5}
& \textbf{6} & \textbf{7} & \textbf{8} & \textbf{9} & \textbf{10} \\
\midrule
Mean Expr. dist. $\pm$ sd
& 0.55$\pm$0.37 & 0.72$\pm$0.33 & 0.82$\pm$0.30 & 0.91$\pm$0.23 & 0.96$\pm$0.20
& 0.99$\pm$0.17 & 1.00$\pm$0.16 & 1.02$\pm$0.15 & 1.04$\pm$0.13 & 1.06$\pm$0.12 \\
\bottomrule
\end{tabular}%
}
\end{table}

\begin{figure*}[ht]
  \centering
  \includegraphics[width=\textwidth]{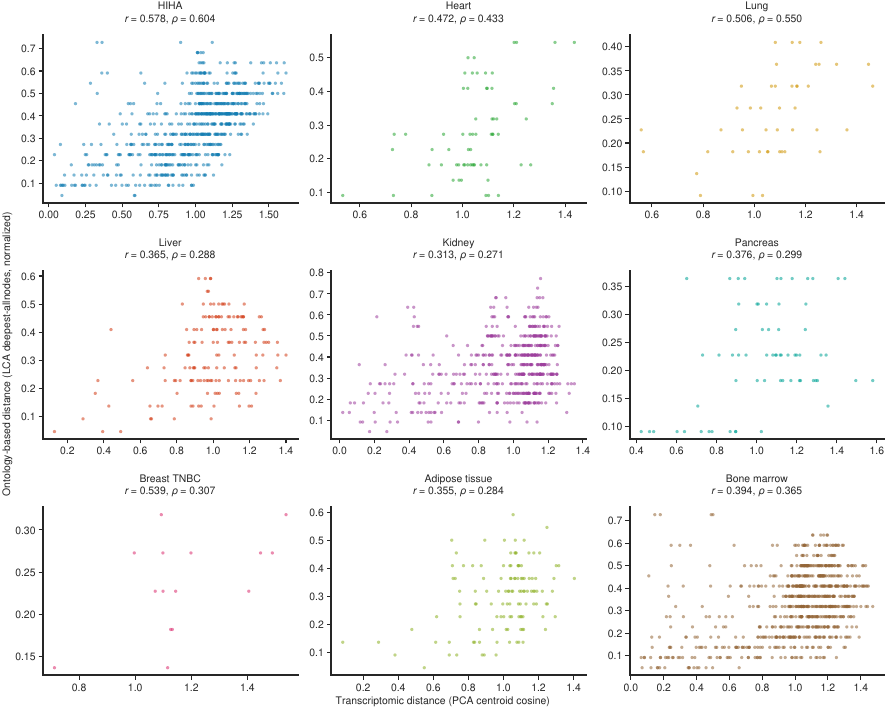}
  \vspace{-2mm}
  \caption{Scatter plots of LCA distance versus cell-type pairwise transcriptomic distance for each of the nine tissues.}
  \label{fig:nine_tissue_lca_scattered}
\end{figure*}

\paragraph{Generalization Across Nine Zero-shot Clustering Datasets.} As showed in Table~\ref{tab:lca_nine} and AFigure \ref{fig:nine_tissue_lca_scattered}, we observed a consistent positive correlation in every tissue, with an overall Pearson $r = 0.41$ ($p<0.001$) across 130 unique cell types. Correlations are strongest in immune-rich tissues (HIHA, Lung, Breast TNBC), where the ontology is deepest, and weakest in Kidney, Adipose and Liver, where many annotated types are shallow epithelial or stromal nodes.

\begin{figure*}[t]
  \centering

  \begin{minipage}[t]{0.48\textwidth}
    \centering
    \vspace{0pt}
    \begin{minipage}[c][5.5cm][c]{\linewidth}
      \centering
      \includegraphics[width=\linewidth,height=5.2cm,keepaspectratio]{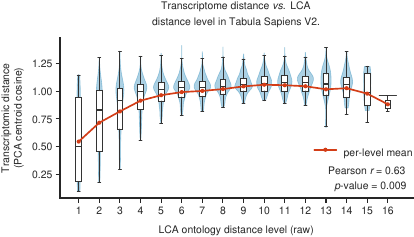}
    \end{minipage}
    \vspace{-2mm}
    \captionof{figure}{Cell-type pairwise transcriptomic distance in PCA space across raw LCA ontology distance over 1 to 16 levels in Tabula Sapiens v2. Boxes show the distance distributions, and the red line denotes the per-level mean.}
    \label{fig:lca_level_expr_boxplot_all_levels}
  \end{minipage}
  \hfill
  \begin{minipage}[t]{0.48\textwidth}
    \centering
    \vspace{0pt}
    \begin{minipage}[c][5.5cm][c]{\linewidth}
      \centering
      \includegraphics[width=\linewidth,height=5.2cm,keepaspectratio]{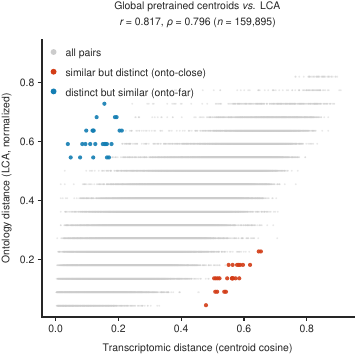}
    \end{minipage}
    \vspace{-2mm}
    \captionof{figure}{Scatter plot of the pairwise transcriptomic cell-type centroid distance versus the normalized LCA distance $\tilde{\Delta}_{i,j}$. The off-diagonal points are highlighted with \textcolor{red}{red} and \textcolor{blue}{blue}.}
    \label{fig:lca_vs_expr}
  \end{minipage}
\end{figure*}

\paragraph{Discordant pairs.} Plotting pre-trained global cell-type centroid distances against $\tilde{\Delta}_{i,j}$ (Figure~\ref{fig:lca_vs_expr}) isolates two kinds of off-diagonal pairs. $(i)$ \emph{Ontology-far, transcriptome-close}: almost entirely functional T-cell subsets that the ontology separates but the transcriptome does not ({\em e.g.}, CD8$^+$ cytotoxic {\em vs.}\ CD4$^+$ cytotoxic T cells; $\gamma\delta$ intra-epithelial {\em vs.}\ effector-memory CD8$^+$ T cells), plus cross-tissue mesenchyme (ileal smooth muscle {\em vs.}\ kidney fibroblast). $(ii)$ \emph{Ontology-close, transcriptome-far}: dominated by generic parent terms (``professional antigen presenting cell'', ``connective tissue cell'') that sit one edge from many children yet are transcriptionally far from all of them. The ontology is thus an informative prior rather than a complete model of expression; the MLM objective keeps the embedding faithful to the transcriptome in exactly these regions, which is why the full model (not the ontology-only ablations) gives the best balance in Figure~\ref{fig:Pareto_ablation_plot}.

\subsection{Pre-trained Model Configuration (Table \ref{tab:pretrained_model_config})}
\label{supp:model_config}
The pre-trained model backbone, architecture, and the training details are listed in Table \ref{tab:pretrained_model_config}.

\begin{table}[h]
\centering
\caption{Model architecture and training configuration.}
\label{tab:pretrained_model_config}
\begin{tabular}{ll}
\toprule
\textbf{Component} & \textbf{Configuration} \\
\midrule
\textbf{Backbone Model} 
& BERT (\texttt{BertForMaskedLM}) \\

\midrule
\textbf{BERT Architecture}
& Vocal size: 19941 \\
& Hidden size: 256 \\
& Layers: 6 \\
& Attention heads: 4 \\
& Intermediate size: 512 \\
& Activation: ReLU \\
& Max position embeddings: 2002 \\

\midrule
\textbf{Regularization} 
& Hidden dropout: 0.02 \\
& Attention dropout: 0.02 \\
& LayerNorm $\epsilon$: $1\times10^{-12}$ \\

\midrule
\textbf{Optimizer} 
& AdamW ($\text{lr}=1\times10^{-3}$, weight decay = 0.001) \\

\midrule
\textbf{Learning Rate Schedule} 
& WarmupDecayLR \\
& Warmup steps: 1\% of total training steps \\

\midrule
\textbf{Distributed Training} 
& DeepSpeed with ZeRO stage 2 \\
& Overlap communication enabled \\
& Reduce / allgather bucket size: $5\times10^{6}$ \\
& Contiguous gradients enabled \\

\midrule
\textbf{Precision} 
& FP16 mixed precision \\

\midrule
\textbf{Training Hyperparameters} 
& Batch size per GPU: 64 \\
& Gradient accumulation steps: 4 \\
& Effective batch size: $64 \times \text{world size} \times 4$ \\
& Epochs: 3 \\
& Gradient clipping: 1.0 \\

\bottomrule
\end{tabular}
\end{table}

\subsection{Computation cost of the Cell-Dataset Index Map ($\Omega$)}
We construct the Cell-Dataset Index Map by extracting specific columns (cell\_type\_ontology\_term\_id, dataset\_id) from the Hugging Face pre-training dataset and applying pandas' {\em groupby} operation. The process takes about 5 minutes and uses 6.5 GB of peak memory on an Intel Xeon Gold 6252 CPU. The final file size is 132 MB, ensuring that scalability to a larger corpus is highly feasible.

\begin{table}[ht]
\centering
\caption{Computational cost of constructing the Cell-Dataset Index Map.}
\label{tab:cell_dataset_index_map_cost}
\scalebox{0.9}{
\begin{tabular}{lc}
\toprule
\textbf{Metric} & \textbf{Value} \\
\midrule


CPU 
& Intel Xeon Gold 6252 \\

Time Consumption
& \(\sim\) 5 minutes \\

Peak Memory Usage 
& 6.5 GB \\

Final File Size 
& 132 MB \\
\bottomrule
\end{tabular}
}
\end{table}

\subsection{Baselines}
\label{suppl:Baselines}
{\bf Geneformer.} We followed the official guideline of Geneformer \cite{Geneformer-Theodoris2023} and downloaded the pre-training weights of Geneformer-V1-10M on \url{https://huggingface.co/ctheodoris/Geneformer/tree/main}.

{\bf scGPT.} We follow the tutorial of running scGPT \cite{scGPT-Cui2024} from \url{https://scgpt.readthedocs.io/en/latest/} and downloaded the whole-human model checkpoint (\url{https://drive.google.com/drive/folders/1oWh_-ZRdhtoGQ2Fw24HP41FgLoomVo-y?usp=sharing}) as indicated in its github repository.

{\bf scCello.} The pre-trained zero-shot scCello \cite{scCello-yuan2024cell} model weights are downloaded from hugging face with following link: \url{https://huggingface.co/katarinayuan/scCello-zeroshot}.

{\bf CellPLM.} We refer the official github repository of CellPLM \cite{cellPLM-wen2024cellplm} (\url{https://github.com/OmicsML/CellPLM}) and download its latest version checkpoint: 20230926\_85M from \url{https://github.com/OmicsML/CellPLM/tree/main/ckpt}.

{\bf SCimilarity.} We refer the official github repository of SCimilarity \cite{SCimilarity-Heimberg2024} (\url{https://github.com/Genentech/scimilarity}) and download its latest model v\_1.1 from \url{https://zenodo.org/records/10685499}.

To provide a clear overview of each baseline, in the following Table \ref{tab:pretrained_model_compare}, we summarize and compare models in terms of pre-training data, parameter size, and training strategy.

\begin{table}[h]

\centering

\caption{Model comparison in terms of pretraining data size, model scale, and modeling strategy.}
\label{tab:pretrained_model_compare}
\scalebox{0.8}{

\begin{tabular}{lccc}

\toprule

\textbf{Model} & \textbf{Pretraining Data Size} & \textbf{\# Parameters} & \textbf{Modeling Strategy} \\

\midrule

geneformer 

& 30M 

& 10{,}316{,}196 

& MLM, no ontology \\

scGPT 

& 33M 

& 51{,}330{,}049 

& Auto-regressive, no ontology \\

cellPLM 

& 11.4M 

& 82{,}402{,}543 

& VAE \\

scCello 

& 22M 

& 10{,}683{,}654 

& MLM + ontology \\

SCimilarity 

& 7.9M 

& 62{,}332{,}627

& Autoencoder + ontology \\

scOntoFM 

& 27.5M 

& 8{,}846{,}848 

& MLM + ontology \\

\bottomrule

\end{tabular}

}

\end{table}

\subsection{Zero-shot Clustering Datasets Details}
\label{suppl:dataset-zero-shot}
For the zero-shot evaluation datasets summarized in Table \ref{tab:zero_shot_dataset_summary}, we retained only primary cells from the original datasets, and no additional filtering was applied. Each data sources are listed down below in the table.

\begin{table}[h]
\centering
\caption{Summary of zero-shot clustering evaluation datasets used in this study.}
\label{tab:zero_shot_dataset_summary}
\scalebox{0.7}{
\begin{tabular}{lrc}
\toprule
\textbf{Dataset} & \textbf{\# Cells} & \textbf{Source} \\
\midrule
Blood (HIHA) 
& 263{,}917 
& \url{https://cellxgene.cziscience.com/e/e522d2cd-7927-4e59-a4ed-064009569279.cxg/} \\

Heart 
& 211{,}061 
& \url{https://cellxgene.cziscience.com/e/d567b692-c374-4628-a508-8008f6778f22.cxg/} \\

Liver 
& 72{,}710 
& \url{https://cellxgene.cziscience.com/e/84cfa5aa-a0d1-415e-ac47-eb66c9cf8170.cxg/} \\

Kidney (Atlas V2) 
& 348{,}984 
& \url{https://cellxgene.cziscience.com/collections/9c9d04c4-8899-417f-bb6f-6107dcadf14f} \\

Pancreas 
& 245{,}878 
& \url{https://cellxgene.cziscience.com/e/e51bae9a-c747-4b64-904a-4da7cda218ab.cxg/} \\

Breast TNBC 
& 52{,}681 
& \url{https://cellxgene.cziscience.com/e/9a64bf99-ebe5-4276-93a8-bee9dff1cd47.cxg/} \\

Lung 
& 30{,}472 
& \url{https://www.ncbi.nlm.nih.gov/geo/query/acc.cgi?acc=GSE131907} \\

Adipose Tissue 
& 166{,}149 
& \url{https://cellxgene.cziscience.com/e/ca46ffe6-0a26-4d8d-82fa-81722daf1102.cxg/} \\

Bone Marrow 
& 263{,}159 
& \url{https://cellxgene.cziscience.com/collections/f6c50495-3361-40ed-a819-fb9644396ed9} \\
\bottomrule
\end{tabular}
}
\end{table}

{\bf Unseen Evaluation Datasets} In Table \ref{tab:unseen_datasetid}, we show the list of query dataset ids from CZI data portal.

\begin{table}[ht]
\centering
\caption{CZI query dataset id for the 212 unseen evaluation datasets.}
\label{tab:unseen_datasetid}
\scalebox{0.7}{
\begin{tabular}{c|c|c}
\toprule
Dataset ID & Dataset ID & Dataset ID \\
\midrule
99e317c0-1fab-4f73-b511-859e752fd1c3 & 6725ee8e-ef5b-4e68-8901-61bd14a1fe73 & e2808a6e-e2ea-41b9-b38c-4a08f1677f02 \\
98f5d518-1e60-4307-b732-ab045cd3cbc2 & bab7432a-5cfe-45ea-928c-422d03c45cdd & 6c87755e-a671-41a8-9c7e-4e43b850a57b \\
e6361237-ac4e-4c5d-ad8f-f16aca0c0a8f & 242c6e7f-9016-4048-af70-d631f5eea188 & d5a2d011-3765-4dad-ac36-3fff087589fe \\
43245158-5ae1-4e71-a9a6-67eef49c26bc & a65bcc2d-4243-44c1-a262-ab7dcddfcf86 & 4e374783-0152-423e-bcb6-b87cc232f64d \\
5e7fff82-b8eb-4220-bf65-c6feb6765ea1 & 8d1dd010-5cbc-43fb-83f8-e0de8e8517da & b252b015-b488-4d5c-b16e-968c13e48a2c \\
72822932-10f6-466f-baf3-a2c1d89364bc & 881fe679-c6e0-45a3-9427-c4e81be6921f & 4dd1cd23-fc4d-4fd1-9709-602540f3ca6f \\
79ef1959-a6b4-4cac-82ca-30feaec48df1 & 01209dce-3575-4bed-b1df-129f57fbc031 & 30f5e171-83d7-4fc0-bf75-384f122346b3 \\
5d3fc988-765d-48ba-bfb5-151ef2988cac & 31285bcd-9a28-48ce-8cc0-543af80261d3 & dc30c3ec-46d6-4cd8-8ec1-b544a3d0f503 \\
3c361813-5d5e-4868-bfe5-e7591e7b5381 & 234c23d0-d6cd-4612-a40d-e5811727564b & 76150f40-1989-4977-9e23-696e72d59d9e \\
171b03c9-516b-499c-a434-fa4f70191988 & 13149914-ea03-4f01-8bf6-b793b667127b & dbf0bd35-87f8-4b25-bc90-a3c54f379907 \\
d7f5d8d0-6150-48d7-b094-c34286ad11a1 & f8d8b443-bca6-4c3c-9042-669dfb7f8030 & eeacb0c1-2217-4cf6-b8ce-1f0fedf1b569 \\
3e524d45-2265-4440-b903-3c30d51a0b17 & a20e2f2a-e171-450a-8ddc-8935c2977ad6 & 1fc48734-35ee-4372-885c-95fd2674c75a \\
731e6380-879f-4b0b-9a1f-2150208852ef & ab326369-b63c-48d8-8a7f-82ffa0df7002 & 8a554710-08bc-4005-87cd-da9675bdc2e7 \\
0c9a8cfb-6649-4d52-b418-6d8e56bd7afe & 1a974af7-8b4b-4abb-8aba-23e2979d9518 & d18736c3-6292-4379-919a-d6d973204c87 \\
9ea52f85-a35e-4b5b-9207-e2f1de7dbbcd & 87a06a2a-16c6-42f7-af1b-529fad431051 & 51d9716c-50bb-4fd8-8d1e-7d9329969cb7 \\
7c497c66-c2e2-4291-8b96-339e5e65f314 & ef1a4b05-540c-4f73-8ceb-49dfa800645d & bd3e3487-c1e6-4298-b9de-8bdcf3f938d7 \\
14363b6e-3428-45b0-8704-10fd5d6316df & 17e9d436-a264-4c94-a42d-48c8daf6fdd9 & e34f3f89-3a48-4173-a06a-9e9803f2e2b9 \\
31f657dc-1875-4c4b-a5ca-ce63b3ef3a82 & 3079e9b0-c6db-45c8-b998-a4555a73968e & d9f3a82c-24bb-4277-b715-8ec92a33dd32 \\
2856d06c-0ff9-4e01-bfc9-202b74d0b60f & 74014ef8-d2d0-4cbc-8ba2-037f30753ffd & c893ddc3-f25b-45e2-8c9e-155918b4261c \\
6f7fd0f1-a2ed-4ff1-80d3-33dde731cbc3 & e500acbf-f166-4c46-b290-f93506cf26d3 & a4157949-6f2a-40e2-b960-63f6e2bde918 \\
0de831e0-a525-4ec5-b717-df56f2de2bf0 & 93091496-be48-4122-b945-9af9c22a7535 & 73b7d0aa-035f-4a15-9f1d-f56b70204591 \\
492d9f77-c4c8-485d-92d7-7686fe6c8482 & f7ec7bd5-04ab-453b-a8a7-c9d14812affb & 2e9d2f32-4cfb-49b5-b990-cbf4c241214e \\
d6505c89-c43d-4c28-8c4f-7351a5fd5528 & 251b1a7e-d050-4486-8d50-4c2619eb0f46 & 6e00ccf7-0749-46ef-a999-dba785630d52 \\
94c41723-b2c4-4b59-a49a-64c9b851903e & b25f3834-69b3-4d87-a272-3938432d1f30 & 34f5307e-7b4d-4a48-b68f-2ba844c6414b \\
7f7faf6b-f11d-4f07-bc1c-188a4472748d & a9bedd04-51cb-42ad-83fe-cd749de71f1b & 78c62d4e-abf1-405b-bb51-368720814529 \\
9fcb0b73-c734-40a5-be9c-ace7eea401c9 & f0f0d7c4-3bec-428e-9539-c99d36548d96 & 4a5b00e0-1ba3-4fd4-af89-d3512eb20720 \\
b6e502d7-0d07-408e-a291-91b63af0a40a & 5b25a46f-f1cf-4caa-9e10-6ab81dfe9174 & 5e717147-0f75-4de1-8bd2-6fda01b8d75f \\
50eb1e23-b8d4-4f76-a184-44e5541fa05a & 55ca4411-8c7d-4aef-a572-50852e030c05 & 4023a2bc-6325-47db-bfdf-9639e91042c2 \\
cd2f23c1-aef1-48ae-8eb4-0bcf124e567d & 93cb76aa-a84b-4a92-8e6c-66a914e26d4c & e347396c-a7ff-4691-9f7a-99a43555ca18 \\
cb252df6-6e49-4553-abd1-495a00006fb1 & 92161459-9103-4379-ae34-73a38eee1d1d & 4a90bbba-df97-4a28-83ae-7386fe7d9167 \\
9c1b5626-58df-4401-ae7b-f66d068c1551 & 68b6114f-e990-4033-bfb3-c35536633aaa & 093d3bfe-6f0f-4ac0-a7a1-829f94d0a49f \\
07760522-707a-4a1c-8891-dbd1226d6b27 & b62f728b-a7eb-4889-8162-b3d457de93b9 & f5be4b96-f5a3-4c3d-84ac-6f69daf744d5 \\
595c9010-99ec-462d-b6a1-2b2fe5407871 & ce009dc1-ac57-4386-b72f-5c575701c253 & 8fbed309-d3d4-441b-b3ff-e2dcbcec2d35 \\
cab0bc48-744c-461b-aaf4-7bf2cb7af00d & cff99df2-4904-44f7-9173-ff837f95606e & 06ef6b36-6c9b-4e10-8a94-d0baf274276e \\
f7ca7eb9-0919-4eba-99a1-21ba382b4174 & 8dc29642-be06-46dc-a289-8075915745d4 & b3c9904b-2ffd-4038-8620-ee1c5b072e9c \\
8c42cfd0-0b0a-46d5-910c-fc833d83c45e & 8e5978bd-0e48-4a09-a92d-fa2985059f05 & 429d21fd-bb9a-4abf-922b-c15138524bdd \\
5829c7ba-697f-418e-8b98-d605b192dc48 & 6929d6d1-6f81-44d7-bd35-5a580259af81 & b6b5ea88-e092-46e4-b9c5-e93b52d5c195 \\
4859642c-d155-437b-adf5-83ca6c1bef67 & 821a8e3e-87ec-41ee-bd69-cd4377057d38 & 0bae7ebf-eb54-46a6-be9a-3461cecefa4c \\
a30db232-08a6-48f0-ab9f-ad6977b8a6b1 & be41a86a-b606-4b1c-8055-32f334898775 & 76af615b-12fe-4ebb-b8c6-239ad29e26e3 \\
2a65bd41-4a3f-42f4-98b2-c53962811152 & ed9e9f96-4f08-49d2-bef5-b2c29adf3edc & b7efc0ba-9136-4f3f-adb2-d07849e59a3d \\
19e46756-9100-4e01-8b0e-23b557558a4c & e595c979-c2de-4dad-8c02-8334a8e56de6 & eec804b9-2ae5-44f0-a1b5-d721e21257de \\
ca46ffe6-0a26-4d8d-82fa-81722daf1102 & 2727d83a-0af0-443a-bff8-58dc7028289a & e7c5ba7e-4be0-4696-af50-a07d0dc7aac5 \\
a82c43bb-a703-446d-aa24-049bf013121f & 716a4acc-919e-4326-9672-ebe06ede84e6 & 12c3efd8-d515-42e6-b93c-4961e599d524 \\
9fddb063-056d-4202-8b8a-4b0ee531d3ce & 0bc7235a-ae5a-479d-a487-510435377e55 & 8d73847b-33e7-48f0-837f-e3e6579bf12d \\
e3cc5b85-62ec-4add-bee3-73c8fd11e59d & e2a325e2-5633-4950-ae46-ebff70979e3b & 0a2d7e87-c3c0-4ed2-86df-ae18811fcc16 \\
01c93cf6-b695-4e30-a26e-121ae8b16a9e & be401db3-d732-408a-b0c4-71af0458b8ab & 2f05ab20-a092-4bab-9276-3e0eb24e3fee \\
b550fd94-97e0-42e2-90f7-7434781ea2f8 & 43560d7c-acd6-4ede-830d-c9ffc6cd0862 & e5f5d954-cf0e-4bd8-9346-8d1ddf15a08b \\
a19d1667-a7b5-4556-9e5f-f9bfa690c0f1 & a12ccb9b-4fbe-457d-8590-ac78053259ef & 7b55fe5c-d5c8-48e0-a1a3-54c5b9074f3f \\
1873a18a-66fd-4a4d-8277-a872c93f5b59 & 4ce8d8ec-b94c-4b79-afcb-b094336d8a78 & 149b2c3f-ee11-47a7-984b-923570280bd7 \\
1a38e762-2465-418f-b81c-6a4bce261c34 & 51f476f7-b24d-42f3-8871-7dab3fa35e96 & 3966ba97-beb8-4d0b-9954-d3775cd2cd61 \\
d567b692-c374-4628-a508-8008f6778f22 & 4993d61c-1d04-4630-9c61-8d9431f39adc & df1edb87-e512-43ae-b5f4-cb179cfc2bb4 \\
6c600df6-ddca-4628-a8bb-1d6de1e3f9b4 & f9cfac8d-bff6-47a2-a1f6-503827d375f5 & 15d374d6-0dfd-4d8e-ade7-81f73dc921ee \\
bcdec5fa-a7fa-4806-92bc-0cd02f40242f & 6ceeaa86-9ceb-4582-b390-6d4dd6ff0572 & c54c9659-1b6b-4d4a-96f0-7dc987410f2d \\
59b69042-47c2-47fd-ad03-d21beb99818f & 33911db3-f461-464b-8083-a397ab616a09 & cf83c98a-3791-4537-bbde-a719f6d73c13 \\
29244e1d-02e6-4133-b411-516ef7474638 & 7d98cc44-b090-4dc8-804f-2750c84fe9d7 & ac0c6561-7a48-4185-af6f-af799f699172 \\
9df60c57-fdf3-4e93-828e-fe9303f20438 & 350237e0-9f48-4cbd-9140-3b44495549f3 & 390b376d-8857-4b4b-991f-0d7326938172 \\
02792605-4760-4023-82ad-40fc4458a5db & 8dc858b0-b076-4326-94bb-7c2d1cd721a3 & ea786a06-5855-48b7-80d7-0313a21a2044 \\
f1606894-59df-4794-a37f-baa7c6fb6de1 & da75ce6d-a395-4abd-962b-267aadb99666 & 9a281de7-cee5-4e80-8584-1929f46f152f \\
2dd9d7a2-4bee-4c3f-a1f5-89bc3271c4f0 & 62de80d7-e6c4-4ff0-ad4c-d3d36f57cb93 & d967b47c-a9e6-4337-b2f4-977f690cb67f \\
84230ea4-998d-4aa8-8456-81dd54ce23af & 0087cde2-967d-4f7c-8e6e-40e4c9ad1891 & db59611b-42de-4035-93aa-1ed39f38b467 \\
72955cdb-bd92-4135-aa52-21f33f9640db & 364bd0c7-f7fd-48ed-99c1-ae26872b1042 & 7e7f63c5-d964-40be-83de-ecbcccafd233 \\
bdd109f1-ca27-45dc-bb9f-473008449428 & 96baf2e0-60c5-48cf-b1d1-cba657851507 & b61a921b-7fa3-4b42-b455-aaaf32447920 \\
3b8b5de4-3aa1-4ac6-8890-8d03c8219981 & 2ac76f1b-43ef-4271-8686-2f165570989f & 3241c34f-581a-4de3-81fe-28e3de182680 \\
f5be9ed2-5d41-4a52-91e4-4ff24ff84900 & c874f155-9bf9-4928-b821-f52c876b3e48 & 8f4f8502-9170-4ac2-9707-3b6985ebfe5f \\
55003f67-c494-46f1-83fb-902745646379 & 78f40dd6-ed22-470d-8547-6fb5746c25da & f156606a-dd9a-49fd-bc40-0e069b6cf07c \\
460dfe21-0550-4521-92fe-db0f3036d8b0 & 2f7230e7-4706-4924-b571-473fc1e52c30 & b6ee0939-b5b9-4e03-9d69-96fb6bf5d434 \\
9a64bf99-ebe5-4276-93a8-bee9dff1cd47 & 1196334f-a441-474a-95ee-400780941690 & f5b0810c-1664-4a62-ad06-be1d9964aa8b \\
601c7454-17ff-432c-8e0d-902bfddb8013 & ecce9688-66ea-4600-98f7-84aed7421c01 & eb499fd8-7000-419f-8854-926b9b61b11f \\
d3cb449b-c2b1-4b50-a7f1-21203535fe61 & 0436a180-cb44-47ba-8ffa-807b7a468469 & d3d4baff-d142-4f13-acc6-96a4ef6a3e95 \\
a199ca73-035d-44e2-9893-4c493151db21 & dea717d4-7bc0-4e46-950f-fd7e1cc8df7d & 34229bdc-a895-4394-8820-574e4028d8c6 \\
7b20c613-9add-43d1-87e9-defd3d9b9f8c & ea95cdd5-cf7e-4d5d-a0bd-032776d32ae5 \\
\bottomrule
\end{tabular}
}
\end{table}

\subsection{Cell Type Annotation Fine-tuning}
\label{suppl:cell-anno-finetune}
We peformed cell type classification on the HIHA and Pancreas datasets, with fixed held-out testing donors (UP1008 and UP1012 in HIHA; SAMN16547514, SAMN17209843, SAMN17528599, SAMN17831932 in Pancreas) and identical hyperparameters across all scFMs. To ensure a fair comparison, we utilized a standardized Multi-Layer Perceptron (MLP) head classifier for all foundational models. The network takes the extracted 2D continuous cell embeddings as input and processes them through two hidden layers (1024 and 512 dimensions), each followed by 1D Batch Normalization, a ReLU activation, and Dropout regularization (rate of 0.2) to mitigate overfitting. The model was optimized to minimize Cross-Entropy Loss using the AdamW optimizer with a learning rate of $1 \times 10^{-3}$ and weight decay of $1 \times 10^{-4}$, while reserving 10\% of the training donor cells as a validation set for early stopping. We also compared with a specialized cell annotation model with hierarchical cross-entropy (HCE) loss \cite{HCE-2026}, using the authors’ released MLP and TabNet checkpoints and their penultimate-layer embeddings, the same donor splits, and the same MLP probe. The results (Table \ref{tab:cellAnno_finetune}, ATable \ref{tab:hce_cell_annotation}) are reported as mean±std over 5 random seeds and confirm that scOntoFM also maintain its significant performance advantage during fine-tuning stage.

\begin{table}[ht]
\centering
\caption{Benchmark HCE model in cell-type annotation (fine-tuning) on the HIHA and Pancreas datasets. Results are reported as mean $\pm$ standard deviation.}
\label{tab:hce_cell_annotation}
\scalebox{0.9}{%
\begin{tabular}{llcc}
\toprule
\textbf{Dataset} & \textbf{Model} & \textbf{Accuracy} & \textbf{Macro-F1} \\
\midrule
\multirow{3}{*}{HIHA}
& scOntoFM   & \textbf{0.92 $\pm$ 0.00} & \textbf{0.77 $\pm$ 0.01} \\
& HCE-TabNet & \underline{0.91 $\pm$ 0.00} & \underline{0.71 $\pm$ 0.02} \\
& HCE-MLP    & 0.90 $\pm$ 0.00 & 0.70 $\pm$ 0.00 \\
\midrule
\multirow{3}{*}{Pancreas}
& scOntoFM   & \textbf{0.99 $\pm$ 0.00} & \textbf{0.97 $\pm$ 0.01} \\
& HCE-TabNet & \underline{0.98 $\pm$ 0.00} & \underline{0.90 $\pm$ 0.01} \\
& HCE-MLP    & 0.95 $\pm$ 0.00 & 0.88 $\pm$ 0.01 \\
\bottomrule
\end{tabular}%
}
\end{table}






\subsection{Novel Cell Identification}
\label{suppl:novel_cell}
This approach assumes that the biological relationships defined in a Cell Ontology (hierarchical closeness) translate directly to geometric closeness in the cell representation space. Cell types that are closely related in the Cell Ontology ({\em e.g.}, "CD4+ T cell" and "CD8+ T cell") share significant transcriptomic programs. Therefore, the embedding (centroid) of an unseen cell type can be reliably approximated as a weighted interpolation of its known evolutionary or functional relatives. The novel cell identification process could be summarized as follows:

\paragraph{1. Compute Ontology Affinity Matrix}
We define an affinity matrix $\mathbf{A} \in \mathbb{R}^{m \times n}$ between unseen and seen cell types using a Gaussian kernel on the LCA distances:
\begin{equation}
    A_{ij} = \exp\left(-\gamma \cdot \mathbf{\Delta}(u_i, c_j)\right) ,
\end{equation}
where $\gamma$ is a scaling hyperparameter, we set it to 1 by default, and $u_i$ is the unknown cell type, and $c_j$ is the known cell type. The Gaussian kernel enables us to transform the raw LCA (Lowest Common Ancestor) distances into normalized similarity weights.

\paragraph{2. Normalize Weights}
We normalize the affinities so that the weights for each unseen cell type sum to 1:
\begin{equation}
    W_{ij} = \frac{A_{ij}}{\sum_{k=1}^{n} A_{ik}} .
\end{equation}
Let $\mathbf{W} \in \mathbb{R}^{m \times n}$ be the resulting weight matrix.

\paragraph{3. Generate Novel Prototypes}
We calculate the predicted centroid $\hat{\mathbf{p}}_{u_i}$ for each novel cell type by linearly combining the known centroids $\mathbf{Q} \in \mathbb{R}^{n \times d}$:
\begin{equation}
    \hat{\mathbf{P}}_{\mathcal{U}} = \mathbf{W} \cdot \mathbf{Q} ,
\end{equation}
where $\hat{\mathbf{P}}_{\mathcal{U}} \in \mathbb{R}^{m \times d}$ contains the estimated embeddings for the unseen type centroids, $\mathcal{U}=\{u_i\}_{i=1}^m$ is the novel cell-type set containing all the unknown cell types

\paragraph{4. Classification}
Given a novel cell and and its corresponding embedding $\mathbf{z}$, which could be classified by finding the nearest unseen centroid using cosine similarity:
\begin{equation}
    \hat{y} = \underset{u_i \in \mathcal{U}}{\operatorname{argmax}} \left( \frac{\mathbf{z} \cdot \hat{\mathbf{p}}_{u_i}}{\|\mathbf{z}\| \|\hat{\mathbf{p}}_{u_i}\|} \right) .
\end{equation}

{\bf Novel Cell-type List} A summary of the 99 novel cell types is provided in ATable \ref{tab:novelCL_full_table}, along with the number of cells for each cell type. For cell types with more than 10,000 cells, we randomly subsample 10,000 cells. In total, the novel cell evaluation dataset contains 774,215 cells.

\begin{table}[ht]
\centering
\caption{Predicted accuracy per novel cell type across SOTA methods. The best result for each cell type is shown in \textbf{bold}.}
\label{tab:novelCL_full_table}

\resizebox{\textwidth}{!}{%
\begin{tabular}{lccccccc|lccccccc}
\toprule
CL ID & cellPLM & Geneformer & Ours & scCello & scGPT & SCimilarity & count & CL ID & cellPLM & Geneformer & Ours & scCello & scGPT & SCimilarity & count \\
\midrule
CL:0000030 & 0.015 & 0.004 & \textbf{0.041} & 0.02 & 0.002 & 0 & 10000 & CL:0009070 & 0 & 0.061 & 0.003 & \textbf{0.304} & 0.001 & 0 & 5888 \\
CL:0000047 & 0 & 0 & \textbf{0.024} & 0 & 0 & 0 & 10000 & CL:0009071 & \textbf{0.949} & 0.748 & 0.886 & 0.433 & 0.836 & 0.787 & 3641 \\
CL:0000060 & 0 & 0 & 0.001 & 0 & \textbf{0.002} & 0 & 4451 & CL:0010012 & 0.05 & 0.01 & 0.001 & 0 & 0 & \textbf{0.082} & 10000 \\
CL:0000083 & 0 & 0.028 & 0.004 & 0.342 & 0.183 & \textbf{0.856} & 9257 & CL:0011031 & 0.576 & 0.675 & 0.938 & 0.983 & 0.873 & \textbf{0.998} & 10000 \\
CL:0000121 & 0 & 0.001 & 0 & \textbf{0.17} & 0 & 0 & 10000 & CL:0019021 & 0.408 & 0.089 & \textbf{0.649} & 0.447 & 0.357 & 0.167 & 4533 \\
CL:0000159 & \textbf{1} & 0.157 & \textbf{1} & 0.131 & \textbf{1} & 0.063 & 5088 & CL:1000022 & 0.498 & 0.59 & 0.882 & \textbf{0.946} & 0.498 & 0.052 & 3513 \\
CL:0000338 & 0 & 0 & 0 & 0 & 0 & 0 & 10000 & CL:1000042 & 0 & 0 & 0 & 0 & 0 & 0 & 10000 \\
CL:0000515 & 0 & 0 & \textbf{0.829} & 0.577 & 0 & 0.296 & 10000 & CL:1000398 & 0 & 0 & 0 & 0 & 0 & 0 & 7050 \\
CL:0000529 & 0.154 & 0 & 0.404 & \textbf{0.773} & 0 & 0 & 10000 & CL:1000443 & 0.996 & 0.002 & \textbf{0.997} & 0.766 & 0.996 & 0.994 & 10000 \\
CL:0000549 & 0 & 0.012 & 0 & 0.001 & \textbf{0.158} & 0 & 5253 & CL:1000500 & 0.008 & 0 & 0.004 & \textbf{0.089} & 0.004 & 0.003 & 8409 \\
CL:0000550 & 0 & \textbf{0.008} & 0.001 & 0 & 0.005 & 0 & 5061 & CL:1001567 & 0.018 & 0.001 & 0.012 & \textbf{0.116} & 0.023 & 0.023 & 10000 \\
CL:0000552 & 0.974 & 0.973 & 0.975 & 0.973 & 0.968 & \textbf{0.986} & 3734 & CL:1001569 & 0 & 0 & 0 & 0 & 0 & 0 & 9976 \\
CL:0000553 & 0.663 & 0.848 & 0.708 & 0.254 & \textbf{0.917} & 0.899 & 2564 & CL:1001576 & 0.019 & 0.022 & \textbf{0.209} & 0.167 & 0.14 & 0.005 & 7386 \\
CL:0000598 & 0 & 0 & \textbf{0.045} & 0.039 & 0.005 & 0 & 10000 & CL:1001609 & 0 & 0 & 0.046 & 0.041 & 0 & \textbf{0.048} & 7010 \\
CL:0000700 & 0 & 0 & 0 & 0 & 0 & 0 & 6364 & CL:2000016 & 0.008 & 0.006 & 0.018 & \textbf{0.102} & 0.013 & 0.004 & 3430 \\
CL:0000790 & 0.005 & 0.115 & \textbf{0.438} & 0.354 & 0.157 & 0.385 & 10000 & CL:2000029 & 0.031 & 0 & 0.084 & \textbf{0.238} & 0.057 & 0.006 & 10000 \\
CL:0000867 & 0.015 & 0.099 & 0.071 & \textbf{0.313} & 0.014 & 0.094 & 4554 & CL:2000054 & 0.564 & 0.536 & \textbf{0.747} & 0.576 & 0.584 & 0.64 & 9320 \\
CL:0000912 & 0.332 & 0.122 & \textbf{0.341} & 0.208 & 0.315 & 0.168 & 10000 & CL:4023009 & 0 & 0 & 0.421 & \textbf{0.642} & 0 & 0 & 10000 \\
CL:0000920 & 0.054 & 0.03 & 0.253 & \textbf{0.402} & 0.176 & 0.21 & 8377 & CL:4023026 & 0 & 0 & 0 & 0 & 0 & 0 & 5385 \\
CL:0001033 & \textbf{0.401} & 0.011 & 0.018 & 0.078 & 0.147 & 0.001 & 10000 & CL:4023029 & 0 & 0 & 0 & 0 & 0 & 0 & 6172 \\
CL:0001034 & 0 & 0 & 0 & 0 & 0 & 0 & 10000 & CL:4023069 & 0 & 0 & \textbf{0.619} & 0.404 & 0 & 0 & 6404 \\
CL:0001049 & 0.082 & 0.028 & \textbf{0.661} & 0.079 & 0.054 & 0.131 & 5698 & CL:4023111 & 0 & 0 & 0 & 0 & 0 & 0 & 10000 \\
CL:0001055 & 0.813 & 0.848 & 0.842 & 0.831 & 0.847 & \textbf{0.915} & 2494 & CL:4023161 & 0 & 0.001 & 0 & 0.009 & 0 & \textbf{0.301} & 3847 \\
CL:0002032 & 0 & 0.075 & \textbf{0.63} & 0.497 & 0.364 & 0 & 6718 & CL:4028002 & 0 & 0 & 0 & 0 & 0 & 0 & 6368 \\
CL:0002046 & 0.965 & 0.885 & 0.96 & 0.577 & 0.952 & \textbf{0.992} & 10000 & CL:4030006 & 0.001 & 0.017 & 0.079 & 0.019 & \textbf{0.116} & 0.096 & 10000 \\
CL:0002178 & 0.584 & 0.552 & 0.606 & 0.155 & \textbf{0.868} & 0.819 & 10000 & CL:4030034 & \textbf{0.997} & 0.983 & 0.986 & 0.962 & 0.978 & 0.972 & 4462 \\
CL:0002211 & 0.196 & 0.195 & 0.302 & 0.176 & 0.176 & \textbf{0.742} & 10000 & CL:4030055 & 0.422 & 0.323 & 0.33 & \textbf{0.846} & 0.8 & 0.364 & 10000 \\
CL:0002248 & 0 & 0 & 0 & 0 & 0 & 0 & 2942 & CL:4030056 & 0 & 0 & 0 & 0 & 0 & 0 & 3945 \\
CL:0002259 & 0 & 0 & 0 & 0 & 0 & 0 & 4308 & CL:4030059 & \textbf{0.87} & 0.107 & 0.118 & 0.125 & 0.837 & 0.005 & 10000 \\
CL:0002303 & 0 & 0 & 0 & 0 & 0 & 0 & 10000 & CL:4030063 & 0 & 0 & 0 & 0 & 0 & 0 & 3978 \\
CL:0002304 & 0 & 0 & 0 & 0 & 0 & 0 & 5330 & CL:4033008 & 0.841 & \textbf{0.969} & 0.964 & 0.484 & 0.827 & 0.912 & 8339 \\
CL:0002325 & 0.739 & 0.07 & \textbf{0.997} & 0.88 & 0.671 & 0.874 & 10000 & CL:4033014 & 0 & 0 & \textbf{0.006} & \textbf{0.006} & 0 & 0 & 9787 \\
CL:0002334 & 0.003 & 0.004 & \textbf{0.011} & 0.002 & 0.005 & 0.01 & 10000 & CL:4033019 & 0.906 & 0.017 & 0.872 & 0.115 & 0.086 & \textbf{0.953} & 2830 \\
CL:0002362 & \textbf{0.198} & 0.065 & 0.058 & 0.04 & 0.151 & 0.002 & 2714 & CL:4033027 & 0.081 & 0.003 & 0.173 & 0.569 & \textbf{0.57} & 0.004 & 10000 \\
CL:0002547 & \textbf{0.003} & 0.002 & 0 & 0 & 0 & 0 & 4757 & CL:4033028 & 0 & 0 & 0 & 0 & 0 & 0 & 10000 \\
CL:0002549 & 0.005 & 0 & 0.032 & 0.005 & 0.001 & \textbf{0.133} & 2058 & CL:4033029 & 0 & 0 & 0 & 0 & 0 & 0 & 10000 \\
CL:0002551 & 0.047 & 0.558 & 0.622 & 0.627 & 0.082 & \textbf{0.793} & 10000 & CL:4033030 & 0 & 0 & 0 & 0 & 0 & 0 & 10000 \\
CL:0002615 & 0.996 & 0.476 & 0.991 & \textbf{0.999} & \textbf{0.999} & 0.931 & 10000 & CL:4033031 & 0 & 0 & 0 & 0 & 0 & 0 & 10000 \\
CL:0002627 & 0.578 & 0.809 & 0.893 & 0.115 & \textbf{0.904} & 0.418 & 10000 & CL:4033032 & 0 & 0 & 0 & 0 & 0 & 0 & 10000 \\
CL:0002629 & 0.028 & 0.258 & 0.472 & 0.794 & \textbf{0.922} & 0.895 & 10000 & CL:4033033 & 0 & 0 & 0 & 0 & 0 & 0 & 10000 \\
CL:0002671 & \textbf{0.067} & 0.008 & 0.01 & 0.034 & 0.019 & 0.034 & 4225 & CL:4033034 & 0 & 0 & 0 & 0 & 0 & 0 & 10000 \\
CL:0002672 & 0.002 & 0 & \textbf{0.016} & 0.009 & 0 & 0.005 & 5149 & CL:4033035 & 0 & 0 & 0 & 0 & 0 & 0 & 10000 \\
CL:0003050 & \textbf{1} & 0.985 & \textbf{1} & \textbf{1} & \textbf{1} & \textbf{1} & 7710 & CL:4033036 & 0 & 0 & 0 & 0 & 0 & 0 & 10000 \\
CL:0004217 & \textbf{0.999} & 0.675 & 0.987 & 0.995 & \textbf{0.999} & \textbf{0.999} & 10000 & CL:4033039 & \textbf{0.656} & 0.274 & 0.287 & 0.212 & 0.176 & 0.188 & 4520 \\
CL:0004218 & 0 & 0 & 0 & 0 & 0 & 0 & 10000 & CL:4033046 & 0.064 & 0 & 0.633 & \textbf{0.841} & 0.722 & 0.002 & 10000 \\
CL:0004232 & 0 & 0 & 0.497 & 0.547 & 0.13 & \textbf{0.957} & 10000 & CL:4033047 & 0 & 0 & 0 & 0 & 0 & 0 & 10000 \\
CL:0005022 & 0 & 0.003 & \textbf{0.019} & \textbf{0.019} & \textbf{0.019} & 0.001 & 8796 & CL:4033048 & 0 & 0.001 & 0.001 & 0.009 & 0.002 & \textbf{0.02} & 4318 \\
CL:0005026 & 0.272 & 0.478 & 0.516 & \textbf{0.903} & 0.532 & 0.387 & 10000 & CL:4033051 & 0 & 0 & 0 & 0 & 0 & 0 & 10000 \\
CL:0008031 & 0.165 & 0.088 & 0.168 & 0.059 & \textbf{0.436} & 0.289 & 10000 & CL:4033052 & 0 & 0 & 0 & 0 & 0 & 0 & 6102 \\
CL:0008035 & 0.816 & 0.04 & \textbf{0.898} & 0.727 & 0.476 & 0.866 & 10000 &  &  &  &  &  &  &  &  \\
\bottomrule
\end{tabular}%
}

\vspace{2mm}

\scalebox{0.8}{%
\begin{tabular}{lccccccc}
\toprule
& cellPLM & Geneformer & Ours & scCello & scGPT & SCimilarity & total count \\
\midrule
Macro accuracy
& 0.213 & 0.151 & \textbf{0.286} & 0.254 & 0.244 & 0.240
& 774,215 \\
Average rank
& 3.712 & 4.202 & \textbf{2.848} & 3.237 & 3.485 & 3.515
& 774,215 \\
\bottomrule
\end{tabular}%
}
\end{table}

\subsection{Gene Function Prediction}
\label{suppl:gene_func_pred}
\subsubsection{Static Gene Embedding}

 To evaluate the semantic quality of the \textit{static} gene representations, we perform a supervised multi-lable classification task to predict Gene Ontology (GO) namespace membership by using the \textit{static} gene embedding extracted from scFMs.

{\bf Problem Formulation}
Let $\mathcal{M} = \{m_1, m_2, \dots, m_{|\mathcal{M}|}\}$ be the set of genes. For each gene $m_i$, we extract a \textit{static} embedding vector $\mathbf{e}_i \in \mathbb{R}^d$ directly from the input token embedding table of the pre-trained foundation model ({\em e.g.}, Geneformer, scCello, or scGPT). This approach evaluates the model's \textit{zero-shot} understanding of gene identity without updating the encoder weights. The target labels are defined by the three root namespaces of the Gene Ontology: Biological Process (BP), Cellular Component (CC), and Molecular Function (MF). We define the label vector $\mathbf{y}_i \in \{0, 1\}^3$ for gene $m_i$ based on memberships defined in the MSigDB\footnote{MSigDB website: \url{https://www.gsea-msigdb.org/gsea/msigdb} \label{MSigDB}} C5 gene sets, where $y_{i,j} = 1$ if the gene belongs to namespace $j$.

{\bf Gene Ontology Gene Sets} The GO terms are downloaded from MsiDB C5 collections, encompassing 7583 BP components, 1042 CC components, and 1855 MF components, forming 10480 gene sets in total.

{\bf Model Architecture} We employ a 3-layer Multilayer Perceptron (MLP) as a probing classifier. The model architecture and training details are summarized in Table \ref{tab:gene_funcPred_MLP_config}. To ensure a fair comparison, we apply same settings for all the scFMs without specifically tuning the hyper-parameters.

\begin{table}[h]
\centering
\caption{Summary of the architecture and training configuration for static gene embedding GO function classifier.}
\label{tab:gene_funcPred_MLP_config}
\begin{tabular}{ll}
\toprule
\textbf{Category} & \textbf{Specification} \\
\midrule
Input dimension & $d$ (embedding dimension) \\
Hidden layer & 128 units, ReLU activation \\
Bottleneck layer & 64 units, ReLU activation \\
Output layer & 3 units (multi-label classification) \\
Dropout rate & 0.2 (applied after each hidden layer) \\
\midrule
Loss function & Binary Cross-Entropy with Logits \\
Optimization & Adam optimizer \\
Learning rate & $1 \times 10^{-4}$ \\
Batch size & 64 \\
Training epochs & 15 \\
Task formulation & Multi-label classification ($C=3$) \\
\bottomrule
\end{tabular}
\end{table}





{\bf Experiments and Evaluation} To ensure robust evaluation, we perform cross-validation (set $k$ folder to 20) with stratified sampling for the train/test splits. Performance is reported as the mean and standard deviation across the 20 splits for Micro-F1, Macro-F1, Micro-Average Precision (AP), and Macro-AP, the evaluation metrics will be introduced in more detail in Appendix \ref{suppl:gene_func_metrics}.

\subsubsection{Contextual Gene embedding}
To interpret the biological signals captured by the foundation model, we perform pathway activity inference using the \textit{decoupler} package \cite{decoupleR-BadiaiMompel2022} for the lung adenocarcinoma dataset \cite{Lung-Kim2020}. Unlike standard approaches that utilize raw read counts, we leverage the contextualized gene embedding generated by the model.

{\bf Input Data Preparation}
Let $\mathbf{H}^{(i)} \in \mathbb{R}^{L \times d}$ be the hidden state representation for cell $i$ extracted from the last layer of the pretrained Transformer model, where $L$ is the sequence length (number of genes) and $d$ is the embedding dimension. We obtain a scalar intensity value $E_{ij}$ for the $j$-th gene in cell $i$ by averaging over the embedding dimension:
\begin{equation}
    E_{ij} = \frac{1}{d} \sum_{k=1}^{d} H_{ijk} .
\end{equation}

This operation results in a dense matrix $\mathbf{E} \in \mathbb{R}^{N \times M}$ representing the mean activation intensity of each gene within its specific cellular context.

{\bf Pathway Activity Inference}
We employ the \textbf{Univariate Linear Model (ULM)} in the \textit{decoupler} package \cite{decoupleR-BadiaiMompel2022} for activity estimation, and download the prior knowledge network from the MSigDB Hallmark collection\textsuperscript{\ref{MSigDB}}. For each cell $i$, ULM fits a linear model to determine if the continuous values in $\mathbf{E}_{i}$ corresponding to a specific pathway gene set are significantly higher than the background distribution.
The resulting matrix $\mathbf{Y} \in \mathbb{R}^{N \times K}$ contains $z$-scores (enrichment scores) representing the inferred activity of $K$ biological pathways across $N$ cells. Similarly, we could easily extend this framework to other available gene sets ({\em i.e.}, MSigDB C3 regulatory taget gene sets, C5 ontology gene sets, or resources inside Omnipath \cite{Omnipath-Trei2016}) and perform biological discovery with our model's contextual embedding. More information about the \textit{decoupler} package and the analysis pipeline could be found in the official website\footnote{decoupler single-cell enrichment analysis: \url{https://decoupler.readthedocs.io/en/latest/notebooks/scell/rna_sc.html}}.

{\bf Differential Pathway Analysis}
To identify cell-type-specific functional signatures, a differential activity analysis was performed on the matrix $\mathbf{Y}$, containing the pathway enrichment score for each cell. We group cells by their annotated types $c \in \mathcal{C}$, where $C$ is the total cell types in the dataset, and apply a t-test with overestimated variance:

\begin{equation}
    t = \frac{\mu_{k, c} - \mu_{k, \mathcal{C}\setminus c}}{\sqrt{\frac{\sigma^2_{k, c}}{n_c} + \frac{\sigma^2_{k, \mathcal{C}\setminus c}}{n_{\mathcal{C}\setminus c}}}} ,
\end{equation}

\noindent where $\mu_{k, c}$ is the mean activity of pathway $k$ in cell type $c$, and $\mathcal{C}\setminus c$ denotes all the other  types. This identifies pathways that are uniquely activated within specific cell populations based on the foundation model's learned representations.

\subsection{Spatial Gene Imputation}
\label{suppl:spatial_gene_imputation}
We perform reference-based spatial gene imputation to evaluate the biological fidelity of the learned cell  in zero-shot setting. This task assumes that single-cell and spatial transcriptomics profiles map to a shared latent manifold where proximity implies biological similarity.

{\bf Problem Formulation}
Let $\mathcal{D}_{st}$ and $\mathcal{D}_{sc}$ denote the spatial transcriptomics and single-cell reference datasets, respectively. Let $\mathcal{M}$ be the comprehensive set of genes profiled in the single-cell reference, and $\mathcal{M}'$ be the set of genes captured in the spatial data, such that $\mathcal{M}' \subset \mathcal{M}$.

For an arbitrary sample $i$ from modality $m \in \{sc, st\}$, let $\mathbf{H}^{(m)}_i \in \mathbb{R}^{L_i \times d}$ denote the contextualized hidden states extracted from the last layer of the pretrained foundation model, where $L_i$ represents the sequence length (number of detected genes) and $d$ is the embedding dimension. We derive a global semantic embedding $\mathbf{h}^{(m)}_i \in \mathbb{R}^d$ for this sample by averaging representations across the gene sequence dimension:

\begin{equation}
    \mathbf{h}^{(m)}_i = \frac{1}{L_i} \sum_{t=1}^{L_i} \mathbf{H}^{(m)}_{i, t, :} .
\end{equation}

Formally, we define the spatial dataset as $\mathcal{D}_{st} = \{ (\mathbf{x}^{(st)}_i, \mathbf{h}^{(st)}_i) \}_{i=1}^{N}$, where $\mathbf{h}^{(st)}_i \in \mathbb{R}^d$ is the average semantic embedding for a spot and $\mathbf{x}^{(st)}_i \in \mathbb{R}^{|\mathcal{M}'|}$ is the raw expression vector. Similarly, the reference dataset is defined as $\mathcal{D}_{sc} = \{ (\mathbf{x}^{(sc)}_j, \mathbf{h}^{(sc)}_j) \}_{j=1}^{K}$, where $\mathbf{x}^{(sc)}_j \in \mathbb{R}^{|\mathcal{M}|}$ represents the comprehensive transcriptomic profile. Our objective is to reconstruct the full gene expression profile (specifically for genes in $\mathcal{M} \setminus \mathcal{M}'$ or a hold-out test set) for the spatial spots by leveraging the single-cell reference data.

{\bf Imputation Workflow}
We employ a \textbf{weighted $k$-nearest neighbors} approach. Given a spatial spot $i$, we would like to find the k-nearest cells (which we define as $\mathcal{N}_\text{k}^i$) in the embedding space, where k we set to 50 in our experiment. Then we perform the imputation for the unseen genes ($\mathcal{M} \setminus \mathcal{M}'$) in spatial data using a weighted average of the comprehensive genes ($\mathcal{M}$) in single cell. More details about how we calculate the weights, and the algorithm implementations are illustrated as follows:

\textbf{1. Cosine Similarity Computation:} We compute the cosine similarity score between spatial spot $i$ and reference single cell $j$:
\begin{equation}
    S_{ij} = \frac{\mathbf{h}^{st}_i \cdot \mathbf{h}^{sc}_j}{\|\mathbf{h}^{st}_i\|_2 \|\mathbf{h}^{sc}_j\|_2} .
\end{equation}

\textbf{2. Neighboring Weights Normalization:} For each spatial spot $i$, we identify the set $\mathcal{N}_\text{k}^i$ of the top-k most similar reference cells. We compute normalized weights using the softmax function:
\begin{equation}
    w_{ij} = \frac{\exp(S_{ij})}{\sum_{m \in \mathcal{N}_\text{k}^i} \exp(S_{im})} \quad \forall j \in \mathcal{N}_\text{k}^i .
\end{equation}

\textbf{3. Imputation:} The imputed expression vector $\hat{\mathbf{x}}^{st}_i \in \mathbb{R}^{|M|}$ is calculated as:
\begin{equation}
    \hat{\mathbf{x}}^{st}_i = \sum_{j \in \mathcal{N}_\text{k}^i} w_{ij} \mathbf{x}^{sc}_j .
\end{equation}

This operation effectively transfers the comprehensive gene information from the reference scRNA-seq dataset to the spatial spots.

{\bf Spatial Imputation Datasets} The spatial imputation datasets, their paired scRNA-seq data, and the shared common genes for model evaluation, are provided in ATable \ref{tab:spatial_imputation_dataset}. the single cell reference data of lung cancer and breast cancer were retrieved from \citet{Lung-Kim2020} and \cite{breastCancer-NG-Wu2021} (GEO accession number: GSE131907 and GSE176078 respectively).
The scRNA-seq of brain tissue was downloaded from SEA-AD (donor\_H21\_032) \cite{SEAAD-Gabitto2024}.
The corresponding lung-tissue spatial dataset (\textit{Lung cancer 2}) was obtained from the Vizgen MERFISH FFPE Human Immuno-oncology dataset\footnote{\scriptsize\url{https://info.vizgen.com/ffpe-showcase?submissionGuid=88ba0a44-26e2-47a2-8ee4-9118b9811fbf}}, the paired breast cancer was sourced from donor (CID3586) in \cite{breastCancer-NG-Wu2021} and the brain tissue spatial data was retrieved for donor\_H21\_032 in SEA-AD.

\begin{table}[htbp]
  \centering
  \caption{Summary of spatial transcriptomics (ST) and single-cell (SC) datasets.}
  \label{tab:spatial_imputation_dataset}
  \begin{tabular}{lccc}
    \toprule
    Dataset & ST (spots $\times$ genes) & SC (cells $\times$ genes) &  Shared genes for imputation \\
    \midrule
    Lung cancer   & $31,520 \times 423$   & $207,853 \times 20,793$ & 422 \\
    Breast cancer & $4,784 \times 1,000$  & $9,234 \times 19,098$   & 891 \\
    SEA-AD        & $18,085 \times 180$   & $15,876 \times 20,080$  & 132 \\
    \bottomrule
  \end{tabular}
\end{table}

{\bf Experiments and Evaluation} We first retrieve the shared common gene sets ($\mathcal{M} \cap \mathcal{M}'$) between single cell and spatial transcriptomics data. Then the shared genes are partitioned into 10-fold, we employ k-fold cross-validation to assess the imputed spatial gene performance for each held-out gene sets. Performance is quantified using Pearson Correlation ($r$), Spearman Rank Correlation ($\rho$), and Kendall's Tau ($\tau$) between the imputed gene vector $\hat{\mathbf{x}}$ and ground truth $\mathbf{x}$ across all test spatial spots. the evaluation metrics will be introduced in more detail in Appendix \ref{suppl:correlation_metrics}.

\subsection{Evaluation Metrics}
\label{suppl:evaluation_metrics}
We employ different evaluation metrics depending on the downstream tasks. For zero-shot clustering, we report standard scIB metrics \cite{scIB-Luecken2021}: adjust random index (ARI), normalized mutual information (NMI), average silhouette width (ASW$_{label}$), and their arithmetic mean AvgBio to quantify the separation of different cluster/cell-types. To assess batch mixing, we use ASW$_{label/batch}$ and GraphConn and report their mean (AvgBatch). And the overall scIB score is calculated as a weighted average of these two average metric (overall = 0.6$\times$ AvgBio + 0.4$\times$ AvgBatch).
We also report scGraph-RWR \cite{Wu2025-my} and scGraph-LCA.
The former uses the Random Walk with Restart (RWR) algorithm to obtain cell-type hierarchical distances on the ontology graph and the latter is adapted from scGraph-RWR by replacing the ontology-RWR reference distance matrix with LCA-based distance matrix.
Both metrics quantify graph-level similarity between the reference ontology distance matrix and the cell-type centroid embedding distance matrix, providing a complementary evaluation of the intrinsic biological structure of cell embeddings in scFMs \cite{scGraph-Wang2025}.
For cell type annotation fine-tuning, we compute overall accuracy to measure general performance and report macro F1 to equally weight all classes regardless of imbalance.
For novel cell identification, we compute per-cell-type accuracy and report macro-accuracy to account for class imbalance.
For gene function prediction, we report micro/macro precision and micro/macro F1.
For spatial gene imputation, we quantify the agreement between imputed and ground-truth spatial expression using Pearson, Spearman, and Kendall’s $\tau$ correlation coefficients \cite{scikit-learn}.
The mathematical formulations of each of these metrics are provided in the following sub-sections.

\subsubsection{Zero-shot clustering evaluation metrics}

\textbf{Normalized Mutual Information (NMI):}
\[
\text{NMI}(C, K) = \frac{I(C; K)}{\frac{1}{2}[H(C) + H(K)]} ,
\]
where:
\begin{itemize}
    \item $C$ and $K$ are the true cell type labels and predicted cluster labels.
    \item $H(C) = -\sum_{i} P(c_i) \log P(c_i)$ is the entropy of the labels.
    \item $I(C; K) = \sum_{i,j} P(c_i, k_j) \log \frac{P(c_i, k_j)}{P(c_i)P(k_j)}$ is the mutual information.
\end{itemize}

\textbf{Adjusted Rand Index (ARI):}
\[
\text{ARI} = \frac{\text{RI} - E[\text{RI}]}{\max(\text{RI}) - E[\text{RI}]} ,
\]
where the unadjusted Rand Index (RI) is:
\[
\text{RI} = \frac{TP + TN}{TP + FP + FN + TN} = \frac{a + b}{\binom{N}{2}} ,
\]
\begin{itemize}
    \item $a$: Number of pairs of cells in the same set in $C$ and the same set in $K$.
    \item $b$: Number of pairs of cells in different sets in $C$ and different sets in $K$.
\end{itemize}

\textbf{Average Silhouette Width (ASW$_{label}$) - Cell Type:}
\[
s_i = \frac{b_i - a_i}{\max(a_i, b_i)}, \quad \text{ASW}_{label} = \frac{1}{N} \sum_{i=1}^{N} s_i .
\]
The commonly used version is the rescaled version down below, and we also use the scaled version in our experiment.
\[
\text{ASW}_{label} = \frac{\text{ASW}_{label} + 1}{2} .
\]

\textbf{Average Biological Score (AvgBio):}
\[
\text{AvgBio} = \frac{1}{3} (\text{NMI} + \text{ARI} + \text{ASW}_{label}) .
\]

\textbf{Average Silhouette Width (ASW$_{label/batch}$) - Batch:}
\[
\text{ASW}_{label/batch} = \frac{1}{|C|} \sum_{c \in C} \left[ 1 - \left| \frac{1}{|N_c|} \sum_{i \in N_c} s_i^{\text{batch}} \right| \right] .
\]
where $s_i^{\text{batch}}$ is the silhouette score calculated using batch labels (instead of cell types) within a specific cell type group $c$.

\textbf{Graph Connectivity (GraphConn):}
\[
\text{GraphConn} = \frac{1}{|C|} \sum_{c \in C} \frac{|\text{LCC}(G_c)|}{|N_c|} ,
\]
where:
\begin{itemize}
    \item $G_c$: The k-nearest neighbor (kNN) subgraph containing only cells of type $c$.
    \item $\text{LCC}(G_c)$: The largest connected component in that subgraph.
\end{itemize}

\textbf{Average Batch Score (AvgBatch):}
\[
\text{AvgBatch} = \frac{1}{2} (\text{ASW}_{label/batch} + \text{GraphConn}) .
\]

{\bf Overall scIB score:} $\text{overall score} = 0.6 \times \text{AvgBio} + 0.4 \times \text{AvgBatch}$.

{\bf scGraph-LCA} 
To quantify how well the learned latent embedding preserves the hierarchical relationships defined in the Cell Ontology, we employ a correlation-based graph similarity measurement metric, named scGraph-LCA. Our metric generalizes the evaluation protocols proposed in scGraph \cite{scGraph-Wang2025} and scGraph-RWR \cite{Wu2025-my}. It relies on the comparison of two pairwise distance matrices: one derived from the model's latent space ($\Delta^{\text{emb}}$) and one derived from the ontology graph ($\Delta^{\text{onto}}$).
We will illustrate each component one by one in the following: 

{1. Model embedding distance matrix ($\Delta^{\text{emb}}$)}

Following the notation in Section \ref{sec:data_process} in the main paper, here, let $\mathcal{D} = \{x_n, c_n\}_{n=0}^N$ denotes a single-cell RNA-seq dataset, where each cell $n$ is associated with a cell type label $c_n \in \mathcal{V}$. We reuse the notation of Cell-Dataset Index ($\mathcal{I}_{c, k}$) here, let $\mathcal{I}_c = \bigcup_{k \in \mathcal{K}_c} \mathcal{I}_{c,k}$ represent the set of indices for all cells annotated as type $c \in \mathcal{V}$ across all available datasets/batches, here $k$ could interchangeably represent as batches, since if we only evaluate on one dataset, the number of available dataset is always being 1, however, there may exist multiple batches in one dataset.

For each unique cell type $c \in \mathcal{V}$ present in the data, we compute the centroid in the embedding space. To mitigate the effect of outliers, we employ a trimmed mean aggregation. Let $\{\mathbf{z}_n \mid n \in \mathcal{I}_c\}$ be the set of embedding vectors for all cells of type $c$. We compute the trimmed mean centroid $\boldsymbol{\mu}_c$:
\begin{equation}
    \boldsymbol{\mu}_c = \text{TrimmedMean}\left( \{\mathbf{z}_n \mid n \in \mathcal{I}_c\}, \alpha \right),
\end{equation}
where $\alpha$ is the trimming proportion (set to 0.2 in our experiments).

We then construct the pairwise \textbf{Embedding Distance Matrix} $\Delta^{\text{emb}} \in \mathbb{R}^{|\mathcal{V}| \times |\mathcal{V}|}$, where the entry $(i, j)$ represents the Euclidean distance between the centroids of cell types $c_i$ and $c_j$:
\begin{equation}
    \Delta^{\text{emb}}_{ij} = \|\boldsymbol{\mu}_{c_i} - \boldsymbol{\mu}_{c_j}\|_2.
\end{equation}

{2. Ontology Reference Distance Matrix ($\mathbf{\Delta}^{\text{onto}}$)}

The reference matrix encodes the "ground truth" biological distances extracted from the Cell Ontology graph $\mathcal{G} = (\mathcal{V}, \mathcal{E})$. , we first introduce the LCA distance matrix, then in the following we will derive the scGraph-RWR based on RWR distancem matrix.

As introduced in Section \ref{sec:data_process}, we utilize the Lowest Common Ancestor (LCA) to capture the hierarchical distances on the ontology graph. For any two types of cells $c_i, c_j \in \mathcal{V}$, their raw semantic distance $d_{\mathcal{G}}(c_i, c_j)$ is calculated as the sum of the shortest path distances from each node to their common $c_s = \text{LCA}(c_i, c_j)$:
\begin{equation}
    \Delta^{\text{onto, LCA}}_{ij} = \delta_{\mathcal{G}}(c_i, c_j) = \delta_{path}(c_i, c_s) + \delta_{path}(c_j, c_s) ,
\end{equation}
where $\delta_{path}(c_i, c_s)$ and $\delta_{path}(c_j, c_s)$ denote the number of edges on the shortest path between nodes $(c_i, c_j)$ to their lowest common ancestor $c_s$ in $\mathcal{G}$, here we use $\Delta^{\text{onto, LCA}}_{ij}$ to distinguish the RWR-based distance matrix ($\Delta^{\text{onto, RWR}}_{ij}$) in scGraph-RWR.

{3. Graph Similarity Measurement}

To quantify the similarity between the model embedding distance matrix and ontology-based distance matrix, we calculate Pearson Correlation and the rank-based correlation (Spearman, and Kendall's $\tau$) between the columns of the distance matrices. For each cell type $c_i$, let $\mathbf{r}^{\text{emb}}_i$ be the $i$-th column of $\Delta^{\text{emb, LCA}}$ and $\mathbf{r}^{\text{onto}}_i$ be the $i$-th column of $\Delta^{\text{onto, LCA}}$. We compute the agreement score using Pearson Correlation ($r$), Spearman's rank correlation ($\rho$) and Kendall's rank correlation coefficient ($\tau$), the full mathematical equations of the above metrics are illustrated in more detail in Appendix \ref{suppl:correlation_metrics}:

\begin{equation}
    \text{scGraph-LCA}_{r} = \frac{1}{K} \sum_{i=1}^K \rho(\mathbf{r}^{\text{emb}}_i, \mathbf{r}^{\text{onto}}_i) ,
\end{equation}

\begin{equation}
    \text{scGraph-LCA}_{\rho} = \frac{1}{K} \sum_{i=1}^K \rho(\mathbf{r}^{\text{emb}}_i, \mathbf{r}^{\text{onto}}_i) ,
\end{equation}

\begin{equation}
    \text{scGraph-LCA}_{\tau} = \frac{1}{K} \sum_{i=1}^K \tau(\mathbf{r}^{\text{emb}}_i, \mathbf{r}^{\text{onto}}_i) .
\end{equation}

A higher correlation indicates that cell types close in the ontology graph are also geometrically proximal in the model's latent space, demonstrating preservation of biological semantics.

{\bf scGraph-RWR} According to \cite{Wu2025-my}, scGraph-RWR utilizes a weighted ontology graph, in our experiment the weights are download from \cite{onClass-Wang2021}. Regarding the reference distance matrix, we just need to replace the LCA-based ontology distance matrix with the RWR-based ontology distance matrix by following scGraph-LCA procedures. The process to get $\Delta^{\text{onto, RWR}}$ is as follows: 

First, we define the column-normalized transition probability matrix $\mathbf{P}$ as:
\begin{equation}
    \mathbf{P} = \mathbf{A} \mathbf{D}^{-1},
\end{equation}
where $\mathbf{A}$ is the adjacency matrix of ontology graph $\mathcal{G}$ and $\mathbf{D}$ is the diagonal degree matrix. The RWR diffusion matrix $\mathbf{R}$, which captures the stationary probability distribution for walkers starting at each node, is computed analytically as:
\begin{equation}
    \mathbf{R} = \beta \left( \mathbf{I} - (1 - \beta) \mathbf{P} \right)^{-1},
\end{equation}
where $\beta$ is the restart probability (regulating the diffusion scale) and $\mathbf{I}$ is the identity matrix. Finally, the ontology-derived RWR distance between cell types $i$ and $j$ is defined as the complement of the diffusion probability:
\begin{equation}
    \Delta^{\text{onto, RWR}}_{ij} = 1 - R_{ij}.
\end{equation}

\subsubsection{Cell annotation (Fine-tuning) metrics}
Let $C$ be the complete set of unique cell types present in the evaluation dataset. 
The overall classification performance is evaluated using Accuracy and Macro F1 score.
The \textbf{Accuracy} is defined as the overall proportion of correctly classified cells out of the total number of cells:
\[
\text{Accuracy} = \frac{\sum_{c \in C} TP_c}{N} ,
\]
where:
\begin{itemize}
    \item $TP_c$: The number of cells correctly classified as cell type $c$ (True Positives).
    \item $N$: The total number of cells across all types in the evaluation dataset.
\end{itemize}
The \textbf{Macro F1} score is defined as the arithmetic mean of the individual F1 scores computed independently for each cell type $c \in C$, treating all classes equally regardless of their support size:
\[
\text{Macro F1} = \frac{1}{|C|} \sum_{c \in C} \text{F1}_c ,
\]
where $\text{F1}_c$ is the harmonic mean of the Precision and Recall for a specific cell type $c$:
\[
\text{F1}_c = 2 \times \frac{\text{Precision}_c \times \text{Recall}_c}{\text{Precision}_c + \text{Recall}_c} ,
\]
with Precision and Recall defined as:
\[
\text{Precision}_c = \frac{TP_c}{TP_c + FP_c} , \quad \text{Recall}_c = \frac{TP_c}{TP_c + FN_c} ,
\]
\begin{itemize}
    \item $FP_c$: The number of cells incorrectly classified as cell type $c$ (False Positives).
    \item $FN_c$: The number of cells truly belonging to cell type $c$ but incorrectly classified as another type (False Negatives).
    \item $|C|$: The total count of unique cell types in the evaluation set.
\end{itemize}

\subsubsection{Novel cell identification metrics}
Let $U \subset V$ be the set of unseen cell types (where $V$ is the total set of cell ontology nodes).
The Macro Accuracy over the unseen set $U$ is defined as the arithmetic mean of the recall (accuracy) for each unseen class:

\[
\text{Macro Accuracy}_U = \frac{1}{|U|} \sum_{u \in U} \text{Acc}_u ,
\]

where $\text{Acc}_u$ is the accuracy for a specific cell type $u$:

\[
\text{Acc}_u = \frac{TP_u}{N_u} ,
\]

\begin{itemize}
    \item $TP_u$: The number of cells correctly classified as type $u$ (True Positives).
    \item $N_u$: The total number of cells in the dataset with true label $u$ (Support).
    \item $|U|$: The count of unique unseen cell types.
\end{itemize}

\subsubsection{Gene function prediction}
\label{suppl:gene_func_metrics}

We evaluate the multi-label GO function classification performance using Micro-averaged and Macro-averaged metrics. Let $C$ be the total number of classes (in which we have three broad GO terms: Biological Process
(BP), Cellular Component (CC), and Molecular Function
(MF)). For each specific class $i \in \{1, \dots, C\}$, we define:
\begin{itemize}
    \item $TP_i$: True Positives (samples correctly predicted to have GO term $i$).
    \item $FP_i$: False Positives (samples incorrectly predicted to have GO term $i$).
    \item $FN_i$: False Negatives (samples with GO term $i$ that were missed).
\end{itemize}

{\bf Macro-Averaged Metrics}
Macro metrics calculate performance for each GO class independently and then take the unweighted average. This ensures that specific, rare GO class contribute equally to the score as broad, common terms.

\[
\text{Precision}_{\text{macro}} = \frac{1}{C} \sum_{i=1}^{C} \frac{TP_i}{TP_i + FP_i} ,
\]

\[
\text{Recall}_{\text{macro}} = \frac{1}{C} \sum_{i=1}^{C} \frac{TP_i}{TP_i + FN_i} ,
\]

\[
\text{F1}_{\text{macro}} = \frac{1}{C} \sum_{i=1}^{C} \frac{2 \times TP_i}{2 \times TP_i + FP_i + FN_i} .
\]

{\bf Micro-Averaged Metrics}
Micro metrics aggregate the prediction counts across all GO class globally before calculating the score. This gives a measure of the overall system reliability, weighted by the frequency of the terms.

\[
\text{Precision}_{\text{micro}} = \frac{\sum_{i=1}^{C} TP_i}{\sum_{i=1}^{C} (TP_i + FP_i)} ,
\]

\[
\text{Recall}_{\text{micro}} = \frac{\sum_{i=1}^{C} TP_i}{\sum_{i=1}^{C} (TP_i + FN_i)} ,
\]

\[
\text{F1}_{\text{micro}} = \frac{2 \times \text{Precision}_{\text{micro}} \times \text{Recall}_{\text{micro}}}{\text{Precision}_{\text{micro}} + \text{Recall}_{\text{micro}}} .
\]

\subsubsection{Spatial imputation evaluation metrics}
\label{suppl:correlation_metrics}

{\bf Pearson Correlation ($r$)}
\[
r = \frac{\sum (x_i - \bar{x})(y_i - \bar{y})}{\sqrt{\sum (x_i - \bar{x})^2} \sqrt{\sum (y_i - \bar{y})^2}} .
\]

{\bf Spearman's Rank Correlation ($\rho$)}
\[
\rho = 1 - \frac{6 \sum d_i^2}{n(n^2 - 1)} .
\]
where $d_i = \text{rank}(x_i) - \text{rank}(y_i)$.

{\bf Kendall's Tau ($\tau$)}
\[
\tau = \frac{C - D}{\frac{1}{2} n (n-1)} .
\]
where $C$ is the number of concordant pairs and $D$ is the number of discordant pairs.

\section{RESULTS}
\label{suppl:results}
\subsection{Zero-shot clustering SOTA benchmark performance of scIB metrics}
\label{suppl:scIB-zero_shot}

\subsubsection{Overall scIB score across the evaluation tissues} The overall of the scIB scores for each dataset are provided in ATable~\ref{tab:scIB_overall_tissue_performance}.

\begin{table}[h]
\centering
\caption{Performance comparison of zero-shot clustering scIB overall score across tissues/datasets. The best results are shown in \textbf{bold} for each dataset.}
\label{tab:scIB_overall_tissue_performance}
\begin{tabular}{lcccccc}
\toprule
Dataset & Ours & Geneformer & scGPT & scCello & CellPLM & SCimilarity \\
\midrule
HIHA
& 0.7533
& 0.4444
& 0.6002
& 0.5668
& 0.7037
& \textbf{0.7730} \\

Heart
& 0.8324
& 0.6862
& 0.8326
& \textbf{0.8666}
& 0.7847
& 0.8299 \\

Pancreas
& \textbf{0.8140}
& 0.6113
& 0.7967
& 0.7203
& 0.7024
& 0.8023 \\

Lung
& 0.8467
& 0.8395
& 0.8397
& 0.8299
& 0.8379
& \textbf{0.8566} \\

Liver
& 0.7730
& 0.6853
& 0.7886
& 0.7720
& 0.7102
& \textbf{0.7921} \\

Breast TNBC
& \textbf{0.8724}
& 0.6824
& 0.8418
& 0.8423
& 0.8377
& 0.8601 \\

Kidney
& 0.7185
& 0.5785
& 0.7020
& \textbf{0.7601}
& 0.7211
& 0.7149 \\

Adipose Tissue
& \textbf{0.8696}
& 0.6509
& 0.8181
& 0.8589
& 0.8064
& 0.8290 \\

Bone Marrow
& 0.6823
& 0.6498
& 0.6924
& 0.6659
& 0.6772
& \textbf{0.7059} \\

\midrule
Mean
& 0.7958
& 0.6476
& 0.7680
& 0.7647
& 0.7535
& \textbf{0.7960} \\
\bottomrule
\end{tabular}
\end{table}

\subsubsection{Details of each scIB score across the evaluation tissues} For the details of each scIB metrics: ARI, NMI, ASW$_{label}$, GraphConn, ASW$_{label/batch}$, AvgBio, AvgBatch, in HIHA, Heart, Pancreas, Lung, Liver, Breast\_TNBC, Kidney, Adipose Tissue, Bone Marrow. Please refer to the ATable \ref{tab:scIB_full_benchmark_results}.
\begin{table}[h]
\centering
\caption{Zero-shot clustering benchmarking results of detailed scIB metrics for each of the nine evaluation datasets. The best performance in each metric for each dataset is highlighted in \textbf{bold}.}
\label{tab:scIB_full_benchmark_results}
\resizebox{\textwidth}{!}{%
\begin{tabular}{llcccccccc}
\toprule
Dataset & Method & NMI & ARI & ASW$_{label}$ & GraphConn & ASW$_{label/batch}$ & AvgBio & AvgBatch & Overall \\
\midrule
\multirow{6}{*}{HIHA}
 & Ours & 0.7597 & 0.6422 & \textbf{0.6021} & 0.8868 & 0.8756 & 0.6680 & 0.8812 & 0.7533 \\
 & Geneformer & 0.2500 & 0.1355 & 0.4560 & 0.5117 & 0.8687 & 0.2805 & 0.6902 & 0.4444 \\
 & scGPT & 0.5199 & 0.2619 & 0.4977 & 0.8218 & 0.8995 & 0.4265 & 0.8607 & 0.6002 \\
 & scCello & 0.4704 & 0.3245 & 0.4824 & 0.6418 & \textbf{0.9149} & 0.4257 & 0.7784 & 0.5668 \\
 & cellPLM & 0.7101 & 0.4755 & 0.5365 & \textbf{0.9208} & 0.8755 & 0.5740 & \textbf{0.8982} & 0.7037 \\
 & SCimilarity & \textbf{0.7675} & \textbf{0.7375} & 0.5833 & 0.9167 & 0.8602 & \textbf{0.6961} & 0.8884 & \textbf{0.7730} \\
\midrule
\multirow{6}{*}{Heart}
 & Ours & 0.8581 & 0.8204 & \textbf{0.6760} & 0.9297 & 0.8780 & 0.7848 & 0.9039 & 0.8324 \\
 & Geneformer & 0.6464 & 0.4900 & 0.5228 & 0.8878 & 0.8838 & 0.5531 & 0.8858 & 0.6862 \\
 & scGPT & 0.8641 & 0.8905 & 0.5911 & 0.9371 & 0.8803 & 0.7819 & 0.9087 & 0.8326 \\
 & scCello & \textbf{0.9036} & \textbf{0.9383} & 0.6546 & \textbf{0.9456} & \textbf{0.8907} & \textbf{0.8321} & \textbf{0.9182} & \textbf{0.8666} \\
 & cellPLM & 0.8168 & 0.7228 & 0.5891 & 0.9324 & 0.8627 & 0.7096 & 0.8975 & 0.7847 \\
 & SCimilarity & 0.8572 & 0.8182 & 0.6715 & 0.9402 & 0.8626 & 0.7823 & 0.9014 & 0.8299 \\
\midrule
\multirow{6}{*}{Pancreas}
 & Ours & \textbf{0.8598} & 0.7951 & 0.6092 & \textbf{0.9799} & 0.8259 & \textbf{0.7547} & 0.9029 & \textbf{0.8140} \\
 & Geneformer & 0.5045 & 0.3187 & 0.5082 & 0.8726 & 0.8525 & 0.4438 & 0.8625 & 0.6113 \\
 & scGPT & 0.8315 & \textbf{0.8523} & 0.5259 & 0.9199 & 0.8539 & 0.7365 & 0.8869 & 0.7967 \\
 & scCello & 0.6442 & 0.5420 & 0.5782 & 0.9616 & \textbf{0.8753} & 0.5881 & \textbf{0.9185} & 0.7203 \\
 & cellPLM & 0.6277 & 0.5578 & 0.5515 & 0.9247 & 0.8504 & 0.5790 & 0.8875 & 0.7024 \\
 & SCimilarity & 0.8497 & 0.7252 & \textbf{0.6568} & 0.9522 & 0.8276 & 0.7439 & 0.8899 & 0.8023 \\
\midrule
\multirow{6}{*}{Lung}
 & Ours & 0.8722 & 0.8696 & 0.6469 & 0.9797 & 0.8653 & 0.7962 & 0.9225 & 0.8467 \\
 & Geneformer & 0.8797 & 0.8512 & 0.5774 & 0.9775 & \textbf{0.9115} & 0.7694 & 0.9445 & 0.8395 \\
 & scGPT & 0.8813 & 0.8519 & 0.5736 & \textbf{0.9955} & 0.8965 & 0.7689 & \textbf{0.9460} & 0.8397 \\
 & scCello & 0.8664 & 0.8043 & 0.6211 & 0.9781 & 0.8794 & 0.7640 & 0.9287 & 0.8299 \\
 & cellPLM & 0.8823 & 0.8477 & 0.5786 & 0.9942 & 0.8867 & 0.7696 & 0.9404 & 0.8379 \\
 & SCimilarity & \textbf{0.8898} & \textbf{0.8733} & \textbf{0.6820} & 0.9885 & 0.8495 & \textbf{0.8150} & 0.9190 & \textbf{0.8566} \\
\midrule
\multirow{6}{*}{Liver}
 & Ours & 0.7871 & 0.6935 & 0.5824 & 0.9600 & 0.8422 & 0.6877 & 0.9011 & 0.7730 \\
 & Geneformer & 0.6502 & 0.4678 & 0.5230 & 0.9366 & 0.8486 & 0.5470 & 0.8926 & 0.6853 \\
 & scGPT & \textbf{0.7933} & \textbf{0.7549} & 0.5761 & 0.9625 & 0.8564 & 0.7081 & 0.9095 & 0.7886 \\
 & scCello & 0.7674 & 0.6763 & 0.5746 & 0.9659 & \textbf{0.8756} & 0.6728 & \textbf{0.9207} & 0.7720 \\
 & cellPLM & 0.6738 & 0.4850 & 0.5639 & \textbf{0.9742} & 0.8542 & 0.5742 & 0.9142 & 0.7102 \\
 & SCimilarity & 0.7920 & 0.7154 & \textbf{0.6461} & 0.9730 & 0.8340 & \textbf{0.7179} & 0.9035 & \textbf{0.7921} \\
\midrule
\multirow{6}{*}{Breast TNBC}
 & Ours & \textbf{0.8807} & \textbf{0.9090} & 0.7245 & 0.9954 & 0.8527 & \textbf{0.8380} & 0.9241 & \textbf{0.8724} \\
 & Geneformer & 0.6080 & 0.3939 & 0.5279 & 0.9905 & 0.8917 & 0.5099 & 0.9411 & 0.6824 \\
 & scGPT & 0.8543 & 0.8528 & 0.6070 & 0.9941 & \textbf{0.9008} & 0.7714 & \textbf{0.9475} & 0.8418 \\
 & scCello & 0.8316 & 0.8014 & 0.7106 & 0.9895 & 0.8785 & 0.7812 & 0.9340 & 0.8423 \\
 & cellPLM & 0.8215 & 0.8444 & 0.6438 & 0.9929 & 0.8858 & 0.7699 & 0.9393 & 0.8377 \\
 & SCimilarity & 0.8596 & 0.8616 & \textbf{0.7267} & \textbf{0.9963} & 0.8565 & 0.8160 & 0.9264 & 0.8601 \\
\midrule
\multirow{6}{*}{Kidney}
 & Ours & 0.7174 & 0.5711 & 0.5871 & 0.9542 & 0.7627 & 0.6252 & 0.8585 & 0.7185 \\
 & Geneformer & 0.4874 & 0.2008 & 0.4878 & 0.9392 & 0.7772 & 0.3920 & 0.8582 & 0.5785 \\
 & scGPT & 0.7111 & 0.5489 & 0.5331 & 0.9574 & 0.7597 & 0.5977 & 0.8586 & 0.7020 \\
 & scCello & 0.7452 & \textbf{0.6851} & 0.6009 & 0.9641 & \textbf{0.8050} & \textbf{0.6771} & \textbf{0.8846} & \textbf{0.7601} \\
 & cellPLM & 0.7201 & 0.6085 & 0.5357 & \textbf{0.9771} & 0.7643 & 0.6215 & 0.8707 & 0.7211 \\
 & SCimilarity & \textbf{0.7566} & 0.5347 & \textbf{0.6116} & 0.9646 & 0.7072 & 0.6343 & 0.8359 & 0.7149 \\
\midrule
\multirow{6}{*}{Adipose Tissue}
 & Ours & \textbf{0.9098} & \textbf{0.9307} & 0.6411 & 0.9626 & 0.9037 & \textbf{0.8272} & 0.9331 & \textbf{0.8696} \\
 & Geneformer & 0.6091 & 0.3070 & 0.5012 & 0.9390 & 0.8984 & 0.4724 & 0.9187 & 0.6509 \\
 & scGPT & 0.8618 & 0.7557 & 0.5944 & 0.9710 & 0.9074 & 0.7373 & 0.9392 & 0.8181 \\
 & scCello & 0.8786 & 0.8991 & 0.6337 & 0.9580 & \textbf{0.9250} & 0.8038 & \textbf{0.9415} & 0.8589 \\
 & cellPLM & 0.8151 & 0.7359 & 0.6120 & \textbf{0.9775} & 0.8915 & 0.7210 & 0.9345 & 0.8064 \\
 & SCimilarity & 0.8639 & 0.7567 & \textbf{0.6868} & 0.9685 & 0.8692 & 0.7691 & 0.9188 & 0.8290 \\
\midrule
\multirow{6}{*}{Bone Marrow}
 & Ours & 0.6809 & 0.3919 & 0.5135 & 0.9650 & 0.8601 & 0.5288 & 0.9126 & 0.6823 \\
 & Geneformer & 0.6025 & 0.3237 & 0.4939 & 0.9364 & 0.8925 & 0.4734 & 0.9144 & 0.6498 \\
 & scGPT & 0.6654 & 0.4002 & 0.5267 & 0.9693 & \textbf{0.9005} & 0.5307 & \textbf{0.9349} & 0.6924 \\
 & scCello & 0.6256 & 0.3867 & 0.5075 & 0.9484 & 0.8614 & 0.5066 & 0.9049 & 0.6659 \\
 & cellPLM & 0.6245 & 0.3884 & 0.5260 & 0.9653 & 0.8816 & 0.5130 & 0.9235 & 0.6772 \\
 & SCimilarity & \textbf{0.7160} & \textbf{0.4321} & \textbf{0.5406} & \textbf{0.9707} & 0.8703 & \textbf{0.5629} & 0.9205 & \textbf{0.7059} \\
\bottomrule
\end{tabular}%
}
\end{table}

\subsubsection{Zero-shot clustering of scGraph-LCA benchmarking performance (Table \ref{tab:scGraph-LCA_metrics})} 

\begin{table}[h]
\centering
\caption{Dataset-wise comparison of zero-shot clustering scGraph-LCA metric under Spearman, Kendall's $\tau$, and Pearson correlation. Best result per dataset and per metric is shown in \textbf{bold}.}
\label{tab:scGraph-LCA_metrics}
\resizebox{\textwidth}{!}{%
\begin{tabular}{l*{18}{c}}
\toprule
& \multicolumn{6}{c}{Spearman}
& \multicolumn{6}{c}{Kendall's $\tau$}
& \multicolumn{6}{c}{Pearson} \\
\cmidrule(lr){2-7} \cmidrule(lr){8-13} \cmidrule(lr){14-19}
Dataset
& Ours & Geneformer & scGPT & scCello & CellPLM & SCimilarity
& Ours & Geneformer & scGPT & scCello & CellPLM & SCimilarity
& Ours & Geneformer & scGPT & scCello & CellPLM & SCimilarity \\
\midrule
HIHA
& \textbf{0.7157} & 0.3890 & 0.4174 & 0.4724 & 0.3772 & 0.6201
& \textbf{0.5688} & 0.2849 & 0.3171 & 0.3590 & 0.2895 & 0.4921
& \textbf{0.7669} & 0.4527 & 0.5300 & 0.5946 & 0.5215 & 0.7294 \\

Heart
& \textbf{0.7074} & 0.4220 & 0.6605 & 0.4163 & 0.5100 & 0.5090
& \textbf{0.5783} & 0.3477 & 0.5275 & 0.3277 & 0.3929 & 0.4117
& 0.7787 & 0.6073 & \textbf{0.7914} & 0.6462 & 0.6969 & 0.6909 \\

Pancreas
& 0.3992 & 0.4040 & 0.3978 & 0.3313 & 0.5470 & \textbf{0.6646}
& 0.2960 & 0.3213 & 0.3135 & 0.2390 & 0.4454 & \textbf{0.5699}
& 0.7398 & 0.7139 & 0.7660 & 0.6478 & 0.7780 & \textbf{0.8412} \\

Lung
& \textbf{0.6777} & 0.6719 & 0.4047 & 0.4655 & 0.4624 & 0.5958
& 0.5615 & \textbf{0.5621} & 0.3270 & 0.3684 & 0.3745 & 0.4890
& \textbf{0.8381} & 0.8127 & 0.7396 & 0.7519 & 0.7474 & 0.7886 \\

Liver
& \textbf{0.5704} & 0.4905 & 0.4824 & 0.4925 & 0.5023 & 0.4842
& \textbf{0.4691} & 0.3857 & 0.3799 & 0.3923 & 0.4008 & 0.3781
& \textbf{0.7411} & 0.6457 & 0.6423 & 0.6718 & 0.6716 & 0.6501 \\

Breast TNBC
& 0.5112 & 0.4000 & 0.4328 & \textbf{0.5969} & 0.5487 & 0.4226
& 0.3956 & 0.3261 & 0.3470 & \textbf{0.5085} & 0.4390 & 0.3461
& 0.8699 & 0.8298 & 0.8774 & \textbf{0.8808} & 0.8693 & 0.8425 \\

Kidney
& \textbf{0.6378} & 0.4423 & 0.2760 & 0.3554 & 0.3912 & 0.5204
& \textbf{0.5083} & 0.3368 & 0.1921 & 0.2573 & 0.2878 & 0.3910
& \textbf{0.6857} & 0.5748 & 0.4880 & 0.5579 & 0.5382 & 0.6273 \\

Adipose Tissue
& 0.6180 & 0.4910 & 0.4144 & 0.5398 & 0.4644 & \textbf{0.6355}
& 0.4762 & 0.3874 & 0.3267 & 0.4051 & 0.3606 & \textbf{0.5030}
& 0.7830 & 0.6737 & 0.6450 & 0.7102 & 0.6617 & \textbf{0.7834} \\

Bone Marrow
& \textbf{0.6630} & 0.3306 & 0.2431 & 0.5207 & 0.2891 & 0.5980
& \textbf{0.5370} & 0.2470 & 0.1794 & 0.3984 & 0.2202 & 0.4749
& \textbf{0.7093} & 0.4151 & 0.4120 & 0.6115 & 0.4437 & 0.6849 \\

\midrule
Mean
& \textbf{0.6112} & 0.4490 & 0.4143 & 0.4656 & 0.4547 & 0.5611
& \textbf{0.4879} & 0.3554 & 0.3233 & 0.3618 & 0.3568 & 0.4506
& \textbf{0.7681} & 0.6362 & 0.6546 & 0.6747 & 0.6587 & 0.7376 \\
\bottomrule
\end{tabular}%
}
\end{table}

\subsubsection{Zero-shot clustering of scGraph-RWR benchmarking performance(Table \ref{tab:scGraph-RWR_performance})} 

\begin{table}[t]
\centering
\caption{Performance comparison across datasets of scGraph-RWR. Best result per dataset and best mean performance are shown in \textbf{bold}.}
\label{tab:scGraph-RWR_performance}
\resizebox{\textwidth}{!}{
\begin{tabular}{lccccccccccccccc}
\toprule
& \multicolumn{5}{c}{Spearman} 
& \multicolumn{5}{c}{Kendall's $\tau$} 
& \multicolumn{5}{c}{Pearson} \\
\cmidrule(lr){2-6} \cmidrule(lr){7-11} \cmidrule(lr){12-16}
Dataset
& Ours & Geneformer & scGPT & scCello & CellPLM
& Ours & Geneformer & scGPT & scCello & CellPLM
& Ours & Geneformer & scGPT & scCello & CellPLM \\
\midrule
HIHA
& 0.4939 & 0.4584 & 0.5000 & \textbf{0.5253} & 0.4951
& 0.3668 & 0.3292 & 0.3677 & \textbf{0.3802} & 0.3618
& \textbf{0.7998} & 0.6188 & 0.6230 & 0.6855 & 0.7118 \\

Heart
& 0.3438 & 0.4141 & 0.3190 & \textbf{0.4372} & 0.3950
& 0.2264 & \textbf{0.3223} & 0.2132 & \textbf{0.3223} & 0.3058
& 0.8926 & 0.8406 & 0.8857 & \textbf{0.9146} & 0.9022 \\

Pancreas
& 0.5373 & 0.5693 & 0.3700 & \textbf{0.6562} & 0.5029
& 0.4116 & 0.4697 & 0.2980 & \textbf{0.5202} & 0.4040
& 0.8831 & 0.8923 & 0.8612 & \textbf{0.9007} & 0.8579 \\

Lung
& \textbf{0.5370} & 0.3794 & 0.3297 & 0.2933 & 0.2970
& \textbf{0.4489} & 0.2756 & 0.2311 & 0.2400 & 0.2133
& 0.9285 & 0.8861 & 0.9153 & \textbf{0.9513} & 0.9037 \\

Liver
& \textbf{0.4974} & 0.2292 & 0.2393 & 0.3665 & 0.3649
& \textbf{0.3771} & 0.1677 & 0.1594 & 0.2552 & 0.2573
& 0.8525 & 0.7604 & 0.8307 & \textbf{0.8898} & 0.8258 \\

Breast TNBC
& \textbf{0.6200} & 0.6000 & 0.5400 & 0.4800 & 0.3800
& 0.4800 & \textbf{0.5200} & 0.4400 & 0.4400 & 0.2800
& 0.9601 & 0.9702 & \textbf{0.9793} & 0.9639 & 0.9464 \\

Kidney
& 0.4045 & \textbf{0.5174} & 0.4825 & 0.4294 & 0.4472
& 0.2823 & \textbf{0.3801} & 0.3489 & 0.3120 & 0.3202
& 0.7419 & 0.7105 & 0.7917 & \textbf{0.8268} & 0.7128 \\

Adipose Tissue
& \textbf{0.5853} & 0.4151 & 0.4574 & 0.4410 & 0.3744
& \textbf{0.4458} & 0.3073 & 0.3375 & 0.3250 & 0.2646
& 0.8354 & 0.7821 & 0.8654 & \textbf{0.8688} & 0.8113 \\

Bone Marrow
& \textbf{0.5563} & 0.3358 & 0.2973 & 0.4434 & 0.3810
& \textbf{0.4040} & 0.2387 & 0.2167 & 0.3151 & 0.2820
& \textbf{0.7476} & 0.5311 & 0.6396 & 0.7473 & 0.6543 \\

\midrule
Mean
& \textbf{0.5084} & 0.4354 & 0.3928 & 0.4525 & 0.4042
& \textbf{0.3825} & 0.3345 & 0.2903 & 0.3456 & 0.2988
& 0.8491 & 0.7769 & 0.8213 & \textbf{0.8610} & 0.8140 \\
\bottomrule
\end{tabular}
}
\end{table}

\subsubsection{Zero-shot clustering of 340 in-distribution cell types in unseen datasets} We report the AvgBio metrics (ARI, NMI, ASW$_{label}$) in scIB (ATabel \ref{tab:seenCL_new_scib_scgraphlca_multicol}), since we randomly down-sample the cell types with more than 10,000 cells, it is hard for us to observe the batch/dataset id in this process, for simplicity of the implementation, we only calculate the AvgBio related metrics. Additionally, we also report scGraph-LCA and scGraph-RWR in ATabel \ref{tab:seenCL_new_scib_scgraphlca_multicol}. Note that for the scGraph-RWR metric, 48 cell types are not present in reference embedding, so the total cell types for this metric is 292.
\begin{table}[h]
\centering
\caption{Overall performance of zero-shot clustering of 340 in-distribution cell types in unseen datasets across SOTA methods using scIB metrics, scGraph-LCA, and scGraph-RWR. Best result per column is shown in \textbf{bold}.}
\label{tab:seenCL_new_scib_scgraphlca_multicol}
\resizebox{\textwidth}{!}{
\begin{tabular}{lcccc ccc ccc}
\toprule
& \multicolumn{4}{c}{scIB Metrics}
& \multicolumn{3}{c}{scGraph-LCA}
& \multicolumn{3}{c}{scGraph-RWR} \\
\cmidrule(lr){2-5} 
\cmidrule(lr){6-8}
\cmidrule(lr){9-11}
Method 
& NMI & ARI & ASW & Avg Bio
& Spearman & Kendall's $\tau$ & Pearson
& Spearman & Kendall's $\tau$ & Pearson \\
\midrule
Ours
& 0.7804 & \textbf{0.3431} & 0.5032 & 0.5422
& \textbf{0.5114} & \textbf{0.3887} & \textbf{0.5611}
& 0.0937 & 0.0569 & 0.4427 \\

Geneformer
& 0.7213 & 0.2241 & 0.4717 & 0.4724
& 0.2962 & 0.2157 & 0.3278
& 0.1369 & 0.0916 & 0.3605 \\

scGPT
& 0.7740 & 0.3014 & 0.5074 & 0.5276
& 0.1339 & 0.0962 & 0.2058
& 0.0682 & 0.0447 & 0.3638 \\

CellPLM
& \textbf{0.7915} & 0.3334 & \textbf{0.5243} & \textbf{0.5498}
& 0.0980 & 0.0687 & 0.1933
& \textbf{0.1436} & \textbf{0.0962} & 0.3980 \\

scCello
& 0.7693 & 0.3092 & 0.5199 & 0.5328
& 0.2417 & 0.1751 & 0.3384
& 0.0670 & 0.0427 & \textbf{0.4768} \\
\bottomrule
\end{tabular}
}
\end{table}

\subsubsection{Bone marrow clustering and graph distance visualization}
\begin{figure}[h]
\centering
\includegraphics[width=\columnwidth]{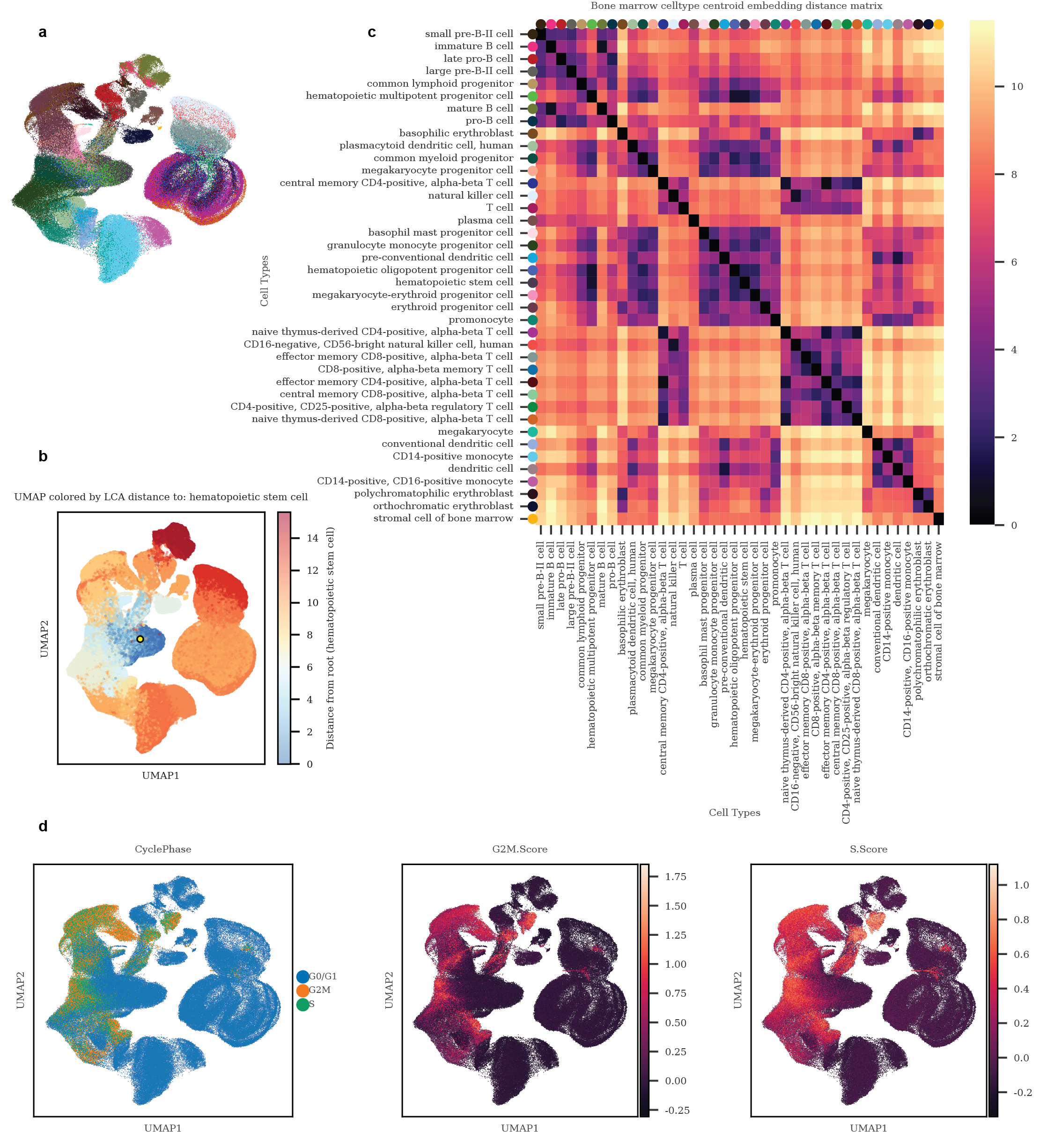}
\caption{(a) Clustering results of Bone marrow dataset using our model embeddings, colors represent different cell-types. (b) Distance heatmap from root centroid cell-type (hematopoietic stem cell, highlighted as \textcolor{yellow}{yellow dot} in the umap) to all the other cell-types in the dataset. (c) Cell-type centroid embedding distances heatmap for the bone marrow dataset. (d) Cell cycle scores heatmap from original paper \cite{BoneMarrow-Zeng2025}.}
\label{fig:BoneMarrow_emb_heatmap}
\end{figure}

\begin{figure}[h]
\centering
\includegraphics[width=\columnwidth]{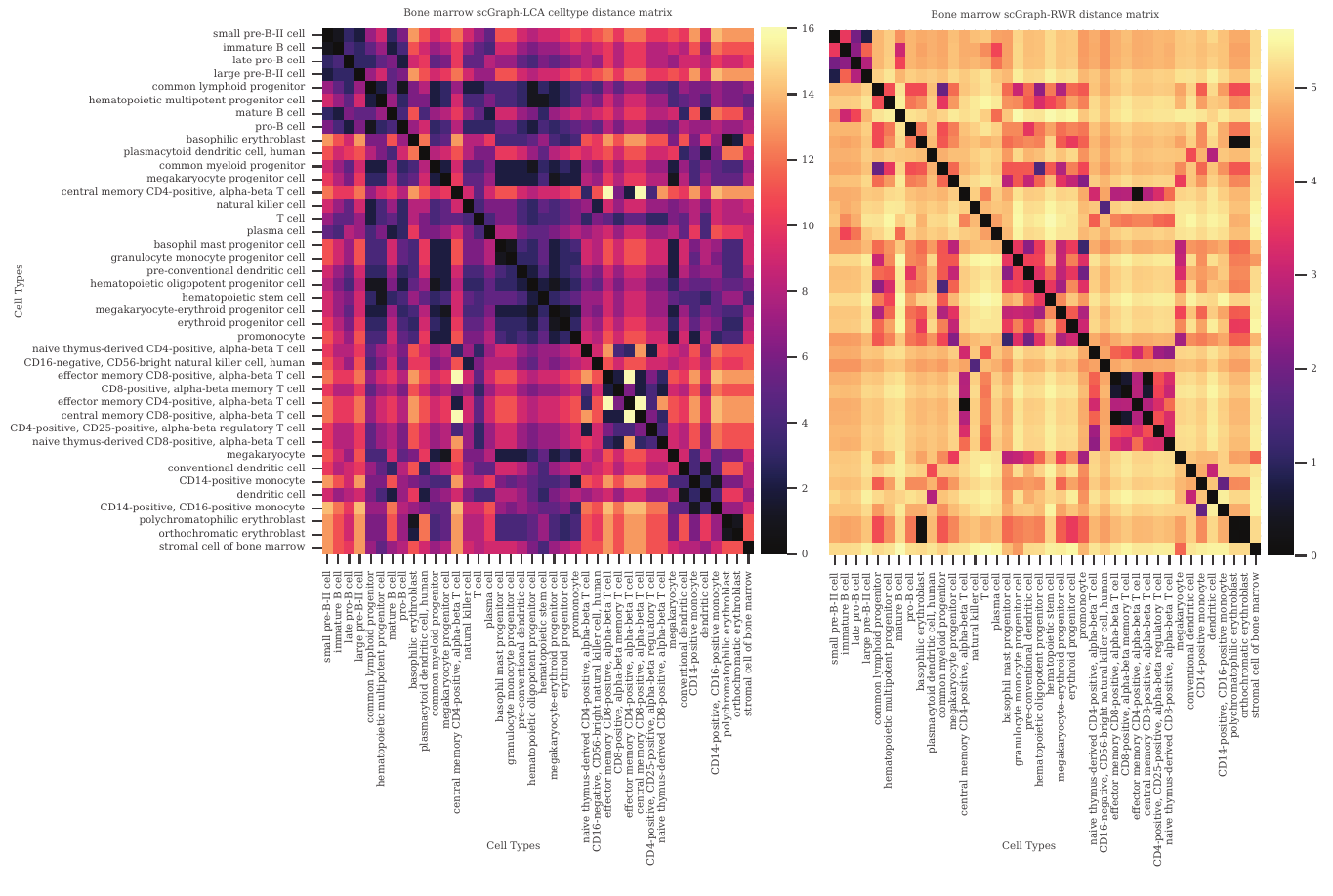} 
\caption{Left: LCA distance reference heatmap of scGraph-LCA in Bone marrow datast; Right: Random walk with restart (scGraph-RWR) of ontology graph distance in Bone marrow dataset.}
\label{fig:scGraph_reference_heatmap}
\end{figure}

We also demonstrated that the learned embeddings of scOntoFM well reflect the intrinsic biologically hierarchical structure of hematopoietic related cell types (AFigure \ref{fig:BoneMarrow_emb_heatmap}c), showing strong empirical consistency with ontology-based distances (AFigure \ref{fig:scGraph_reference_heatmap}). Furthermore, the ontology-derived distance heatmap in AFigure \ref{fig:BoneMarrow_emb_heatmap}b closely aligns with the G2/M and S-phase cell cycle scores reported in the original study AFigure \ref{fig:BoneMarrow_emb_heatmap}d. Taken together, these results indicate that our model could effectively capture the change in cell states along continuous developmental trajectories.

\subsection{Novel cell identification results}
\begin{figure}[h]
\centering
\includegraphics[width=1\columnwidth]{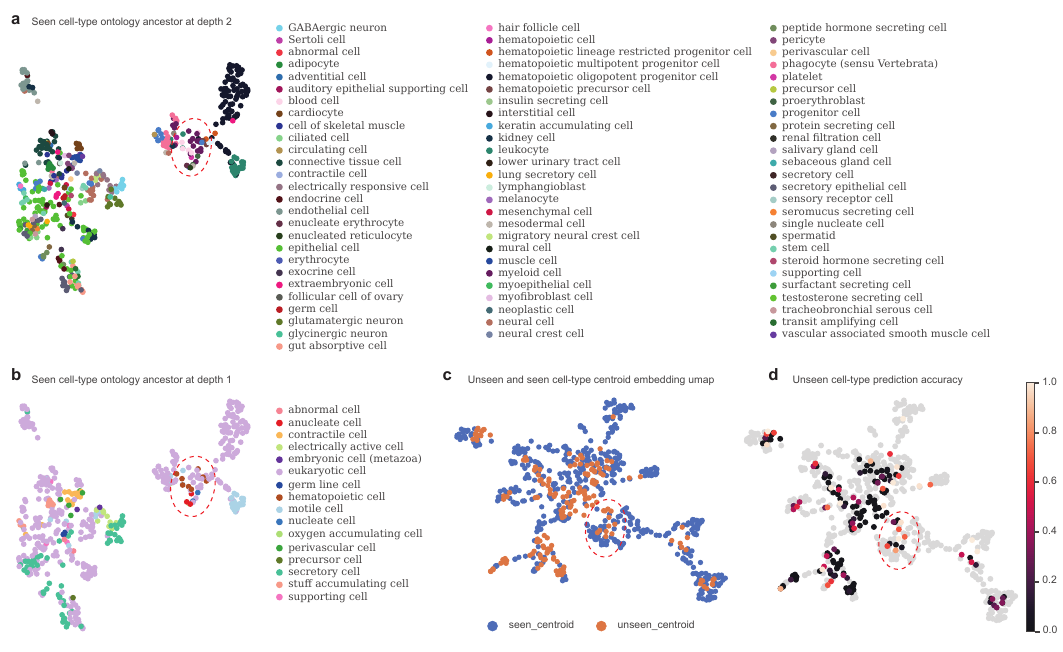}
\caption{(a) Umap visualization of seen cell-type centroid embedding, cell-types are colored by its ontology ancestor at depth 2. (b) Same as (a), but cell-types are colored by its ontology ancestor at depth 1. (c) Co-embedding seen centroid embeddings and unseen cell-type centroid embedding, dots are colored by seen centroid (\textcolor{blue}{blue}) and unseen centroid (\textcolor{orange}{orange}). (d) Our model predicted accuracy heatmap for unseen cell-types. The \textcolor{red}{red} circles in all the umaps are the cell-types coming from hematopoietic or myeloid cell lineage, in which the unseen cell-types (megakaryocyte progenitor cell, immature neutrophil, CD14-positive, CD16-low monocyte) in Figure \ref{fig:case_LCA_cluster_plot} are selected from.}
\label{fig:coarse_seenEmb_umap_plot}
\end{figure}

\begin{figure}[h]
\centering
\includegraphics[width=1\columnwidth]{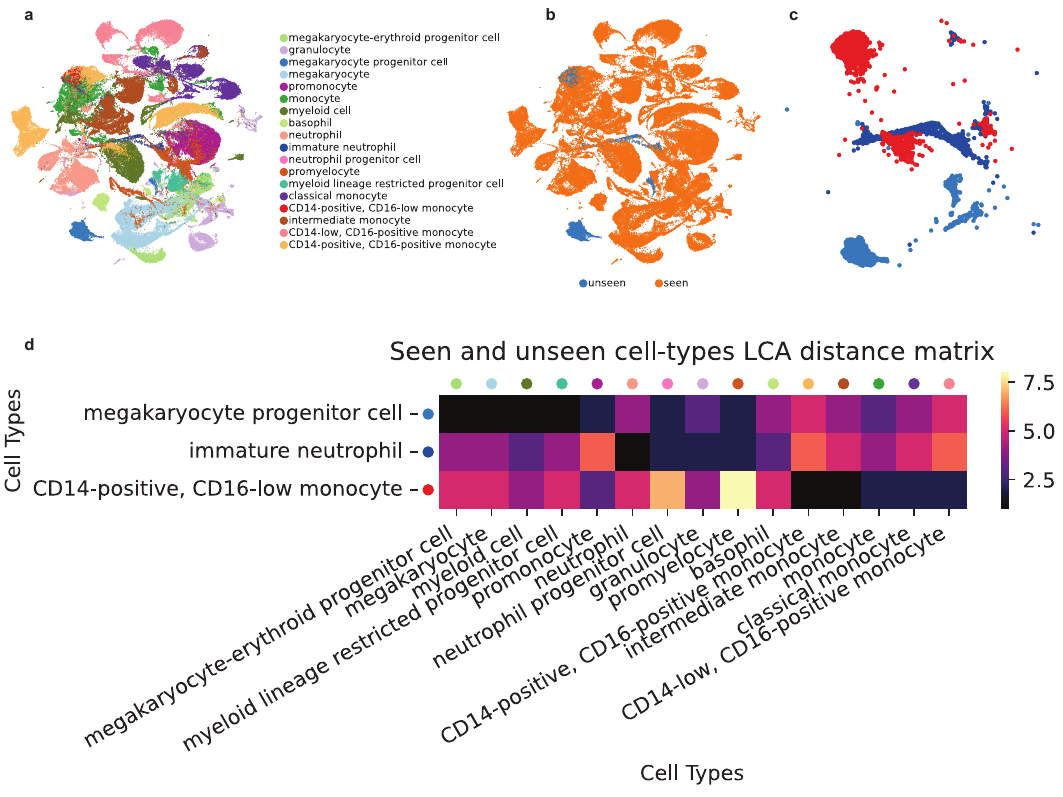}
\caption{(a) Umap of our model embeddings on three unseen cell types and the top 5 closest seen cell types for each unseen cell type. (b) Umap visualization of seen (\textcolor{orange}{orange}) and unseen (\textcolor{blue}{blue}) labels. (c) Umap visualization of the three query unseen cell types. (d) LCA distance matrix of unseen cell types and the top 5 closest seen cell types for each unseen cell type.}
\label{fig:case_LCA_cluster_plot}
\end{figure}

In AFigure \ref{fig:coarse_seenEmb_umap_plot}(a, b), we first demonstrate that our model has learned biological meaningful structure of known cell types, cell-type labels are aggregated into coarser level according to the lineage of ontology tree. Then built on known cell-type centroid embeddings and the comprehensive LCA-based ontology affinity matrix introduced in Appendix \ref{suppl:novel_cell}, scOntoFM could infer the unseen cell-type centroids as shown in AFigure \ref{fig:coarse_seenEmb_umap_plot}(c, d). Concretely, we conducted a specific case study on three unseen hematopoietic cell types: megakaryocyte progenitor cells, immature neutrophils, and CD14-positive/CD16-low monocytes. As AFigure \ref{fig:case_LCA_cluster_plot} shows, our model accurately positions unseen cells near their biological relatives. This alignment with ground-truth ontology distances confirms the model's ability to correctly infer the biological neighborhood of unknown cell types

\subsection{Gene function prediction results}
\label{suppl:geneFunc_plot}
\begin{figure}[h]
\centering
\includegraphics{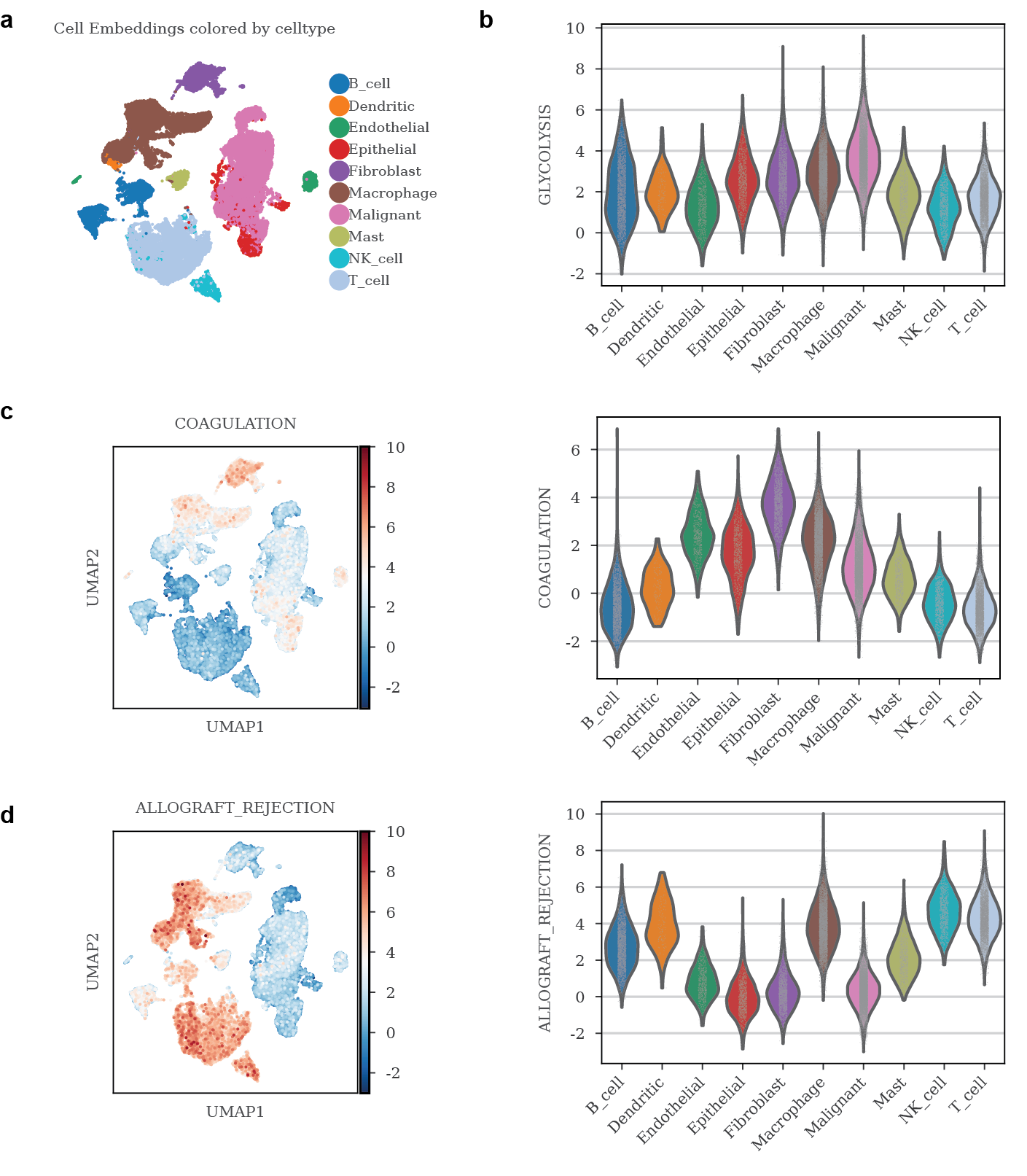} 
\caption{(a) Our model's umap embedding visualization colored by cell-types. (b) Violin plot of \textit{GLYCOLYSIS} hallmark enrichment score distribution for each cell type. (c) Left: \textit{COAGULATION} hallmark enrichment score heatmap. Right: Violin plot of \textit{COAGULATION} hallmark enrichment score distribution for each cell type. (d) Left: \textit{ALLOGRAFT\_REJECTION} hallmark enrichment score heatmap. Right: Violin plot of \textit{ALLOGRAFT\_REJECTION} hallmark enrichment score distribution for each cell type.}
\label{fig:geneFunc_violin_plot}
\end{figure}

In AFigure \ref{fig:geneFunc_violin_plot}, we show additional results of the \textit{COAGULATION} hallmark gene set and the \textit{ALLOGRAFT\_REJECTION} hallmark gene set enrichment score in the lung adenocarcinoma dataset \cite{Lung-Kim2020}. The enrichment scores for the \textit{ALLOGRAFT\_REJECTION} hallmark are notably elevated in immune-related populations—specifically macrophages, T cells, and NK cells—while remaining low in non-immune lineages. Particularly, the \textit{COAGULATION} hallmark is highly enriched in the cancer-associated fibroblasts (CAFs), which contributes hypercoagulable state and facilitate tumor immune evasion \cite{coagulation-pathway-Yang2023}. These distinct enrichment patterns confirm that our model accurately disentangles the complex functional heterogeneity within the tumor microenvironment.

\subsubsection{GO namespace classification by aggregating gene embeddings of the GO terms.}
Beyond classifying semantic meaning of single gene to GO terms, we further evaluated GO namespace classification by aggregating the gene embeddings for specific GO terms. The results are reported as mean±std over 5 random seeds. As ATable \ref{tab:aggregated_GO_embedding} shows, scOntoFM achieves highly competitive results, ranking second only to scGPT, while still leading other Geneformer-like backbone. The performance difference is expected given scGPT's significantly larger vocabulary size (~60,000 {\em vs.} ~20,000 genes), and embedding dimensionality (512 {\em vs.} 256) compared to scOntoFM.
\begin{table}[h]
\centering
\caption{GO namespace classification by aggregated gene embeddings. \textbf{bold} indicates the best performance and \underline{underlining} denotes the second-best results}
\label{tab:aggregated_GO_embedding}
\scalebox{0.8}{
\begin{tabular}{lcccc}
\toprule
\textbf{Model} & \textbf{F1 Micro} & \textbf{F1 Macro} & \textbf{AP Micro} & \textbf{AP Macro} \\
\midrule
Ours 
& \underline{0.783} $\pm$ 0.01 & \underline{0.632} $\pm$ 0.02 & \underline{0.832} $\pm$ 0.01 & \underline{0.632} $\pm$ 0.02 \\

Geneformer 
& 0.767 $\pm$ 0.01 & 0.593 $\pm$ 0.02 & \underline{0.832} $\pm$ 0.01 & 0.613 $\pm$ 0.02 \\

scCello 
& 0.781 $\pm$ 0.01 & 0.627 $\pm$ 0.02 & 0.823 $\pm$ 0.01 & 0.620 $\pm$ 0.02 \\

scGPT 
& \textbf{0.809} $\pm$ 0.01 & \textbf{0.683} $\pm$ 0.02 & \textbf{0.861} $\pm$ 0.01 & \textbf{0.678} $\pm$ 0.02 \\
\bottomrule
\end{tabular}
}
\end{table}

\subsection{Spatial imputation results}
\begin{figure}[h]
\centering
\includegraphics{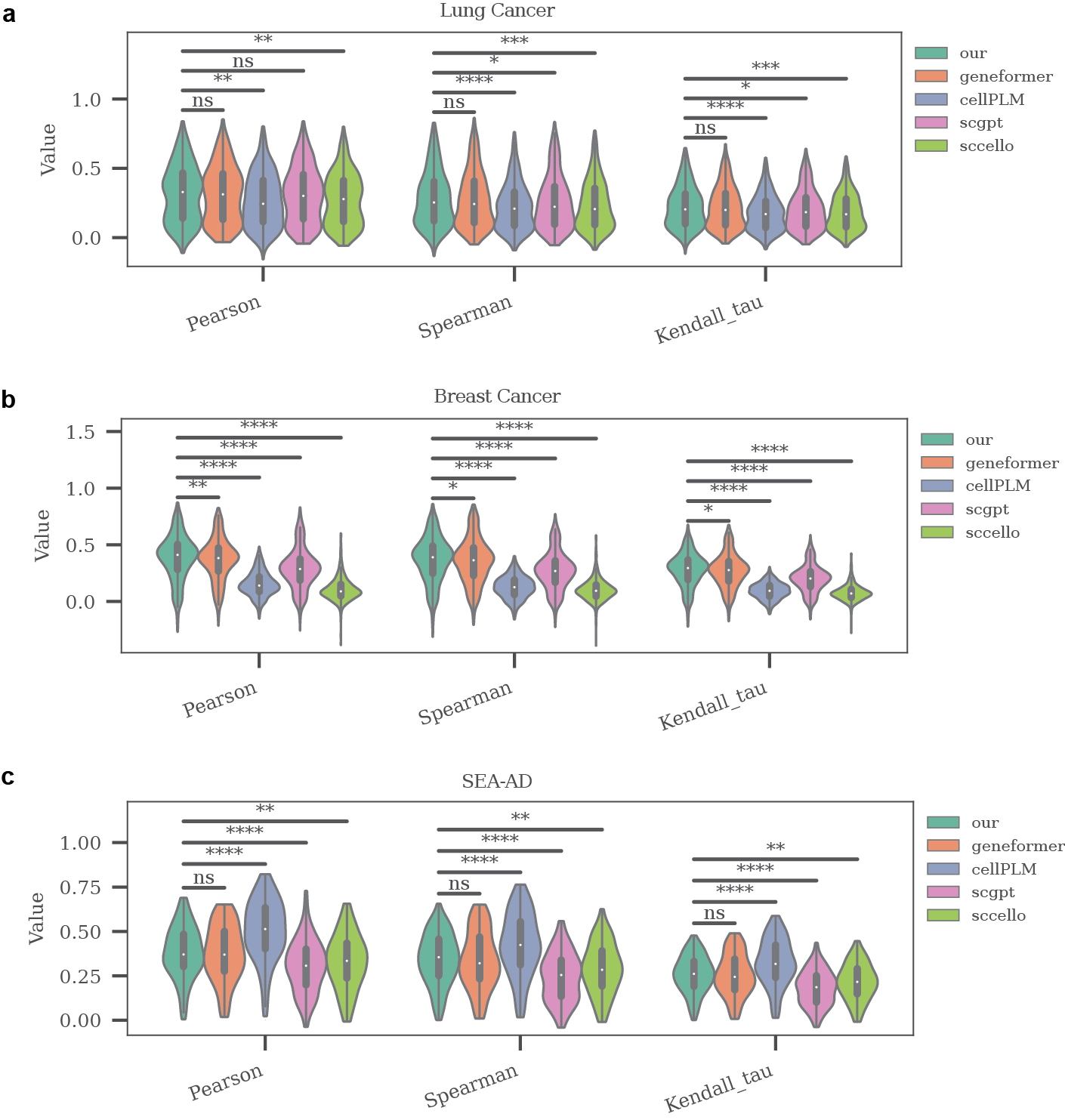} 
\caption{(a) Violin plot of the distribution for Pearson, Spearman, and Kendall’s $\tau$ correlation across all the imputed genes in MERSCOPE Lung cancer 2 dataset. (b) Violin plot of the distribution for Pearson, Spearman, and Kendall’s $\tau$ correlation across all the imputed genes in breast cancer dataset. (c) Violin plot of the distribution for Pearson, Spearman, and Kendall’s $\tau$ correlation across all the imputed genes in SEA-AD dataset. Statistical significance is indicated by stars (*), with “ns” denoting no significant difference.}
\label{fig:imputation_violin_plot}
\end{figure}

In AFigure \ref{fig:imputation_violin_plot}, we show an additional violin plot for spatial gene imputation in terms of Pearson, Spearman, and Kendall’s $\tau$ correlation metrics across the three evaluation datasets (MERSCOPE Lung cancer 2, breast cancer and SEA-AD). We conducted the Mann-Whitney U tests between our method other scFMs on spatial gene imputation task. Our method is statistical significant {\em vs.} other scFMs in both the Lung and breast cancer data except Geneformer.

\begin{table}[h]
\centering
\caption{Spatial gene imputation performance (mean $\pm$ std) of Pearson, Spearman, and Kendall’s $\tau$ correlations in SEA-AD dataset. Higher is better ($\uparrow$). Best results are shown in \textbf{bold}.}
\label{tab:AD_imputation_results}
\begin{tabular}{lccc}
\toprule
Method 
& Pearson $\uparrow$ 
& Spearman $\uparrow$ 
& Kendall's $\tau \uparrow$ \\
\midrule
Ours        
& $0.3820 \pm 0.1450$ 
& $0.3429 \pm 0.1424$ 
& $0.2581 \pm 0.1025$ \\

Geneformer 
& $0.3769 \pm 0.1580$ 
& $0.3388 \pm 0.1627$ 
& $0.2552 \pm 0.1175$ \\

CellPLM    
& $\mathbf{0.5047 \pm 0.1705}$ 
& $\mathbf{0.4251 \pm 0.1765}$ 
& $\mathbf{0.3247 \pm 0.1302}$ \\

scCello    
& $0.3286 \pm 0.1543$ 
& $0.2901 \pm 0.1429$ 
& $0.2167 \pm 0.1026$ \\

scGPT      
& $0.3056 \pm 0.1470$ 
& $0.2435 \pm 0.1361$ 
& $0.1823 \pm 0.0980$ \\
\bottomrule
\end{tabular}
\end{table}

\subsection{Ablation study about the loss terms in scOntoFM (ATable \ref{tab:ablation_full_table})}
\label{suppl:ablation_study}

\begin{table}[t]
\centering
\caption{Ablation study of model loss components.}
\label{tab:ablation_components}

\setlength{\tabcolsep}{4pt}
\renewcommand{\arraystretch}{1.0}

\begin{tabular}{l|ccc|cc}
\hline
\textbf{Model} & MLM & DynamicMargin & OntoAlign & scIB $\uparrow$ & scGraph-LCA $\uparrow$ \\
\hline
MLM only 
& \textcolor{green}{\ding{51}} 
& \textcolor{red}{\ding{55}} 
& \textcolor{red}{\ding{55}} 
& 0.7597 
& 0.4237 \\

w/o OntoAlign 
& \textcolor{green}{\ding{51}} 
& \textcolor{green}{\ding{51}} 
& \textcolor{red}{\ding{55}} 
& \textbf{0.8066} 
& 0.3607 \\

w/o DynamicMargin 
& \textcolor{green}{\ding{51}} 
& \textcolor{red}{\ding{55}} 
& \textcolor{green}{\ding{51}} 
& 0.7841 
& \textbf{0.7211} \\

\textbf{scOntoFM} 
& \textcolor{green}{\ding{51}} 
& \textcolor{green}{\ding{51}} 
& \textcolor{green}{\ding{51}} 
& 0.7958 
& 0.6112 \\
\hline
\end{tabular}
\end{table}

\begin{table}[t]
\centering
\caption{Ablation study of the model's loss terms across the nine evaluation datasets of the overall scIB score and scGraph-LCA metric.}
\label{tab:ablation_full_table}
\scalebox{0.65}{
\begin{tabular}{lcccc|cccc}
\toprule
& \multicolumn{4}{c}{\textbf{scIB}} 
& \multicolumn{4}{c}{\textbf{scGraph-LCA}} \\
\cmidrule(lr){2-5} \cmidrule(lr){6-9}
Dataset 
& scOntoFM 
& w/o OntoAlign 
& w/o DynamicMargin 
& MLM only
& scOntoFM 
& w/o OntoAlign 
& w/o DynamicMargin 
& MLM only \\
\midrule
HIHA            

& 0.7533 & 0.7745 & \textbf{0.7848} & 0.7403

& 0.7157 & 0.4282 & \textbf{0.8301} & 0.4787 \\

Heart           

& 0.8324 & \textbf{0.8549} & 0.8480 & 0.7908

& \textbf{0.7074} & 0.4171 & 0.6824 & 0.5511 \\

Pancreas        

& 0.8140 & \textbf{0.8388} & 0.7767 & 0.7191

& 0.3992 & 0.3957 & \textbf{0.4408} & 0.3351 \\

Lung            

& \textbf{0.8467} & 0.8332 & 0.8119 & 0.8091

& 0.6777 & 0.2689 & \textbf{0.8622} & 0.4425 \\

Liver           

& 0.7730 & 0.7773 & \textbf{0.7837} & 0.7553

& 0.5704 & 0.3691 & \textbf{0.6911} & 0.4847 \\

Breast\_TNBC    

& 0.8724 & \textbf{0.8759} & 0.8272 & 0.8530

& 0.5112 & 0.2792 & \textbf{0.7478} & 0.2966 \\

Kidney          

& 0.7185 & \textbf{0.7252} & 0.7037 & 0.6629

& 0.6378 & 0.3333 & \textbf{0.7333} & 0.3638 \\

Adipose\_tissue 

& \textbf{0.8696} & 0.8643 & 0.8185 & 0.8154

& 0.6180 & 0.3389 & \textbf{0.8345} & 0.3986 \\

Bone\_marrow    

& 0.6823 & \textbf{0.7153} & 0.7021 & 0.6915

& 0.6630 & 0.4155 & \textbf{0.6673} & 0.4624 \\

\midrule
\textbf{Mean}
& 0.7958 & \textbf{0.8066} & 0.7841 & 0.7597
& 0.6112 & 0.3607 & \textbf{0.7211} & 0.4237 \\
\bottomrule
\end{tabular}
}
\end{table}

\subsubsection{Pareto analysis across the nine tissues in our study}
The highlighted points represent the average values across each ablation setting. The plot (Figure \ref{fig:Pareto_ablation_plot}) demonstrates that scOntoFM achieves an optimal balance between standard integration metrics (scIB) and ontology-based metric (scGraph-LCA).


\subsection{Robustness of scOntoFM to Cell Type Annotation Noise}
Cell type annotation noise and inconsistency across datasets can introduce substantial bias during pre-training. To mitigate this, we pre-trained scOntoFM on the CZI CELLxGENE platform, where metadata curation—particularly cell type standardization—is a large-scale, expert-driven effort involving resources such as the Human Cell Atlas (HCA), Tabula Sapiens, and HuBMAP. Pre-training on this well-curated and standardized corpus not only avoids performance degradation but also enables scOntoFM to better capture hierarchical relationships between cell types and preserve biologically meaningful structure.

\begin{figure}[h]
\centering
\includegraphics[width=\columnwidth]{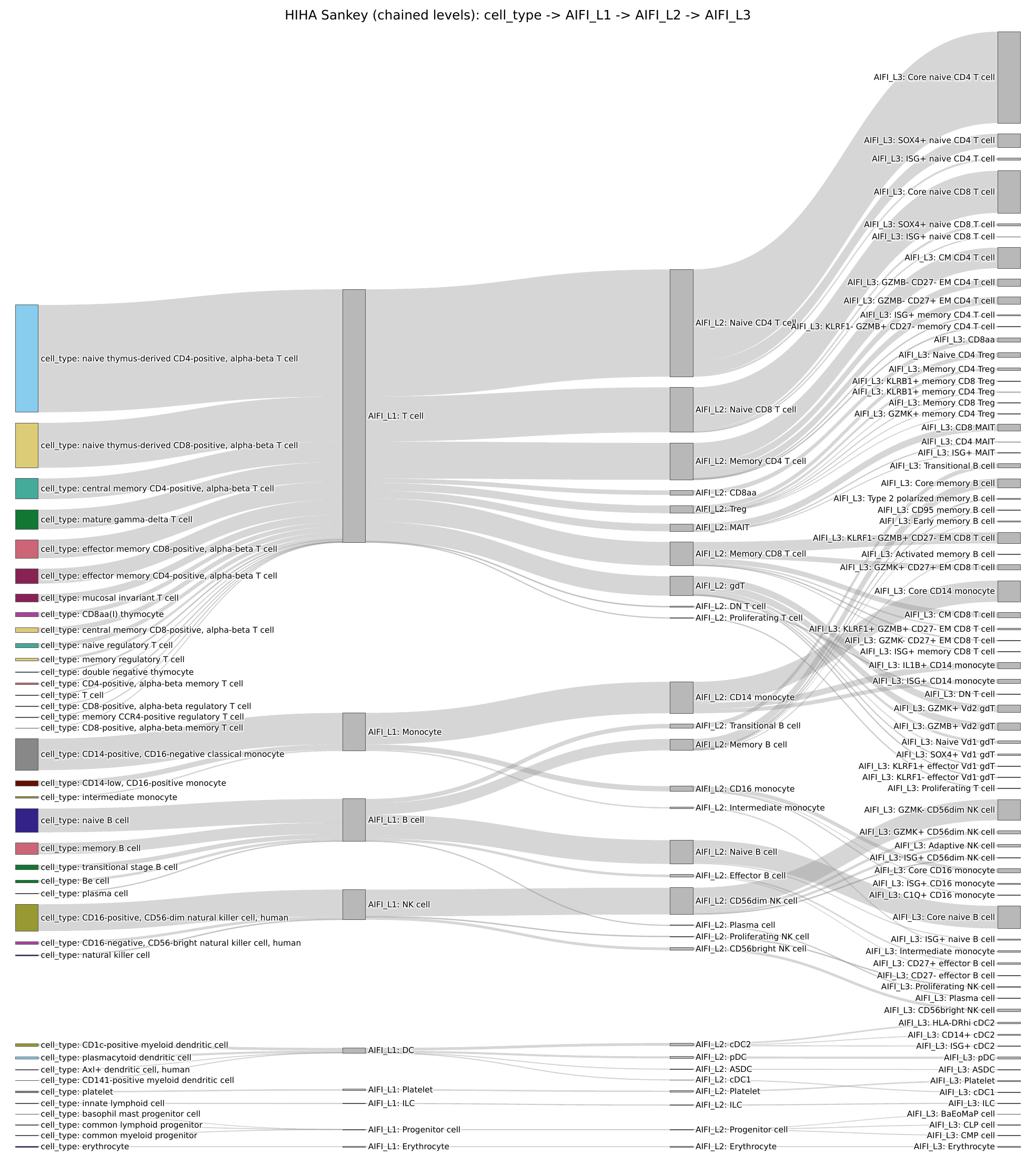} 
\caption{From left to right: Sankey plot of cell type annotation flow from cell\_type (CZI provided ontology labels) to author annotated labels (from coarse to fine-grained) in HIHA dataset }
\label{fig:HIHA_sankey_plot}
\end{figure}

\begin{figure}[h]
\centering
\includegraphics[width=\columnwidth]{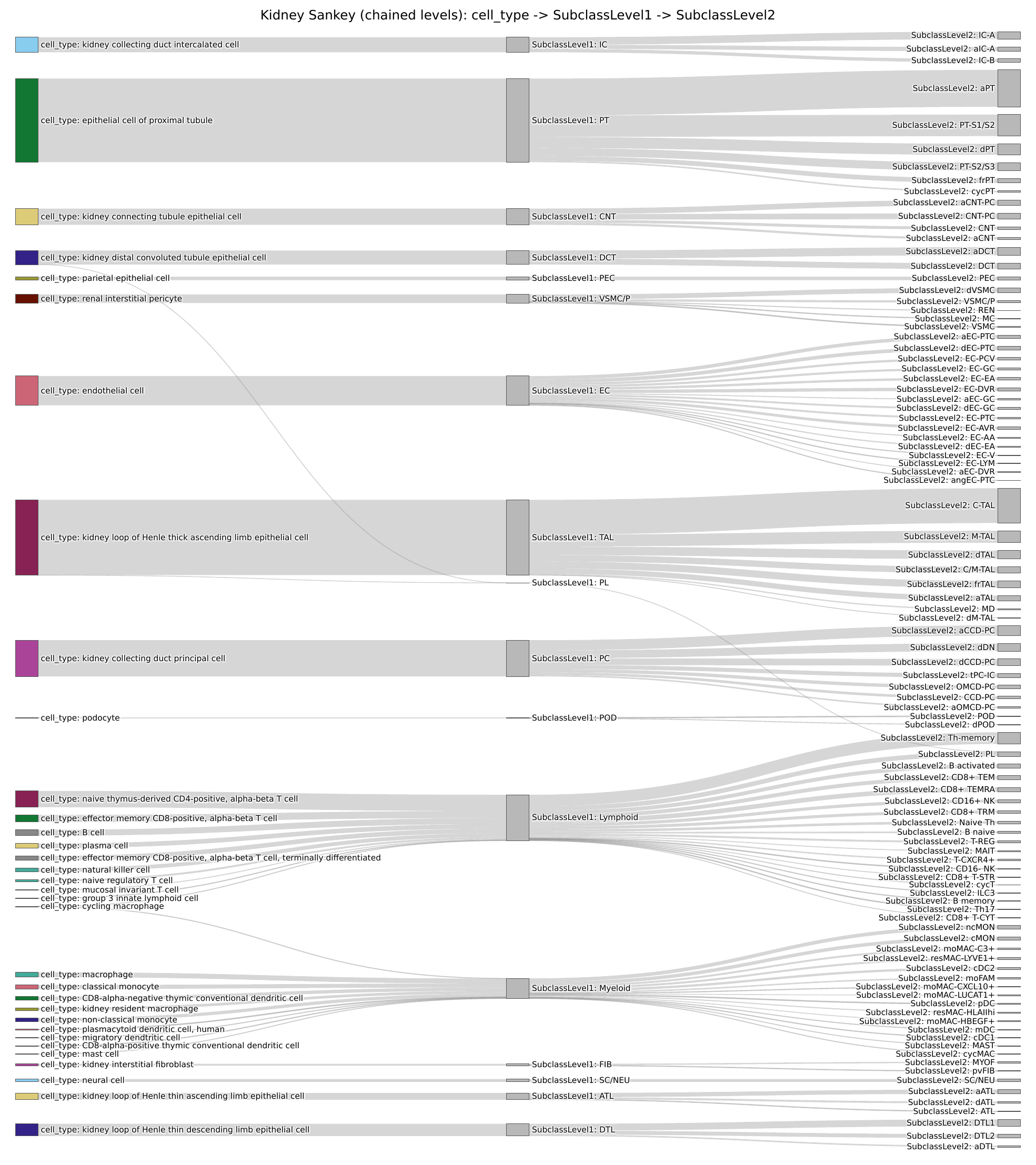} 
\caption{From left to right: Sankey plot of cell type annotation flow from cell\_type (CZI provided ontology labels) to author annotated labels (from coarse to fine-grained) in Kidney dataset }
\label{fig:Kidney_sankey_plot}
\end{figure}

We evaluate robustness from two complementary aspects. First, robustness across annotation levels in zero-shot clustering: scOntoFM consistently maintains strong performance under varying label granularities (Table \ref{tab:hiha_kidney_robust_labels}), particularly at fine-grained levels ({\em e.g.}, AIFI\_L1/L2 in HIHA and SubclassLevel2 in Kidney), highlighting its ability to capture subtle cell states and complex lineage structures. This is further supported by high agreement scores (ARI) between cell ontology references and annotations at different resolutions (see ATable \ref{tab:agreement_scores}, Figure \ref{fig:HIHA_sankey_plot}, \ref{fig:Kidney_sankey_plot}), demonstrating stability under label mismatch. Second, resilience to label noise: under controlled noise injection on the Azimuth PBMC dataset, scOntoFM maintains high accuracy and stable performance as noise increases (ATable \ref{tab:label_noise_analysis}). Notably, at higher noise levels ({\em e.g.}, 50\%), the model predominantly assigns cells to their parent categories, reflecting biologically consistent behavior when fine-grained labels become unreliable.







\begin{table}[ht]
\centering
\caption{Agreement scores between reference and target annotations across datasets.}
\label{tab:agreement_scores}
\scalebox{0.8}{
\begin{tabular}{lccccc}
\toprule
\textbf{Dataset} & \textbf{Reference} & \textbf{Target} & \textbf{\# Ref Labels} & \textbf{\# Target Labels} & \textbf{Agreement Score} \\
\midrule
HIHA 
& cell\_type & AIFI\_L1 & 38 & 9 & 0.259 \\

HIHA 
& cell\_type & AIFI\_L2 & 38 & 29 & 0.970 \\

HIHA 
& cell\_type & AIFI\_L3 & 38 & 71 & 0.830 \\

Kidney 
& cell\_type & SubclassLevel1 & 33 & 17 & 0.933 \\

Kidney 
& cell\_type & SubclassLevel2 & 33 & 97 & 0.407 \\
\bottomrule
\end{tabular}
}
\end{table}

\begin{table}[ht]
\centering
\caption{Impact of noise proportion on performance and hierarchical error distribution.}
\label{tab:label_noise_analysis}
\scalebox{0.8}{
\begin{tabular}{lccccc}
\toprule
\textbf{Noise} & \textbf{Accuracy (mean $\pm$ std)} & \textbf{Macro F1 (mean $\pm$ std)} & \textbf{Parent (mean $\pm$ std)} & \textbf{Child (mean $\pm$ std)} & \textbf{Sibling (mean $\pm$ std)} \\
\midrule
0.1 & 0.941 $\pm$ 0.001 & 0.806 $\pm$ 0.017 & 0.001 $\pm$ 0.000 & 0.003 $\pm$ 0.001 & 0.014 $\pm$ 0.000 \\
0.2 & 0.940 $\pm$ 0.001 & 0.752 $\pm$ 0.013 & 0.001 $\pm$ 0.000 & 0.003 $\pm$ 0.001 & 0.015 $\pm$ 0.000 \\
0.3 & 0.934 $\pm$ 0.002 & 0.691 $\pm$ 0.008 & 0.001 $\pm$ 0.000 & 0.005 $\pm$ 0.001 & 0.015 $\pm$ 0.001 \\
0.4 & 0.834 $\pm$ 0.036 & 0.572 $\pm$ 0.024 & 0.001 $\pm$ 0.001 & 0.053 $\pm$ 0.021 & 0.018 $\pm$ 0.010 \\
0.5 & 0.150 $\pm$ 0.018 & 0.207 $\pm$ 0.025 & 0.000 $\pm$ 0.000 & 0.552 $\pm$ 0.002 & 0.035 $\pm$ 0.003 \\
\bottomrule
\end{tabular}
}
\end{table}

\section{Broader Impact of scOntoFM}
\label{suppl:broadImpact}
This paper presents work whose goal is to advance the field of Machine Learning and Computational Biology. There are many potential societal consequences of our ontology-based foundation model, particularly in the interpretation of complex single-cell atlases. We hope that by integrating hierarchical biological priors, our model will aid researchers in uncovering accurate biological insights and accelerating disease characterization. However, we acknowledge that pretraining such foundation models is computationally intensive, and future efforts should address the energy efficiency and accessibility of these large-scale systems.


\newpage
\section*{NeurIPS Paper Checklist}

\begin{enumerate}

\item {\bf Claims}
    \item[] Question: Do the main claims made in the abstract and introduction accurately reflect the paper's contributions and scope?
    \item[] Answer: \answerYes{} 
    \item[] Justification: In this paper, we propose scOntoFM, a novel foundation model that integrates hierarchical cell ontologies directly into the pre-training framework. This main claim is accurately and clearly stated and emphasized in the abstract and introduction section. It reflects our core contribution and scopes. 

\item {\bf Limitations}
    \item[] Question: Does the paper discuss the limitations of the work performed by the authors?
    \item[] Answer: \answerYes{} 
    \item[] Justification: The paper includes an explicit \textit{Limitations} paragraph in Section \ref{sec:conclusion}, discussing reliance on explicit cell-type labels and scalability constraints from the indexing map, and outlining future directions.

\item {\bf Theory assumptions and proofs}
    \item[] Question: For each theoretical result, does the paper provide the full set of assumptions and a complete (and correct) proof?
    \item[] Answer: \answerNA{} 
    \item[] Justification: This paper does not present theorem- or proposition-style theoretical claims requiring formal proofs. Instead, it primarily focuses on methodological development and empirical evaluation in the context of computational biology and single-cell foundation models.

    \item {\bf Experimental result reproducibility}
    \item[] Question: Does the paper fully disclose all the information needed to reproduce the main experimental results of the paper to the extent that it affects the main claims and/or conclusions of the paper (regardless of whether the code and data are provided or not)?
    \item[] Answer: \answerYes{} 
    \item[] Justification: We provide all necessary details to reproduce our work. The method is fully specified in Section \ref{sec:methods} (including objectives), and experimental setup details are provided in Section \ref{sec:experiments} plus Appendix \ref{suppl:experiments}/\ref{suppl:results} (model/training configuration, dataset lists/sources, task protocols, and additional quantitative results).

\item {\bf Open access to data and code}
    \item[] Question: Does the paper provide open access to the data and code, with sufficient instructions to faithfully reproduce the main experimental results, as described in supplemental material?
    \item[] Answer: \answerNo{} 
    \item[] Justification: Our code and datasets will be released upon acceptance.

\item {\bf Experimental setting/details}
    \item[] Question: Does the paper specify all the training and test details (e.g., data splits, hyperparameters, how they were chosen, type of optimizer) necessary to understand the results?
    \item[] Answer: \answerYes{} 
    \item[] Justification: All the  Experimental setting/details about pre-training and downstream tasks are illustrated in Appendix \ref{suppl:experiments}.

\item {\bf Experiment statistical significance}
    \item[] Question: Does the paper report error bars suitably and correctly defined or other appropriate information about the statistical significance of the experiments?
    \item[] Answer: \answerYes{} 
    \item[] Justification: The paper reports uncertainty for key experiments using mean $\pm$ standard deviation (e.g., Table \ref{tab:geneFunc_performance} with 20-fold CV for gene function prediction, Table \ref{tab:impute_correlation_metrics} for spatial imputation), with additional distributional violin plots in Figure \ref{fig:geneFunc_violin_plot} for hallmark gene set enrichment score, and error bars in Figure \ref{fig:novelCL} for novel cell identification, as well the statistical significance test in Figure \ref{fig:imputation_violin_plot} for spatial gene imputation performance.

\item {\bf Experiments compute resources}
    \item[] Question: For each experiment, does the paper provide sufficient information on the computer resources (type of compute workers, memory, time of execution) needed to reproduce the experiments?
    \item[] Answer: \answerYes{} 
    \item[] Justification: We report the major pre-training details of scOntoFM (six NVIDIA H200 GPUs, effective batch size, and $\sim$30-hour pretraining). We also report the computational resources required when constructing the Cell-Dataset Index Map, which takes about 5 minutes and uses 6.5 GB of peak memory on an Intel Xeon Gold 6252 CPU. The final file size is 132 MB. 
    
\item {\bf Code of ethics}
    \item[] Question: Does the research conducted in the paper conform, in every respect, with the NeurIPS Code of Ethics \url{https://neurips.cc/public/EthicsGuidelines}?
    \item[] Answer: \answerYes{} 
    \item[] Justification: We follow the NeurIPS Code of Ethics throughout the entire project.

\item {\bf Broader impacts}
    \item[] Question: Does the paper discuss both potential positive societal impacts and negative societal impacts of the work performed?
    \item[] Answer: \answerYes{} 
    \item[] Justification: Broader impacts are addressed throughout the introduction and discussion. And we also provide a detailed impact statement in Appendix \ref{suppl:broadImpact}.
    
\item {\bf Safeguards}
    \item[] Question: Does the paper describe safeguards that have been put in place for responsible release of data or models that have a high risk for misuse (e.g., pre-trained language models, image generators, or scraped datasets)?
    \item[] Answer: \answerNA{} 
    \item[] Justification: This work focuses on representation learning for single-cell transcriptomic analysis and does not involve high-risk generative modeling or biological sequence design. The released code and model checkpoints are intended solely for research use in computational biology. All datasets used in this study are publicly available or distributed under their respective data usage agreements, and no identifiable personal information is included. As such, we do not believe the model poses substantial misuse risks requiring specialized safeguards.

\item {\bf Licenses for existing assets}
    \item[] Question: Are the creators or original owners of assets (e.g., code, data, models), used in the paper, properly credited and are the license and terms of use explicitly mentioned and properly respected?
    \item[] Answer: \answerYes{} 
    \item[] Justification: All external assets used in this work, including datasets, software libraries, and pre-existing models, are appropriately cited and credited in the paper. 

\item {\bf New assets}
    \item[] Question: Are new assets introduced in the paper well documented and is the documentation provided alongside the assets?
    \item[] Answer: \answerNo{} 
    \item[] Justification:  Code and datasets will be released upon acceptance. Clear documentations will be provided along the assets.

\item {\bf Crowdsourcing and research with human subjects}
    \item[] Question: For crowdsourcing experiments and research with human subjects, does the paper include the full text of instructions given to participants and screenshots, if applicable, as well as details about compensation (if any)? 
    \item[] Answer: \answerNA{} 
    \item[] Justification: We did not perform crowdsourcing experiments and research with human subjects.

\item {\bf Institutional review board (IRB) approvals or equivalent for research with human subjects}
    \item[] Question: Does the paper describe potential risks incurred by study participants, whether such risks were disclosed to the subjects, and whether Institutional Review Board (IRB) approvals (or an equivalent approval/review based on the requirements of your country or institution) were obtained?
    \item[] Answer: \answerNA{} 
    \item[] Justification: This paper does not involve crowdsourcing nor research with human subjects.

\item {\bf Declaration of LLM usage}
    \item[] Question: Does the paper describe the usage of LLMs if it is an important, original, or non-standard component of the core methods in this research? Note that if the LLM is used only for writing, editing, or formatting purposes and does \emph{not} impact the core methodology, scientific rigor, or originality of the research, declaration is not required.
    \item[] Answer: \answerNA{} 
    \item[] Justification: Large language models (LLMs) are not a component of the proposed methodology, experimental pipeline, or scientific contributions of this work.

\end{enumerate}

\end{document}